\documentclass[11pt]{article}

\usepackage{arxiv}

\usepackage[utf8]{inputenc} % allow utf-8 input
\usepackage[T1]{fontenc}    % use 8-bit T1 fonts
\usepackage{hyperref} 
\usepackage{booktabs}       % professional-quality tables
\usepackage{nicefrac}       % compact symbols for 1/2, etc.
\usepackage{microtype}      % microtypography

\usepackage{times}
\usepackage{latexsym}
\usepackage{graphicx, color}
\usepackage{booktabs}
\usepackage{wrapfig}
\usepackage{amsmath}
\usepackage{amsfonts}
\usepackage{mathtools}
\usepackage[authormarkuptext=name,addedmarkup=bf,authormarkupposition=left]{changes}
\usepackage[small]{caption}
\usepackage{subcaption}
\usepackage{url}
\usepackage{xspace}
\usepackage{booktabs}
\usepackage{tabularx}
\usepackage{arydshln}
\usepackage{tikz}
\usepackage{verbatim}
\usepackage{multirow}
\usepackage{newfloat}
\usepackage[shortlabels]{enumitem}
\usepackage[vlined, linesnumbered]{algorithm2e}
\usepackage[compress,square,comma]{natbib}
\usetikzlibrary{arrows,decorations.pathmorphing,backgrounds,positioning,fit,matrix,calc,shapes.geometric,shapes.misc, positioning,decorations.pathreplacing,trees,intersections}

\usepackage{tikz-qtree-compat}
\usepackage{pgfplots}
\usepgfplotslibrary{fillbetween}

\hypersetup{colorlinks=true,allcolors=[RGB]{0,20,122}}

\DeclareFloatingEnvironment[fileext=frm,placement={!ht},name=Algorithm]{alg}

\definechangesauthor[name={Matt}, color={blue}]{matt} 
\definechangesauthor[name={Ignacio}, color={purple}]{ignacio}
\definechangesauthor[name={Clemens}, color={orange}]{clemens}

\begin{document}
\setlength{\abovedisplayskip}{2pt}
\setlength{\belowdisplayskip}{2pt}
\setlength{\abovedisplayshortskip}{2pt}
\setlength{\belowdisplayshortskip}{2pt}
\setcounter{bottomnumber}{2}
\renewcommand\bottomfraction{.5}
% \title{Routing Networks and the Challenges of Composing Trainable, Modular Functions}
\title{Routing Networks and the Challenges of Modular and Compositional Computation}

\author{
  Clemens Rosenbaum \\
  College of Information and Computer Sciences\\
  University of Massachusetts Amherst\\
  \texttt{cgbr@cs.umass.edu} \\
  %% examples of more authors
   \And
 Ignacio Cases \\
  Linguistics and Computer Science Departments\\
  Stanford University\\
  \texttt{cases@stanford.edu} \\
   \AND
   Matthew Riemer, Tim Klinger \\
   IBM Research\\
   \texttt{\{mdriemer,tklinger\}@us.ibm.com}
  %% \And
  %% Coauthor \\
  %% Affiliation \\
  %% Address \\
  %% \texttt{email} \\
}

%\editor{Leslie Pack Kaelbling}

\maketitle

\begin{abstract}%   <- trailing '%' for backward compatibility of .sty file
Compositionality is a key strategy for addressing combinatorial complexity and the curse of dimensionality. Recent work has shown that compositional solutions can be learned and offer substantial gains across a variety of domains, including multi-task learning, language modeling, visual question answering, machine comprehension, and others. However, such models present unique challenges during training when both the module parameters and their composition must be learned jointly. In this paper, we identify several of these issues and analyze their underlying causes. Our discussion focuses on routing networks, a general approach to this problem, and examines empirically the interplay of these challenges and a variety of design decisions. In particular, we consider the effect of how the algorithm decides on module composition, how the algorithm updates the modules, and if the algorithm uses regularization.
\end{abstract}


\section{Introduction}\label{sec:intro}
% HUME version
In machine learning, and in deep learning in particular, modularity is becoming a key approach to reducing complexity and improving generalization by encouraging a decomposition of complex systems into specialized sub-systems. Such observations have similarly motivated studies in human cognition (compare \citep{BechtelRichardson2010} for an overview). In theory, models without special affordances for modularity can  learn to specialize groups of neurons for specific subtasks, but in practice this appears not to be the case. To address this issue, a number of models have been introduced to enforce specialization  \citep{Hinton91,Miikkulainen1993,Jordan94,conditionalcomputation,davis2013low,Andreas2015,EBengio,largeneuralnets,Pathnet}.  Much of this earlier work has either fixed the composition strategy and learned the modules (neurons or whole networks) or fixed the modules and learned the composition strategy. But in its most general form, the compositionality problem is to jointly learn both the parameters of the modules  and a strategy for their composition with the goal of solving new tasks. Recently we proposed a new paradigm called, \textit{routing} \citep{routingnets}, which is to our knowledge the first approach to jointly optimize both modules and their compositional strategy in the general setting.

%Modular organization studies play a central role in research strategies on human cognition, where complex systems are hypothesized to divide labor through decomposition into specialized subsystems \citep[i.a.]{BechtelRichardson2010}. 

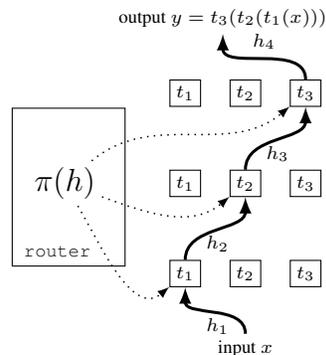
\begin{wrapfigure}{r}{0.35\textwidth}
\vspace{-4mm}
    \centering
    % \vspace{-11mm}
    \begin{tikzpicture}
	\scriptsize
	%\node (full_input) at (-2.0, 2.5) {input: $v$, $t$};
	% the input
	\node (input) at (-.4,0) {input $x$};
	% the fm set: layer 1
	\node[draw,rectangle] (fm11) at (-1.2, 1) {$t_{1}$};
	\node[draw,rectangle] (fm12) at (-0.4,1) {$t_{2}$};
	\node[draw,rectangle] (fm13) at (0.4,1) {$t_{3}$};

    \node[draw,rectangle] (fm21) at (-1.2,2.2) {$t_{1}$};
	\node[draw,rectangle] (fm22) at (-0.4,2.2) {$t_{2}$};
	\node[draw,rectangle] (fm23) at (0.4,2.2) {$t_{3}$};

    \node[draw,rectangle] (fm31) at (-1.2,3.4) {$t_{1}$};
	\node[draw,rectangle] (fm32) at (-0.4,3.4) {$t_{2}$};
	\node[draw,rectangle] (fm33) at (0.4,3.4) {$t_{3}$};
	
	\node (router) at (-2.8, 2.2) {\large $\pi(h)$};
    \draw (-3.5,1.1) -- (-2,1.1) -- (-2,3.2) -- (-3.5,3.2) -- (-3.5,1.1);
    \node (routerbox) at (-2.9, 1.3) {\texttt{router}};
	\draw[-{latex[scale=1.5]},line width=0.2mm,dotted] (router) [out=-60, in=-140] to (fm11);
	\draw[-{latex[scale=1.5]},line width=0.2mm,dotted] (router) [out=-20, in=-140] to (fm22);
	\draw[-{latex[scale=1.5]},line width=0.2mm,dotted] (router) [out=40, in=-140] to (fm33);
	% the information flow
	\draw[-{latex[scale=1.5]},line width=0.4mm] (input) [out=90, in=-90] to node[below] {$~h_1$} (fm11);
	\draw[-{latex[scale=1.5]},line width=0.4mm] (fm11) [out=90, in=-90] to node[below] {$~h_2$} (fm22);
	\draw[-{latex[scale=1.5]},line width=0.4mm] (fm22) [out=90, in=-90] to node[below] {$~h_3$} (fm33);
	% the output
	\node (output) at (-0.7, 4.4) {output $y=t_3(t_2(t_1(x)))$};
	\draw[-{latex[scale=1.5]},line width=0.4mm] (fm33) [out=90, in=-90] to node[above] {$h_4$} (output);
\end{tikzpicture}
    \vspace{-3mm}
    \captionof{figure}{A \textit{routing network}. The router consists of a parameterized decision maker that iteratively selects modules (i.e. trainable functions). Each selected function is then applied to the latest activation, resulting in a new activation, which can then again be transformed. The training of a routing network happens \textit{online}, i.e., the output of the model is used to train the transformations using backpropagation and Stochastic Gradient Descent (SGD), and is simultaneously used to provide feedback to the decision maker.}
    \label{fig:routing-general}
    \vspace{-7mm}
\end{wrapfigure}

A routing network (depicted in Figure \ref{fig:routing-general}) consists of a set of modules (parameterized functions) from which a router (the composition strategy) can choose a composition. In a neural network setting, the modules are sub-networks and the router assembles them into a model that processes the input. In making these decisions the routing network can  be viewed as selecting a model (path) from the set of combinatorially many such models, one for each possible sequence of modules. This connects routing to a body of related work in conditional computation, meta-learning, architecture search and other areas.
Because routing networks jointly train their modules and the module composition strategy, they face a set of challenges which non-routed networks do not. In particular, training a routing network is non-stationary from both the perspective of the router, and from the perspective of the modules, because the optimal composition strategy depends on the module parameters and vice versa. 

There is relatively little research on the unique challenges faced when training these networks, which are highly dynamic. In this paper, we conduct an extensive empirical investigation of these challenges with a focus on routing networks. In Sections \ref{sec:background} we review notable related ideas that share a similar motivation to routing networks. In Section \ref{sec:challenges}, we identify five main challenges for routing. One such challenge is to stabilize the interaction of the module and composition strategy training. Another is \emph{module collapse} \citep{modularnets}, which occurs when the choices made by the composition strategy lack diversity. Overfitting is also a problem which can be severe in a routed network because of the added flexibility of composition order. We also discuss the difficulty extrapolating the learning behavior of heterogeneous modules to better select those with greater potential. Adding to these difficulties is the lack of a compelling mathematical framework for models which perform modular and compositional learning.  This paper is the first to consider all of these challenges jointly. One benefit of such a holistic view of the challenges has been that we were able to identify a clear relationship between collapse and overfitting that holds for any form of modular learning. We discuss this in more detail in Section \ref{sec:challenges}.  In Section \ref{sec:routing}, we present a detailed overview of routing, analyzing two strategies for training the router: reinforcement learning and recently introduced reparameterization strategies. There we also identify important design options, such as the choice of optimization algorithm and the router's architecture. In Section \ref{sec:eval}, we empirically compare these different choices and how they influence the main challenges we have identified. We conclude with thoughts on promising directions for further investigation.

\section{Background and Related Work}\label{sec:background}

Routing networks are clearly related to task-decomposition modular networks \citep{JACOBS1991219}, and to mixtures of experts architectures \citep{Hinton91, Jordan94} as well as their modern attention based \citep{Riemer2016} and sparse \citep{largeneuralnets} variants. The gating network in a typical mixtures of experts model takes in the input and chooses an appropriate weighting for the output of each expert network. This is generally implemented as a soft mixture decision as opposed to a hard routing decision, allowing the choice to be differentiable. Although the sparse and layer-wise variant presented in \citep{largeneuralnets} does save some computational burden, the proposed end-to-end differentiable model is only an approximation and does not model important effects such as exploration vs. exploitation trade-offs, despite their impact on the system. Mixtures of experts models, or more generally soft parameter sharing models, have been considered in the multi-task and lifelong learning setting \citep{Schmidhuber,Aljundi,CrossStitch,SLUICE,AAT}, but do not allow for nearly the level of specialization
as those based on routing. This is because they do not eliminate weight sharing across modules and instead only gate the sharing. In practice, this still leads to significant interference across tasks as a result of a limited ability to navigate the transfer-interference trade-off \citep{MER} in comparison to models that make hard routing decisions. 

\subsection{Existing Approaches to Routing Networks}

Routing networks as a general paradigm of composing trainable transformations were first introduced in \citep{routingnets}. Since then, there have been several approaches extending them. \cite{crl} show that a vanilla routing network can learn to generalize over patterns and thus to classes of unseen samples if it is trained with curriculum learning. \cite{modularnets} identify one of the challenges of compositional computation, collapse, and develop a new policy gradient based routing algorithm that utilizes EM techniques to alternatingly group samples to transformations, and then applies them. \cite{ramachandran2018diversity} investigate the problem of architectural diversity over transformations, but use a top-k routing approach. \cite{routingnaacl} show how routing can be paired with high quality linguistic annotations to learn compositional structures that maximize utilization of task-specific information. \cite{modular_meta_learning} combine a routing-like approach with other meta-learning approaches, to allow for quick adaptation to new tasks. However, this approach relies on \textit{pre-trained} composable transformations.

\subsection{Generalized Architecture based Meta-Learning}

Routing networks extend a popular line of recent research focused on automated architecture search, or more generally, architecture-based meta-learning. In this work, the goal is to reduce the burden on the algorithm designers by automatically learning black box algorithms that search for optimal architectures and hyperparameters. These approaches have been learned using reinforcement learning \citep{zophICLR,bakerICLR}, evolutionary algorithms \citep{evolving,Pathnet}, approximate random simulations \citep{SMASH}, and adaptive growth \citep{Adanet}. \cite{evol_multitask_nets} introduced an evolutionary algorithm approach targeting multi-task learning that comes very close to the original formulation in \cite{routingnets}. However, routing networks are a generalization of these approaches \citep{ramachandran2018diversity} and are highly related in particular to the concept of one-shot neural architecture search \citep{SMASH,pham2018,bender2018}. The main distinction we are making here is that the benefit of routing as not solely related to parameter sharing across examples, but as also related to architectural biases inherent to specific network structures that may make them helpful for specific problems.  

\subsection{Biological Plausibility}

The high-level idea of task specific ``routing" 
% as a \icXX{I'm not comfortable with cognitive here as what follows is a non-cognitive account} cognitive function 
is well founded in biological studies and theories of the human brain \citep{bio2001,bio2010,bio2010_2}. The idea is that regions of the brain collaborate in a complex manner by altering the synchrony in neural activity between different areas and thus changing their effective connectivity so that signals are routed in a task specific coordination. It has been found that coincidence of spikes from multiple neurons converging on a post-synaptic neuron have a super-additive effect \citep{aertsen1989,usrey1999,engel2001,salinas2001,fries2005}. As a result, if neurons tuned to the same stimulus synchronize their firing, that stimulus will be more strongly represented in downstream areas. It is thought that local synchrony may help the brain to improve its signal to noise ratio while, at the same time, reducing the number of spikes needed to represent a stimulus \citep{aertsen1989,tiesinga2002,siegel2003}. %Because filtering out noise is the aim of attention mechanisms, it has been proposed that attention might act by synchronizing stimulus representations in the sensory cortex. 

The routing of modules can also be loosely linked to two fundamental aspects of the organization of the brain: functional segregation of neural populations and anatomic brain regions as specialized and independent cortical areas, and functional integration, the complementary aspect that accounts for coordinated interaction \citep{Tononi1994}. Recently, \cite{Kell2018} have been able to replicate human cortical organization in the auditory cortex using a neural model with two distinct, task-dependent pathways corresponding to music and speech processing. When evaluated in real world tasks, their model performs as well as humans and makes human-like errors. This model is also able to predict with good accuracy fMRI voxel responses throughout the auditory cortex, which suggest certain level of convergence with brain-like representational transformations. \cite{Kell2018} use these promising results to suggest that the tripartite hierarchical organization of the auditory cortex commonly found in other species may also become evident in humans once more data is available and more realistic training mechanisms are employed. In particular, their results suggest that the architectural separation of the processing of the streams is compatible with functional segregation observed in the non-primary auditory cortex \citep[and references therein]{Kell2018}. 

Finally, as we will discuss briefly later, routing networks can be seen as a special case of coagent networks \citep{coagents}, which are believed to be more biologically plausible than full end to end backpropogation networks. 

\subsection{Other Similar Forms of Modular Learning}
Modular organization is central in research strategies in cognitive sciences, particularly in neuropsychology, where decomposition and localization emerge as crucial scientific research strategies \citep{BechtelRichardson2010}. In neuropsychology, deficits in high-level cognitive functions are regularly associated with impairments of particular regions of the brain that are regarded as modules, typically under the assumptions that these modules operate independently and, to variying degrees, that their effects are local \citep{Farah1994, shallice1988neuropsychology,bechtel2002connectionism}. These observations have a long tradition in cognitive science, where the nature of cognitive task-specific modules and the implementation details of compositional computation have been highly debated \citep[i.a.]{Fodor1975-FODTLO, Fodor1983-FODTMO, Fodor1988-FODCAC, TouretzkyHinton1988, Smolensky1988-SMOOTP-2}.\footnote{A review of this long and at times acrimonious debate is outside of the scope of this paper. For particularly interesting accounts we refer to \cite{bechtel2002connectionism} and \cite{Marcus2001-MARTAM-10}.} Among others, there are two notable early attempts to build modular systems with a cognitive focus: \cite{Miikkulainen1993} developed a modular system  for Natural Language Understanding with task-specific components, some of them with correlates to cognitive functions; and \cite{JACOBS1991219}'s modular networks, a modular and conditional computation model that performed task decomposition  based on the input and is, to some extent, a direct ancestor of routing networks.

Another important framework for compositional learning that is also influenced by the challenges of routing discussed in this paper is the options framework \citep{Options} for hierarchical reinforcement learning. In particular, challenges of routing are related to those experienced by end to end architectures for learning sequencing between policies over time with neural networks \citep{optioncritic} and extensions to hierarchical policies \citep{abstractoptions}. Option models, similar to routing networks, are known to experience "option collapse" where either only one option is actually used or all options take on the same meaning. This common difficulty has motivated recent research showing improvement in learned options by imposing information theoretic objectives on learning that incentivize increased option diversity and increased entropy in option selection \citep{VIC,florensa,hausman,DIAYAN,termcritic}. In contrast to routing networks, option models are also concerned with deciding the duration over which its decisions will last. In this paper we focus just on empirical analysis of the simpler case of routing at every time step so that we can directly address diversity of the chosen modules without introducing the additional complicating factor of sequencing selections over time.

% \mrXX{... any other kinds of models that we should mention?}

% \subsection{Applications}
% applications: anything that can be cast as "multi-task"; on-the fly assembly for new datasets; anything that's intrinsically compositional; 

% \subsection{Background and Related Work}
% Ideally we wouldn't need this section or can delay it until much later in the paper. 

%\input{challenges-generic.tex}
\section{Challenges to Compositional Computation}\label{sec:challenges}

We have experienced several challenges particular to training compositional architectures, including training stability, module collapse, and overfitting as well as difficulties with performance extrapolation and formalizing the setting.  Instability in training may occur because of a complex dynamic interplay between the router and module training. Module collapse may occur if module selection collapses to a policy which makes the same decision for all inputs.  Overfitting may be severely exacerbated in modular architectures because of their added flexibility to learn specialized modules for a very narrow subset of samples. Successfully extrapolating the performance of specific modules out over the course of training would potentially allow a more successful selection strategy, but achieving this is a difficult problem itself.  Finally, we lack a good formalization of popular training methodologies, unifying reinforcement learning for the module selection training with supervised training for the modules. There has been some progress on each of these problems (for example, module collapse was discussed in \citep{modularnets}). However, we believe this is the first time they have been collected and given a systematic treatment jointly.

\begin{wrapfigure}{r}{0.32\textwidth}
\vspace{-4mm}
  \centering
    \begin{tikzpicture}[xscale=0.8,>=latex]
\small

% border of the surface1
\path[draw,name path=border1] (0,0) to[out=-10,in=150] (3,-1);
% border of the surface1
\path[draw,name path=border2] (6,0.5) to[out=150,in=-10] (2.75,1.6);
% border of the surface1
\draw[draw,thick,name path=line1] (3,-1) -- (6,0.5);
% border of the surface1
\path[draw,name path=line2] (2.75,1.85) -- (0,0);
% draw the surface1
\shade[left color=gray!10,right color=gray!70] 
  (0,0) to[out=-10,in=150] (3,-1) -- 
  (6,0.5) to[out=150,in=-10] (2.75,1.85) -- cycle;

% gradients
\node at (3,0.5) {\textbullet};
\path[draw,name path=line2,thick,->] (3,0.5) to node [right] {$\nabla_\theta L_i$} (4.1,0.);
\path[draw,name path=line2,thick,->] (3,0.5) to node [below] {$\nabla_\theta L_k$} (1.7,0.7);
\node at(1.2, -1.2) {Interference};
\begin{scope}[shift={(0,-3.7)}]
% border of the surface1
\path[draw,name path=border1] (0,0) to[out=-10,in=150] (3,-1);
% border of the surface1
\path[draw,name path=border2] (6,0.5) to[out=150,in=-10] (2.75,1.6);
% border of the surface1
\draw[draw,thick,name path=line1] (3,-1) -- (6,0.5);
% border of the surface1
\path[draw,name path=line2] (2.75,1.85) -- (0,0);
% draw the surface1
\shade[left color=gray!10,right color=gray!70] 
  (0,0) to[out=-10,in=150] (3,-1) -- 
  (6,0.5) to[out=150,in=-10] (2.75,1.85) -- cycle;

% gradients
\node at (3,0.5) {\textbullet};
\path[draw,name path=line2,thick,->] (3,0.5) to node [right] {$\nabla_\theta L_i$} (4.1,0.);
\path[draw,name path=line2,thick,->] (3,0.5) to node [left] {$\nabla_\theta L_k$} (3.5,-.4);
\node at(1.2, -1.2) {Transfer};
\end{scope}

\end{tikzpicture}
  \captionof{figure}{Depiction of the effects of transfer and interference across examples $i$ and $k$ during learning. %The transfer-interference trade-off illustrated on the gradients of the losses for two different samples
  }
  \label{fig:transfer-interference}
  \vspace{-3mm}
\end{wrapfigure}
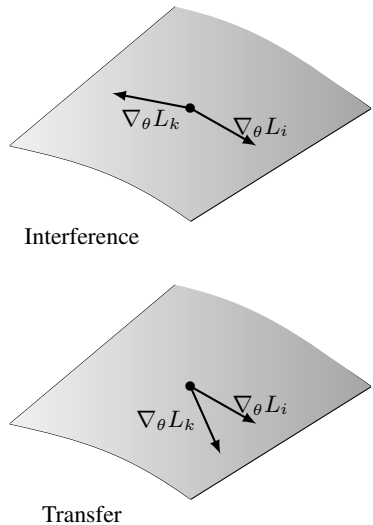

One may argue that core problems of a more general nature underlie these issues. In fact, we can see that learning to compose is difficult because we must simultaneously balance two challenges of learning: the transfer-interference trade-off \citep{MER}, and the exploration-exploitation dilemma \citep{suttonbarto}. 

The ``transfer-interference trade-off'' refers to the problem of choosing which parameters in the model are shared across different input samples or distributions.  When there is a large amount of sharing of model parameters during training we may see better performance as a consequence of \textit{transfer}, since each parameter is trained on more data. But it may also lead to worse performance if the training on different samples produces updates that `cancel' each other out, or cause \textit{interference}. Compositional architectures can offer an interesting balance between the two, as different modules may be active for different samples. 

The ``exploration-exploitation dilemma'', while mostly associated with reinforcement learning, exists for any kind of hard stochastic update. For modular learning, this means that the composer has to strike a balance between `exploiting', i.e. selecting and training already known modules which may perform well at some point in time during training, and `exploring' or selecting and training different modules that may not perform well yet, but which may become the globally optimal choice, given sufficient additional training. Unfortunately, high exploration increases the likelihood of both beneficial transfer and unwanted interference.  But low exploration may bias the model towards selections which limit its ability to achieve an optimal balance between transfer and interference. Striking the right balance between exploration and exploitation can therefore help with interference but is not sufficient to mitigate it entirely. This entanglement complicates the learning and means we cannot treat these tradeoffs in isolation. 

% The first two problems -- collapse and overfitting -- can be cast as special problems within the transfer-interference dilemma \citep{MER}, as collapse is a local transfer optimum, and overfitting in the context of compositional architectures, an optimum minimizing interference on the training data. We will now describe these problems, and try to analyze their causes, before illustrating their consequences empirically in Section \ref{sec:eval}.

% In our experiments we have found three main challenges in training routing networks: overcoming \textit{stability and convergence} issues during training; \textit{overfitting}; and \textit{diversification} of the modules.

\subsection{Training Stability}\label{sec:chall-stability}

One of the most challenging problems for routing networks is overcoming the early learning phase when both modules and the router have just been initialized. In this phase, the router needs to learn the value of the modules, while these are effectively randomly trained and have not yet taken on any real meaning, since pathways and gradient-flows are not yet stable.

\begin{wrapfigure}{r}{0.32\textwidth}
  \centering
  \vspace{-8mm}
    \subcaptionbox{A dataset with two linear modes}{
    \includegraphics[width=0.31\textwidth]{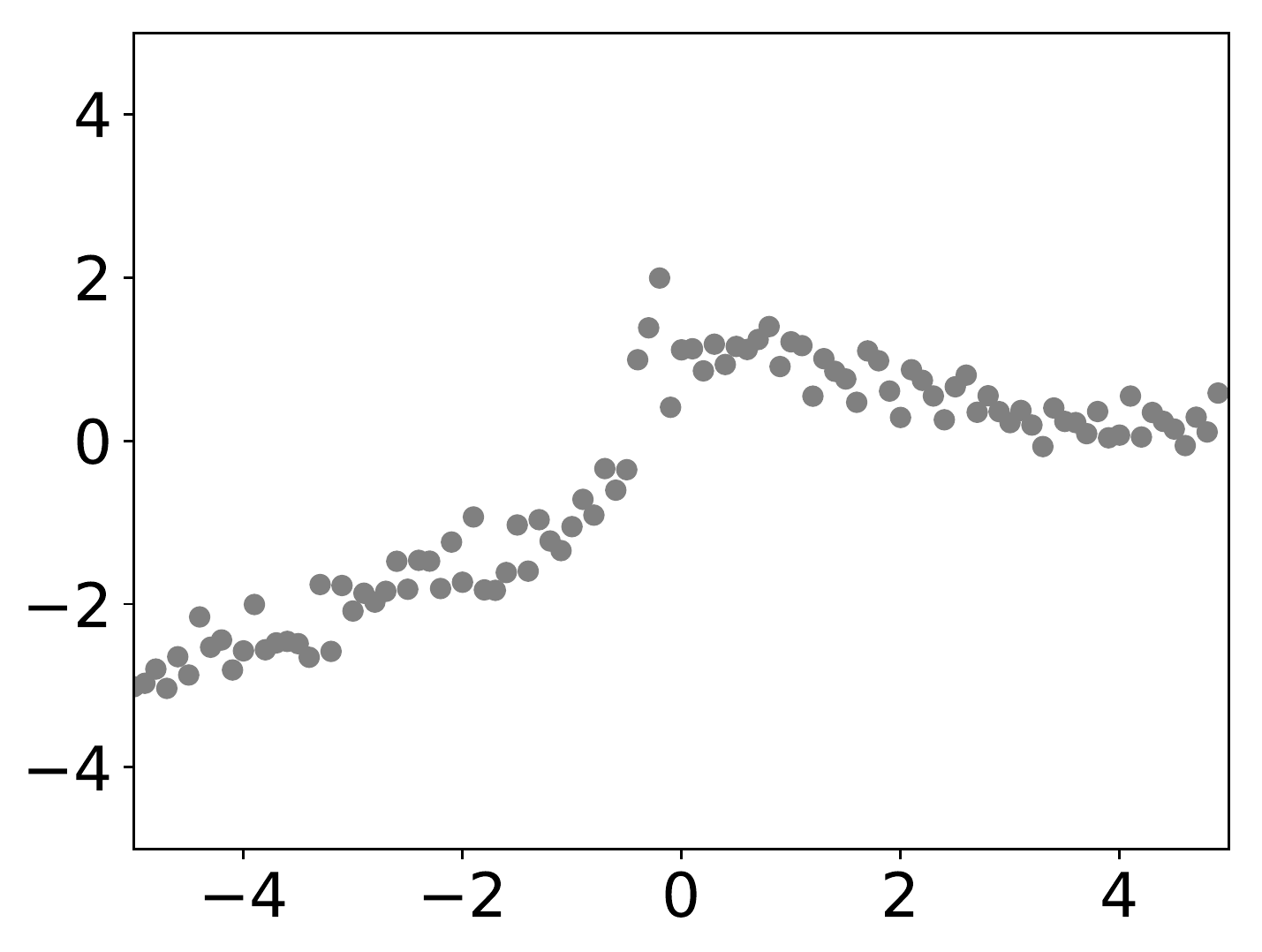}}
    
    \subcaptionbox{The desired routing solution; the model captures both modes}{
    \includegraphics[width=0.31\textwidth]{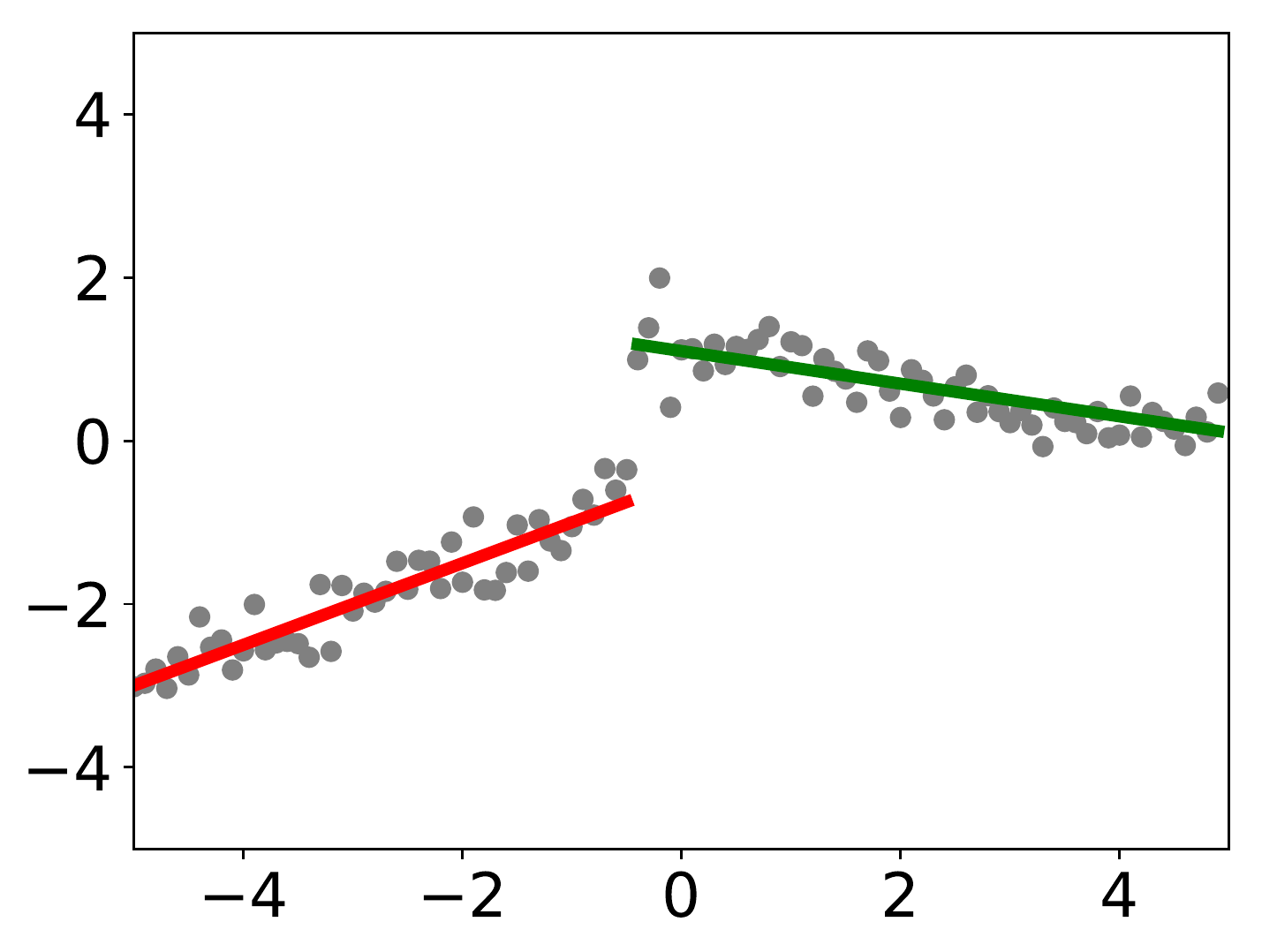}}
    
    \subcaptionbox{The routing modules at initialization}{
    \includegraphics[width=0.31\textwidth]{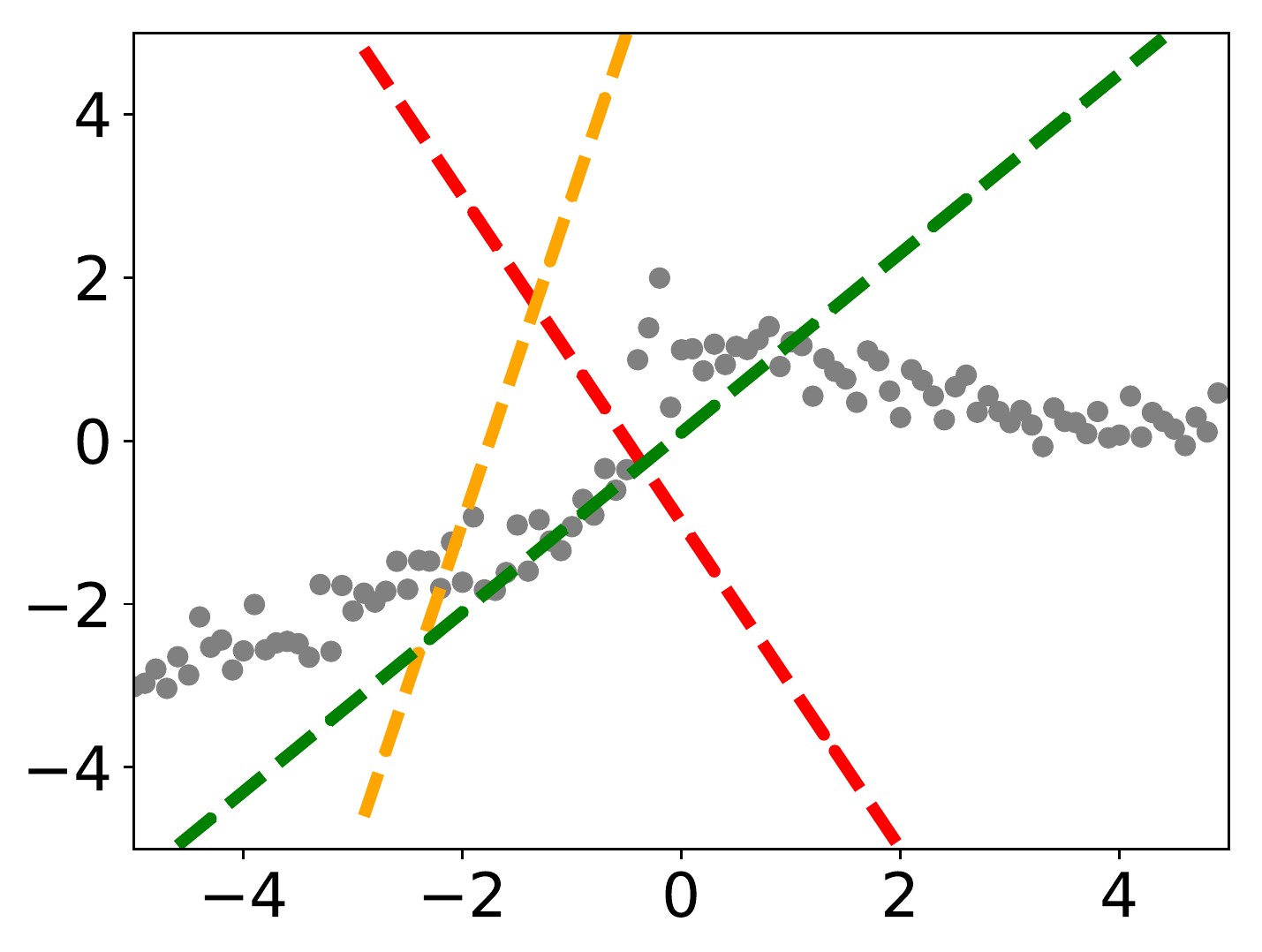}}
    
    \subcaptionbox{Module Collapse: the router reaches a local optimum using the green module only}{
    \includegraphics[width=0.31\textwidth]{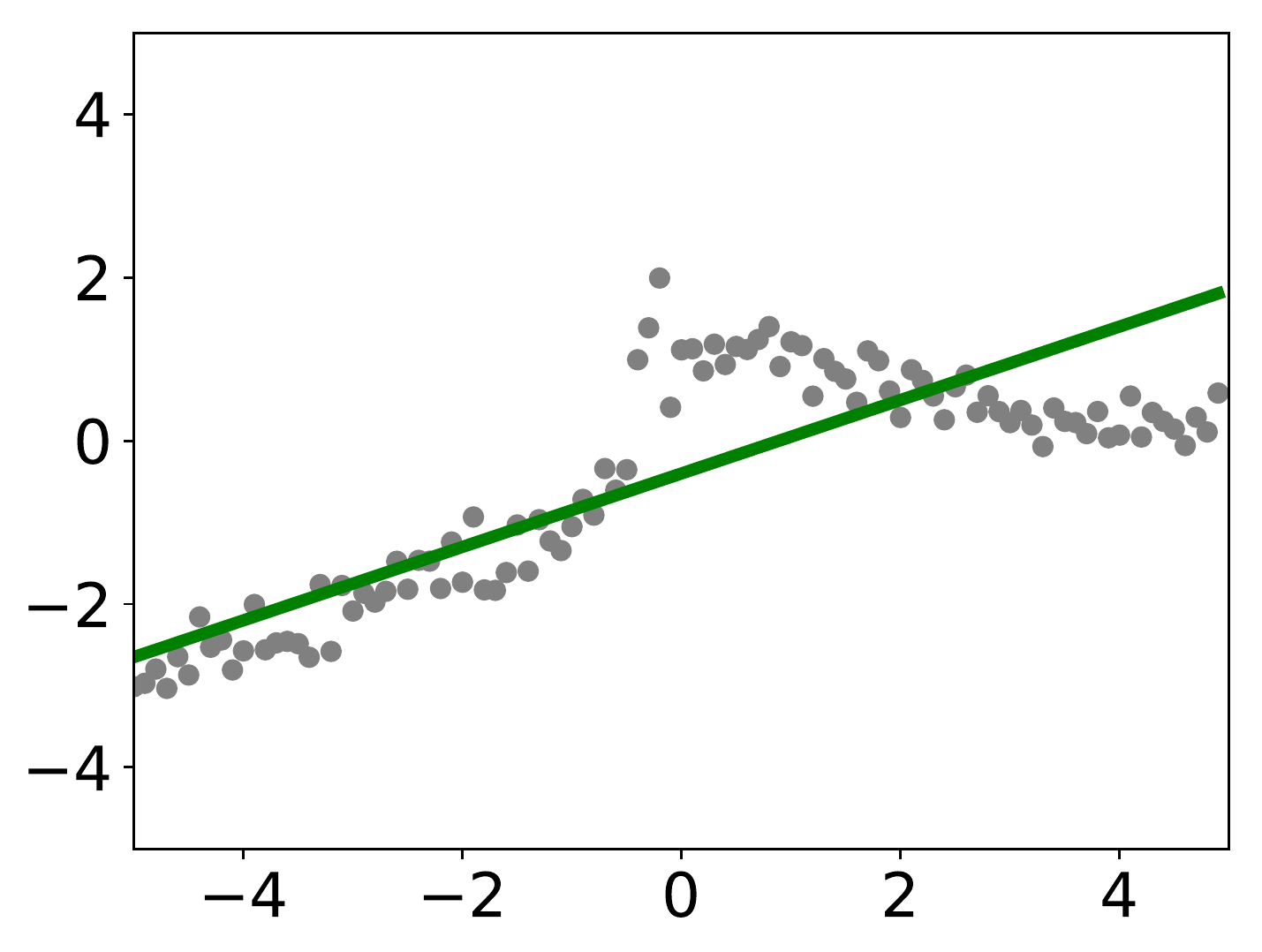}}
  \captionof{figure}{An example of how a 1-dimensional linear routing problem can collapse}
  \label{fig:chall-collapse}
  \vspace{-24mm}
\end{wrapfigure}
In our experience, this ``chicken-and-egg'' problem of stabilizing the early training of a routing network, before paths have had the opportunity to specialize, can result in the routing dynamics never stabilizing, leaving the network at effectively random performance. This problem roughly correlates with the complexity of the decision making problem. If the router only needs to make decisions on a very local distribution of inputs, then stability is less of a problem. If, however, the router has to consider very complex distributions, it consistently struggles.

A mitigating strategy can be to slow down the learning behavior of either the modules or the router, by either fixing their parameters for a short initial period, or by reducing their learning rate. In practice, we have found that simply reducing the learning rate of the router is an effective strategy to stabilize early training.
Another strategy that can be applied is curriculum learning \citep{curriculumlearning} as is done by \cite{crl}. This can be very effective, but can only be applied if the training data can naturally ordered by complexity, which is not the case in many important domains. Neither of these strategies offer a general solution to the problem.  They may work in some settings and fail in others.

\subsection{Module Collapse}\label{sec:chall-collapse}
Another common training difficulty is \textit{module collapse} \citep{modularnets}. This occurs when the router falls into a local optimum, choosing one or two modules exclusively. Module collapse occurs when the router overestimates the expected reward of a selection. This may happen due to random initialization or because one module has a higher expected return during early training. Either way, the module will be chosen more often by the router, and therefore receive more training. This in turn improves the module so that it will be selected yet more often and receive yet more training until the module is dominant and no others seem promising. 

As an example, consider Figure \ref{fig:chall-collapse}, which depicts a routing problem of one dimensional linear regression (since it is easily illustrated). In plot (a), we depict a  dataset with two noisy linear modes. Plot (b) shows the perfect, and desired, routing solution, where the routing model correctly assigns a routing path to each mode. Plot (c) depicts the regression curves produced by the available modules at initialization. Since the red and yellow approximations have been initialized producing too great an initial loss, the router will only update these during exploration phases, and will instead choose the green approximation. This may result in a suboptimal approximation  (depicted in plot (d)), which can form a local optimum that will trap the router without prolonged additional, poorly rewarded exploration.

This illustrates the view of collapse as a local optimum achieved by maximizing for early transfer.  When every selection is ``bad'', training a transformation even on generally less compatible samples, such as shown in Figure \ref{fig:chall-collapse}, will improve its representative power. Similarly, it can be seen as a local optimum in the exploration-exploitation dilemma, since the router correctly believes \textit{at a given time} that one module is better than others, and may thus not explore the other options sufficiently. %As we will see in section \ref{sec:eval}, higher exploration for the selection algorithm tend to be less stable, while lower exploration tends to collapse more.

Existing solutions to this problem include: adding regularization rewards that incentivize diversity \citep{routingnaacl}; separating the training of the router from the training of the modules with an EM-like approach that learns over mini-batches, explicitly grouping samples first \citep{modularnets}; and conditioning the router entirely on discrete meta-data associated with the sample, e.g., task labels \citep{routingnets,routingnaacl}. This has the additional benefit of reducing the dimensionality of the information on which the decisions are made, which, in turn, makes the routing problem easier.

\subsection{Overfitting}\label{sec:chall-overfitting}
\begin{figure}[th!]
    \centering
    \subcaptionbox{A dataset}{
    \includegraphics[width=0.31\textwidth]{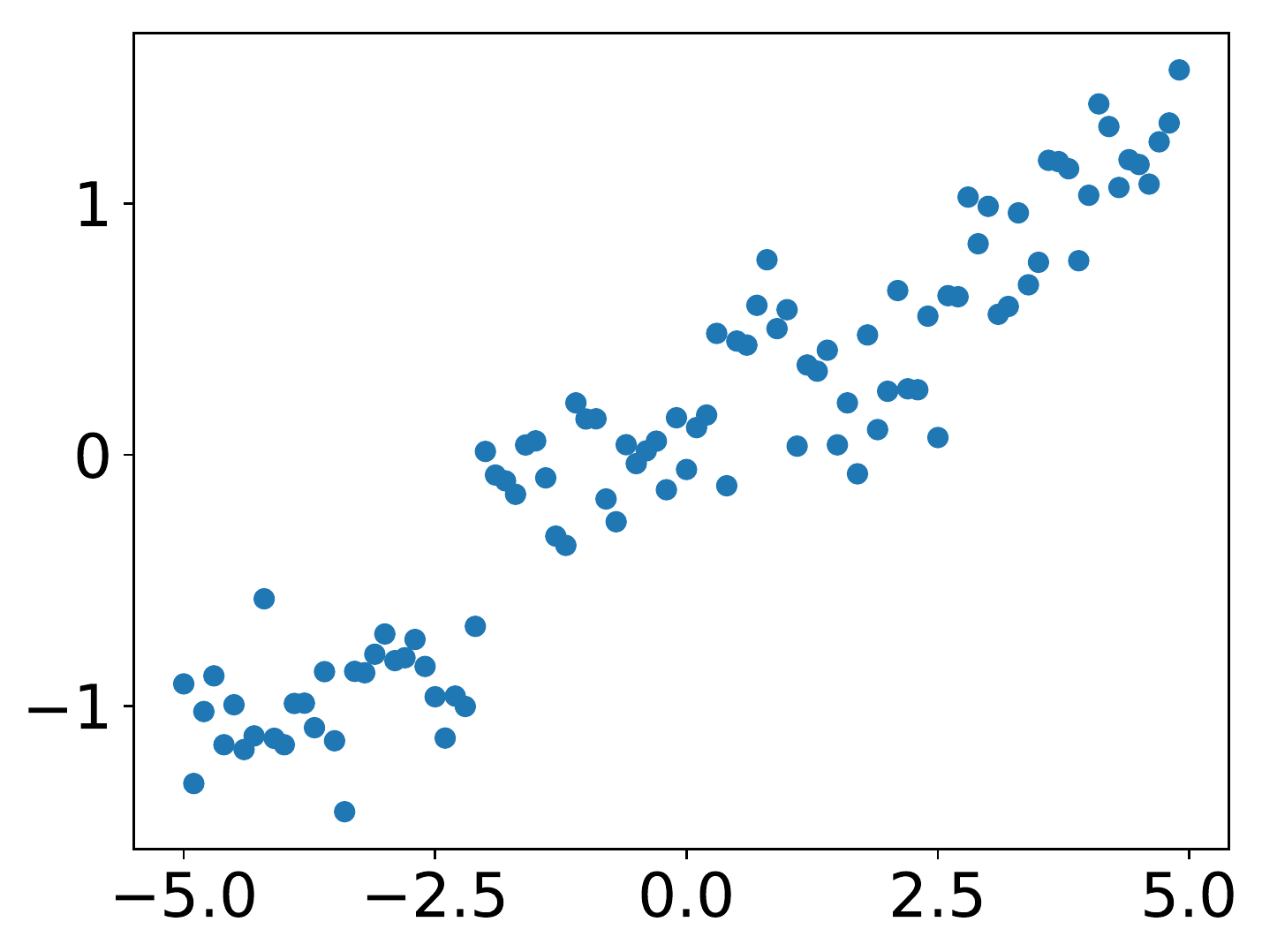}}
    \subcaptionbox{A linear approximation}{
    \includegraphics[width=0.31\textwidth]{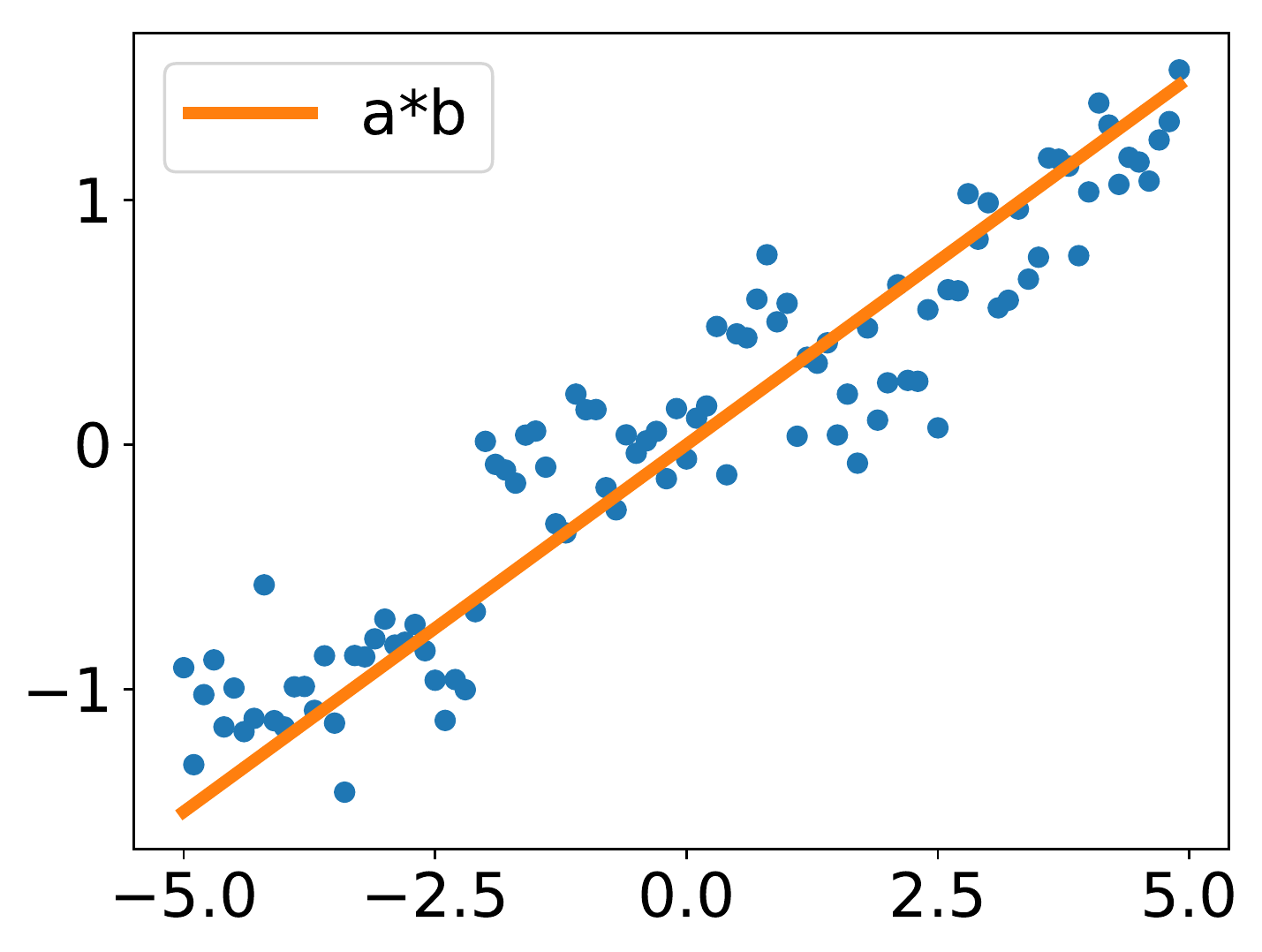}}
    \subcaptionbox{A routed approximation}{
    \includegraphics[width=0.31\textwidth]{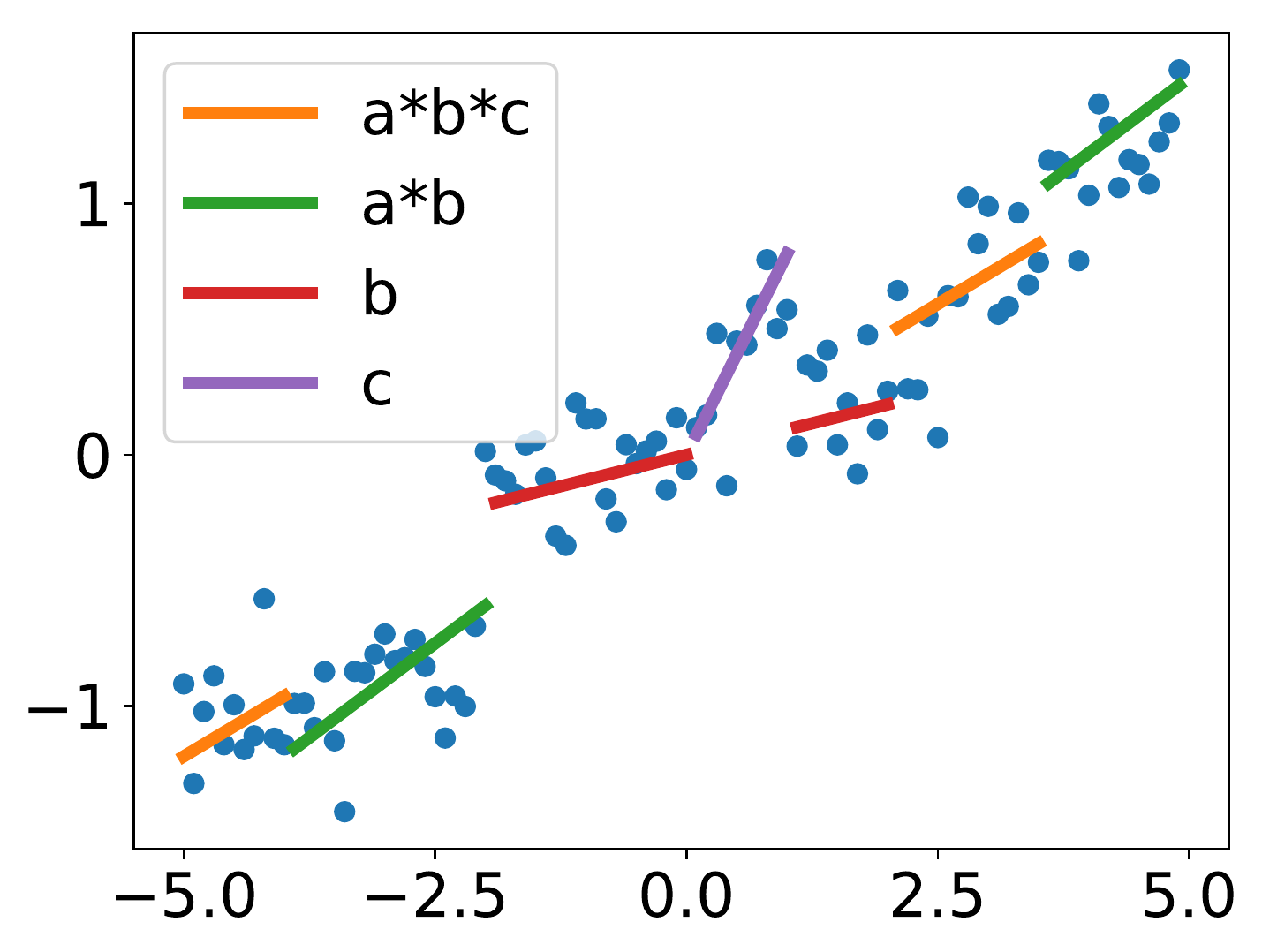}}
    \caption{Illustration of how a routed model may overfit. The learned parameter values of the three linear transformations are $a=3,b=0.1,c=0.8$.}
    \label{fig:overfit}
\end{figure}
Previous work \citep{modularnets,largeneuralnets} observes that models of conditional computation can overfit more than their ``base''-models, but do examine this issue in depth. We have also encountered this problem multiple times, and believe that it stems from the flexibility of routing models to compose highly local approximations. Consider the example in Figure \ref{fig:overfit}, where we again route a linear scalar approximator.

Suppose we have a training dataset consisting of observations of a noisy linear process, such as the one in Figure \ref{fig:overfit}(a). It can be clearly seen that a linear model gives good (and desired) approximations, as in Figure \ref{fig:overfit}(b), and that it does not overfit to the noise present in the data. However, imagine that we wanted to learn such an approximation with a routing model that consists of three parameterized functions, $f_1=a\cdot x, f_2=b\cdot x, f_3=c\cdot x$, which the router can combine to a maximum depth of three. If the router now routes purely based on input information, i.e., if the router can route each sample through a different route, then it may choose a different path for different subsets of the training data, resulting in different (local) approximations as illustrated in Figure \ref{fig:overfit}(c). For neural architectures in particular, this implies that routing breaks an implicit smoothness assumption which, arguably, allows them to generalize well in spite of high dimensionality (see \cite{overfitting} for a further discussion). Since the space of routes or paths grows exponentially with routing depth, deep routing networks can potentially learn highly local approximations with only few samples covered per approximation.

This perspective also explains why \cite{routingnets} and \cite{routingnaacl} do not experience overfitting: if the path of a sample is independent of its input value or its intermediate activations, the resulting paths and approximations cannot learn to match individual areas in activation-space only, but are constrained by the meta-information instead.

Within the transfer-interference trade-off, overfitting is a natural consequence of avoiding training interference. In models of uniform, i.e., non-compositional, computation, the training algorithm always enforces some amount of interference over the model's parameters. While often harmful, it can also steer training away from local suboptimal solutions that cause the model to overfit.

Since this problem is not well investigated for machine learning models, we can only speculate on possible solutions. The most obvious is to apply regularization of some kind that prevents the router from fitting samples to highly expressive local approximators. A very successful solution with a different set of problems is the `routing-by-meta-information' architecture of \citep{routingnets,routingnaacl}. Another possible solution might be module diversity: if only different kinds of approximators can be fit the ``locality'' of routing does not have the same effect. This would explain why \cite{ramachandran2018diversity} were able to achieve very high performance with a modular architecture even when learning with few examples, although we need to take their particular top-k architecture into account. Another uninvestigated area which seems promising for exploration is the tradeoff in capacity between the router and the modules. Is it better to have a `smart' router and `dumb' modules or vice versa? Either will probably have an effect on overfitting.

%Imagine yet another example -- having a number of selections in a routing network large enough to build separate models for each sample in a dataset. In this case, one perfect training solution would be if each sample got assigned networks with non-overlapping modules. However, the network would lose the ability to generalize, as the network can simply remember each sample perfectly.
%While such a scenario is unlikely, as it would require a very large number of modules, the same problem -- if weaker -- exists for any routing network.

%If the problem of Module Collapse is characterized as a local optimum of maximizing early transfer, overfitting can be seen as an optimum of minimizing late interence, as any sample pushing a module towards more general representations will be routed through a different module.

\subsection*{The Flexibility Dilemma for Modular Learning}
\begin{wrapfigure}{r}{0.4\textwidth}
\vspace{-4mm}
  \centering
    \input{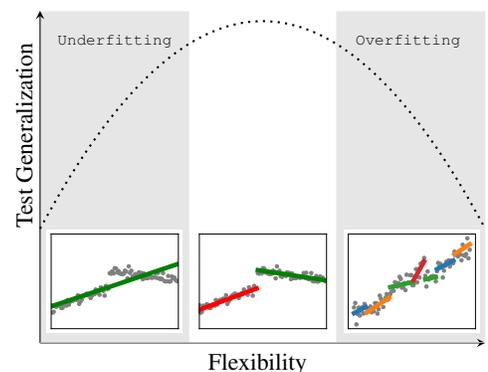}
  \captionof{figure}{The flexibility dilemma. Flexibility is the ability of a routing model to adapt to localized inputs. Low flexibility means few large clusters, and low flexibility means many small clusters. The extrema are only one cluster, i.e., collapsing and underfitting, and one cluster for each sample, i.e., overfitting.}
  \label{fig:flexibility-dilemma}
  \vspace{-3mm}
\end{wrapfigure}
Comparing the two challenges discussed this far, it becomes clear that they are not unrelated, but are an effect of too little -- or too much -- flexibility on the router's part, where flexibility is a routing network's ability to localize its routing decisions. If the router is not flexible enough to realize how truly different modes in a distribution should be treated differently, it will cause underfitting, oftentimes through collapse. If, on the other hand, the router is too flexible in mapping different inputs to different routing paths, it creates hyperlocal approximations that can overfit badly. 

This goes to the core of what routing does, and why it can be such a powerful machine learning model. Routing always has an enormous degree of expressivity, as we have a combinatorial number of implicitly defined models, in the form of paths. If the model finds a good `locality' of approximations, routing can allow a model to achieve truly impressive levels of generalization (compare \citep{crl}, and in particular the section on nested implicatives in \citep{routingnaacl}), as it can adapt to unseen samples by mapping these to known solutions in a highly variable way. However, on both sides of this `locality' of approximation lie the above mentioned pitfalls of collapse and overfitting.

While we illustrated and explained this problem on routing (neural models), it is by no means limited to it. Any modular approach that can treat different samples differently will be exposed to this dilemma between under- and overfitting, as it comes with any form of local approximation approach.

\subsection{Module Diversity}\label{sec:chall-div}

While much of the prior work has focused on the case where each module is of the same kind but with different parameters, routing modules that have different architectures has the potential to be even more powerful and sample efficient. This allows for a larger coverage of functional capabilities, thereby increasing the expressive power of a routing network. Interestingly, this is a specific instance of the ``Algorithm Selection Problem'' \citep{algorithmselection}. For learning problems, however, the problem is complicated by the different learning dynamics that different architectures might exhibit.

\begin{wrapfigure}{r}{0.34\textwidth}
  \centering
  \vspace{-5mm}
    \includegraphics[width=0.32\textwidth]{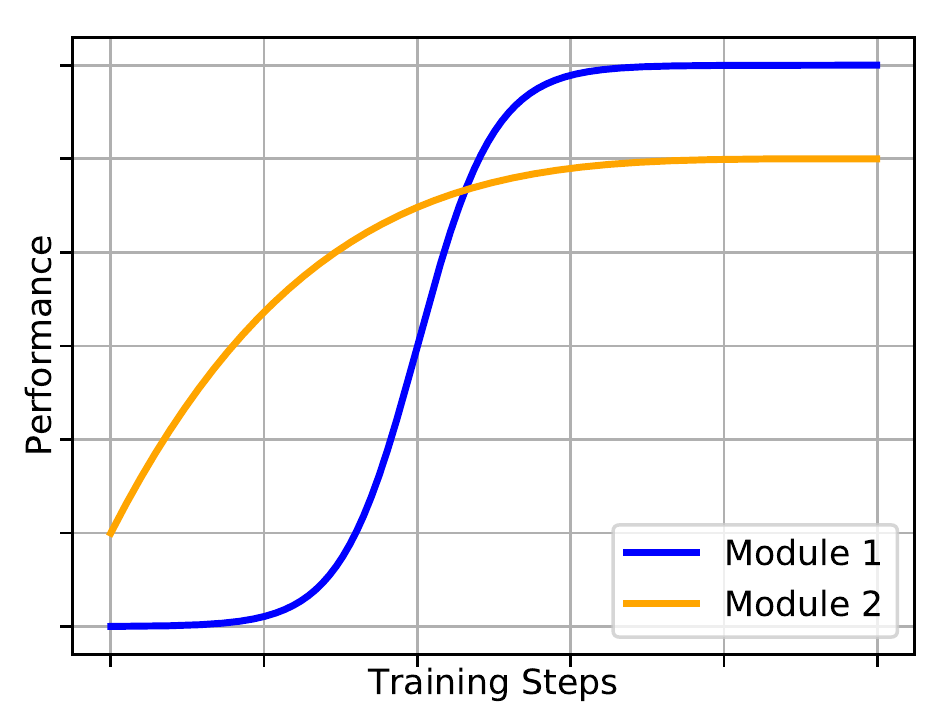}
  \captionof{figure}{Transformation Learning Dynamics}
  \label{fig:learning-dynamics}
     \vspace{-10mm}
\end{wrapfigure}

Consider Figure \ref{fig:learning-dynamics}. For any learning algorithm selection problem, deciding that Module 2 will yield superior performance requires the meta-learning algorithm to sufficiently explore -- and thus train -- both selections. However, for routing networks, exploration correlates with interference. 
This means that even if the router were to explore Module 1 sufficiently, the sampling noise of exploration could potentially interfere with the training procedure of selections taken before or after. While some progress on this problem has been made \citep{ramachandran2018diversity}, it focuses on selections with similar learning characteristics. The full problem, for selections with arbitrary learning characteristics, was only explored in an offline setting, where selections are trained until completion and not varied while training \citep{zophICLR,Liu_2018_ECCV}. Finding optimal selections online remains an open problem due to the difficulty of anticipating future rewards.

\subsection{A uniform formal framework} \label{sec:chall-formalization}
All of the previous challenges are complicated by the lack of a principled mathematical formalization for compositional learning. In particular, it is not clear how the training of the router relates to and interacts with the training of the modules. Although existing approaches have found very successful solutions, even without such a formalization, a principled framework may provide additional insight, convergence guarantees and directions for the development of better algorithms.

Specifically, it is not clear how training of the modules is impacted by the training of the routers, and vice versa. For how compositionality may interfere with the training of the modules, compare section \ref{sec:eval-optimizer}. For the other direction, consider the case of a reinforcement learning strategy for training the routers (arguably the most relevant update strategy). Reinforcement learning relies fundamentally on the assumption that the environment is governed by an underlying Markov Decision Process (MDP) \citep{suttonbarto}.  A naive though intuitive characterization of routing in this context might interpret individual modules as actions.  But an MDP requires a static, non-changing set of actions and here the modules are themselves trained and change over time. Although in practice this strategy has been shown to work in any case \citep{routingnets,crl,routingnaacl}, it lacks theoretical justification, and more principled approaches may yield superior performance.

This critique extends to the two special cases of MDPs introduced in the literature: Meta-MDPs and stochastic games. Meta-MDPs were introduced by \cite{metamdp} and applied to routing by \cite{crl}, and specifically model the computational steps required to make a decision in another underlying MDP. That is, their states are partial results, and their actions are computations. Once they terminate, the resulting computation produces an action in the underlying decision problem. This approach emphasizes that the routing problem consists of computation steps, and not steps taken in an environment. A stochastic game, related to routing by \cite{routingnets} is the multi-agent extension to MDPs and allows for an arbitrary number of agents to exist, which collaborate or compete in the same environment. As other agents interact with the environment, the environment becomes non-stationary from the perspective of any single agent. While this approach consequently models the non-stationarity of a routing problem better than a Meta-MDP, it still does not solve the more fundamental problem of possible interference between update strategies.

There are two scenarios which have more principled solutions to this criticism, although they, too, may benefit from a more targeted solution as they suffer from many of the same problems in practice. In the first, for neural networks, we use some form of reparameterized sampling function to update the composition (compare section \ref{sec:routing-repara}). This addresses the core of the problem since the reparameterized decision making algorithms explicitly use the same update strategy as the (neural) models. In the second scenario, the problem solved with a routing network is optimizing a policy for a given MDP. In this case, the routing network is the policy, and can consequently be modeled as a coagent network \citep{coagents}. A coagent network is a theoretical framework that, like Meta-MDPs, models the decision making in an underlying MDP but explicitly allows for non-stationary interactions between parts of the policy. These parts, called coagents, can update locally on stochastic information. For routing, we can model the policy solving the underlying MDP as a routing network with routing coagents and transformation coagents (i.e., the module parameters are part of the policy). Then the policy gradient theorems in \citep{thomas2011policy} for the acyclic and \citep{kostas2019reinforcement} for the cyclic case hold, as the modules will also be updated with REINFORCE. Extending this work on coagent networks to also cover non-reinforcement learning problems such as supervised learning as well might offer a principled way of modeling routing and similar RL-integrated approaches for conditional computation.

\section{Routing} \label{sec:routing}
\textit{Routing} describes a general framework of repeatedly selecting trainable function modules with a trainable \textit{router}. As such, arbitrary components of a machine learning model can be routed, as long as the model is composed by a sequential application of functions in a set of (compatible) functions. We will focus on routing networks, where the learnable function modules are neural networks and parts thereof. The router receives the state of the computation, the current activation -- the input $x$ at step $0$ -- and evaluates it to select the best function, which will produce a new activation, which then triggers a new computational loop. If the router decides that the result has been processed sufficiently, the last activation will be handed to other computations or interpreted as the output of the model. This result will then be used to jointly train the function modules and the router.
\begin{figure}[b!]
\centering
\begin{minipage}{0.48\textwidth}
\centering
    	\begin{tikzpicture}
		\scriptsize
		%\node (full_input) at (-2.0, 2.5) {input: $v$, $t$};
		% the input
		\node (input) at (0,0) {input: $x$};
		% the fm set: layer 1
		\node[draw,rectangle] (fm11) at (-1.2, 1) {$f_{1}$};
		\node[draw,rectangle] (fm12) at (-0.4,1) {$f_{2}$};
		\node[draw,rectangle] (fm13) at (0.4,1) {$f_{3}$};
		\node[draw,rectangle] (fm14) at (1.2,1) {$\bot$};

        \node[draw,rectangle] (fm21) at (-1.2,2.2) {$f_{1}$};
		\node[draw,rectangle] (fm22) at (-0.4,2.2) {$f_{2}$};
		\node[draw,rectangle] (fm23) at (0.4,2.2) {$f_{3}$};
		\node[draw,rectangle] (fm24) at (1.2,2.2) {$\bot$};

        \node[draw,rectangle] (fm31) at (-1.2,3.4) {$f_{1}$};
		\node[draw,rectangle] (fm32) at (-0.4,3.4) {$f_{2}$};
		\node[draw,rectangle] (fm33) at (0.4,3.4) {$f_{3}$};
		\node[draw,rectangle] (fm34) at (1.2,3.4) {$\bot$};
		
		% the routing selection
		% \node [below right, red] at (.5,.75) {below right};
	    % \draw (0,0) -- (3,1)
 % node[pos=0]{0} node[pos=0.5]{1/2} node[pos=0.9]{9/10};
		\node (a1) at (-2.5,0.5) {\tiny $\mathbf{router}(x, m)$};
		\node (a2) at (-2.5,1.7) {\tiny $\mathbf{router}(f_3(x),m)$};
		\node (a3) at (-2.5,2.9) {\tiny $\mathbf{router}(f_1(f_3(x)), m)$};
		\draw[-{latex[scale=1.5]},line width=0.2mm,dotted] (a1) [out=0, in=-140] to (fm13);
		\draw[-{latex[scale=1.5]},line width=0.2mm,dotted] (a2) [out=0, in=-140] to (fm22);
		\draw[-{latex[scale=1.5]},line width=0.2mm,dotted] (a3) [out=0, in=-140] to (fm34);
		% the information flow
		\draw[-{latex[scale=1.5]},line width=0.4mm] (input) [out=90, in=-90] to (fm13);
		\draw[-{latex[scale=1.5]},line width=0.4mm] (fm13) [out=90, in=-90] to (fm22);
		\draw[-{latex[scale=1.5]},line width=0.4mm] (fm22) [out=90, in=-90] to (fm34);
		% the output
		\node (output) at (0, 4.4) {$\hat{y}=f_{1}(f_{3}(v,t))$};
		\draw[-{latex[scale=1.5]},line width=0.4mm] (fm34) [out=90, in=-90] to (output);
	\end{tikzpicture}\vspace{-2mm}
    \captionof{figure}{Routing (forward) Example}
    \label{fig:forward}
\end{minipage}%
\begin{minipage}{0.48\textwidth}
\centering
    \input{drawings/alg_forward.tex}
    \captionof{alg}{Routing Forward}
    \label{alg:forward}
\end{minipage}%
\end{figure}

As routing relies on \textit{hard} decisions to select the modules (as opposed to `soft' decisions, where several modules are activated and combined in different ways), training algorithms for the router are limited. While other approaches are conceivable (in particular genetic algorithms and other stochastic training techniques), we focus on Reinforcement Learning (RL) and Stochastic Reparameterization in this work.\\

\subsection{Reinforcement Learning}
Reinforcement learning describes a general machine learning paradigm in which an agent that makes a series of hard decisions is trained by providing simple scalar feedback, the reward. Applied to routing, the general outline is that the output of the model $\hat{y}$ has to be translated to a reward $r$ which is then used to improve the router's policy $\pi$.

\subsubsection{Routing MDPs} \label{sec:routing-mdp}
Markov Decision Processes (MDPs) are a common model for formally describing sequential decision making problems. While we have argued in section \ref{sec:chall-formalization} that existing attempts to model compositional computation are problematic, we will still adopt a common formalism for now so that we can discuss some relevant concepts. 

Given a routing MDP $M=\langle S,A,R,P,\gamma \rangle$, a set of applicable transformations $F$, the space of samples $X$, the space of transformations $H$ (i.e., the space of applications of members of $F$ to $X$), and the space of possibly available meta-information $M$, we can define:
\begin{align*}
    \text{the states } S &= (X\cup H) \times M \\
    \text{the actions } A &= F \cup \{\bot\}, \text{where $\bot$ is a termination action}\\
    \text{the transition probability } P(s, a, s') &= 
        \begin{cases}
            1, &\text{if } s = h, a=f, s'=f(h), h\in S, f\in F\\
            0 &\text{otherwise}
        \end{cases}\\
    \text{the discount factor } \gamma &= 1
\end{align*}
The reward function $R$ can freely be designed by the model designer and is not defined any further for now. Solving an MDP generally consists of finding a policy $\pi^*(\theta)$ (parameterized by parameters $\theta$) that maximizes the objective, or score function, $J(\pi(\theta)) := E(\sum_t r_t |\pi(\theta))$.

\subsubsection{Reward Design}\label{sec:rewards}
One of the most important questions when modeling a problem as an MDP is the question of how to design the reward function $R$. For a routing network there are two types of rewards to be considered. The first is the final reward that reflects the models performance of solving the main problem, i.e. of predicting a sample-label. The second are different forms of regularization rewards that can either be computed for entire trajectories or as immediate responses to individual actions.

\paragraph{Final Rewards}
The final reward $r_f$ models the overall performance of the model for a sequence of routing decisions. For classification problems, the outcome is binary -- correct or false -- and an obvious choice is a simple binary reward of $\pm 1$ corresponding to the prediction agreement with the target label. As the objective for the training of the modules is a loss function $\mathcal{L}_{m}(y, \hat{y})$, a different reward design for the same objective can be derived as the negative of the module loss: $r_f= - \mathcal{L}_{m}(y, \hat{y})$. The reward has to be the negative of the module loss as -- by convention -- we minimize model losses but maximize RL rewards. In addition, this allows for a natural application of routing networks to regression problems, as these cannot be translated into a simple binary $\pm 1$ reward. While the second reward design seems like the better choice, as it both richer in information and a more principled meta-learning objective that coincides with the main model's objective, we found that it does not necessarily perform better in practice depending on the problem of focus.

\paragraph{Regularization Rewards}
As the reward signal is sparse for complex compositions, it can also be quite helpful to incorporate additional rewards that can act as an intrinsic reward or regularizer of the choices made by the router. One option is to model problem-specific information in this reward -- e.g. to reflect the cost of computation of choosing a specific function, or to incorporate domain-knowledge, such as intuitions as to which model should be preferable. The second option, as employed in \cite{routingnets}, is to use this regularizer to incentivize transfer between different kinds of samples. There, the reward can be defined to correlate with how often a particular function $a$ is chosen within some window $w$, $C(a) = \frac{\# a}{\sum_w samples}$ -- thereby motivating the router to choose, and thereby share, that function for a wider set of samples. This reward is defined as $R(a)=\frac{\alpha}{t} C(a)$, with a ratio $\alpha$, normalized by the trajectory length $t$ -- such that even for long trajectories this reward may be limited to be smaller than the final reward $r_f$. \cite{routingnets} only investigate values of $\alpha \in [0, 1]$. However, this can lead to a lack of variety in decision making, as the selection collapses -- as discussed in section \ref{sec:chall-div} in detail. To compensate, \cite{routingnaacl} investigate a different set of values for $\alpha$, $\alpha \in [-1, 0]$, with the goal of incentivizing diversity of decision making.

\subsubsection{Algorithms}\label{sec:routing-algs}
Several RL algorithms have been successfully used to train routing policies. \citet{routingnets} compare a variety of algorithms, but settle for a multi-agent algorithm, fitting their multi-agent approach to multi-task learning. \cite{routingnaacl} find that plain Q-Learning performs the best for their domain. %\cite{rec-routing} also compare multiple algorithms, and come to the conclusion that plain QLearning gives state of the art results;
\cite{modularnets} extend a REINFORCE based algorithm with an alternating, EM-like training for modules and router that also optimizes for module diversity and does lookaheads. Meanwhile, \cite{crl} use Proximal Policy Optimization. \\

While this might seem to suggest that there is not necessarily a ``best'' reinforcement learning algorithm, there are some constraints to be considered. The first is that the nonstationarity of the parameters of $F$ prevents the use of replay buffers, as these would be invalidated after each transformation training step. Additionally, we argue that RL algorithms that can accommodate the change in the modules' respective values over training will perform the best. 

Consider the following example:
Given a set of module choices $M$, and a value estimator for a set of states $S$, $\hat{v}(s,m), m \in M, s\in S$. Given some training, the module parameters change, and thereby the expected value for each of the modules given a specific state $s_k$; using some value-based RL algorithm, we also update $\hat{v}$. As, for routing, we generally want to exploit more than explore, the module estimates for state $s_k$ are of different quality, as some modules have been sampled less from state $s_k$ than others. However, these modules may have been trained a lot from a different state $s_l$. Given that there can be transfer from state $s_k$ to state $s_l$, this implies that the value of the less sampled modules from state $s_k$ will probably be underestimated, as the estimator for $\hat{v}$ does not consider the additional training from state $s_l$. Consequently, this may ``lock'' the router in its selection (see also section \ref{sec:chall-collapse} for a discussion of this problem) early on.

A possible solution to this problem -- at least for RL algorithms with a value-based component -- might be to use the Advantage instead of (Q-)values: $A_\pi(s,a) = Q_\pi(s,a)-V_\pi(s)$, as the value component compensates for the inductive bias built up by training all transformations. Put differently, the increase in return for the on-policy average return, even when effectively only training one transformation, is not captured by the Q-component, but instead by the value component of the router. This may keep the difference in Q-value between different transformations low enough to lock the router permanently into a particular transformation, thereby helping to overcome selection collapse.

\subsection{Stochastic Reparameterization}\label{sec:routing-repara}
When it comes to hard decision making in differentiable architectures, stochastic reparameterization was recently introduced as an important alternative to Reinforcement Learning. While earlier work on stochastic reparameterization focused on low-dimensional stochastic distributions, \cite{maddison2016concrete} and \cite{jang2016categorical} discovered that the concrete -- or Gumbel Softmax -- distribution allows to reparameterize $k$ dimensional distributions (we will refer to the corresponding decision making algorithm as ``Gumbel''). Considering that these can be implemented as differentiable `layers' in deep architectures, they appear to be the better choice for a routing networks. \cite{maddison2016concrete} already note that the concrete distribution and its reparameterization do not yield an unbiased estimate of the gradient. The REINFORCE estimator \citep{williams_simple_1992} for the gradients of the  score function, on the other hand, is known to be unbiased (but has very high variance). Therefore two more techniques were introduced to estimate the gradient of discrete random variables. The first, REBAR \citep{rebar-estimator} compensates for REINFORCE's variance by introducing a control variate based on the concrete distribution. The second, RELAX \citep{relax-estimator}, generalizes this for non-discrete and non-differentiable stochastic functions, and allows a neural network to model the control variate.
%However, \cite{rec-routing} as the first to investigate the Gumbel Softmax trick for routing, found that it generally yields inferior results when compared to RL algorithms.

\begin{wrapfigure}{r}{0.34\textwidth}
    \centering
    \vspace{-5mm}
    \includegraphics[width=0.32\textwidth]{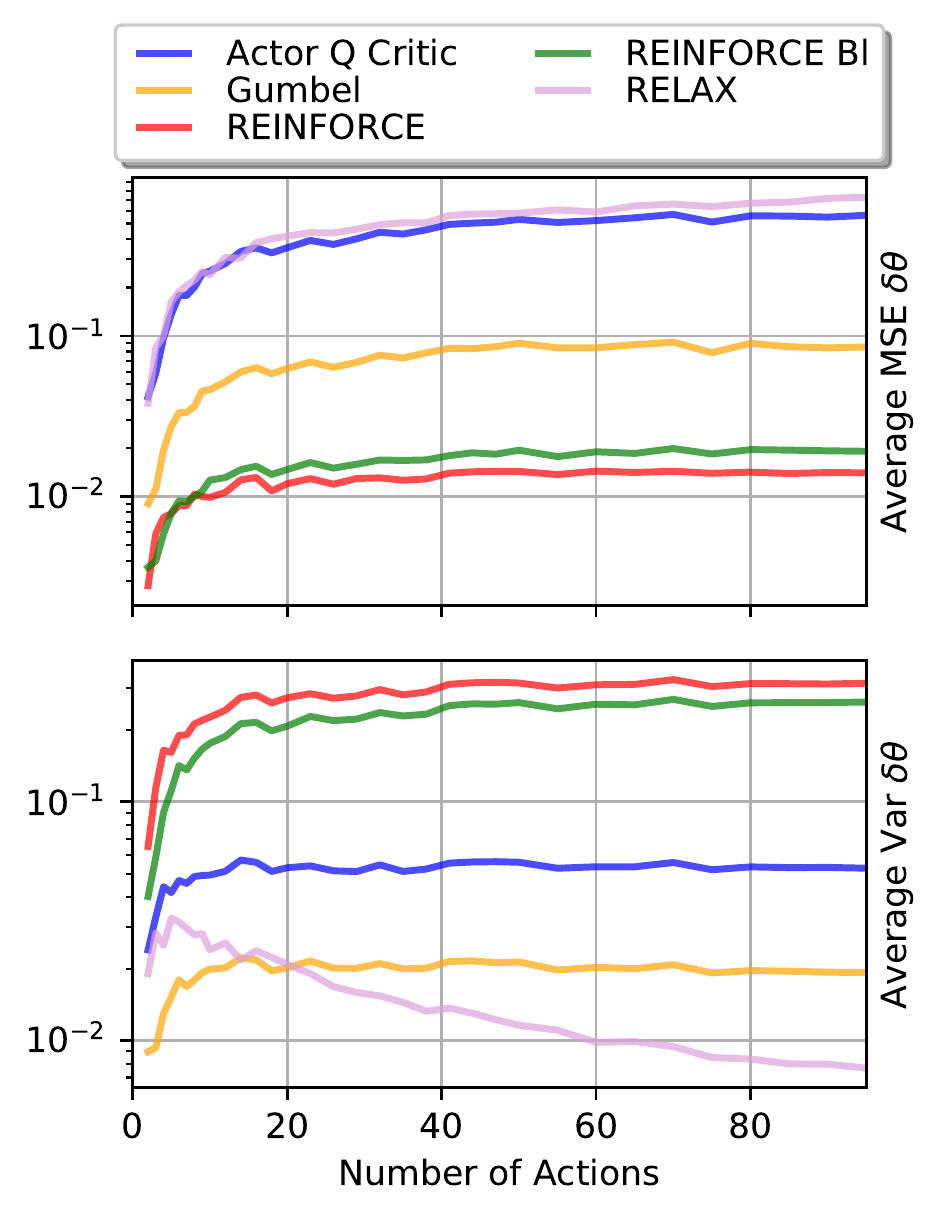}
    \caption{Gradient statistics. The REINFORCE Baseline is computed as $\hat{r} = (1-\alpha) \hat{r} + \alpha r$ with $\alpha=0.1$. The temperature parameter for all approaches is 0.5, which appears to be a common choice in the literature.}
    \label{fig:gradient-comparison}
    \vspace{-3mm}
\end{wrapfigure}
To better understand the relative performance of reparameterized approaches when compared to a reinforcement learning algorithm, we performed the following analysis: Given a known distribution as a policy $\pi$ parameterized by $\theta$ over a known reward function $r$, we do not sample, but instead analytically compute the value of the score-function $J(\theta) = E(r|\pi(\theta))=r\cdot \pi(\theta)$, where $\cdot$ denotes the inner product. We can continue by analytically computing the gradient for all parameters $\theta$. We know that these gradients are the ground truth that policy gradient and reparameterization approaches approximate only. Relying on this ground-truth information, we can then approximate the same gradients $\frac{\delta J}{\delta \theta}$ using REINFORCE, Gumbel and RELAX, and finally compute the average difference between the ground truth values and the approximations.

For Figure \ref{fig:gradient-comparison}, we use this approach to compute both the average gradient differences and the average variances over all parameters. More specifically, we initialize a random reward function (where each reward is uniformly sampled from $[0,1]$), of dimensionality $k$. We also randomly initialize a policy of dimensionality $k$, parameterized by $\theta$. After computing the ground truth gradient for a given pair of reward function and policy, we use the same reward function and policy to sample the gradients for $\theta$ $22*k$ times (so that on average each action will be sampled equally often, even with increasing dimensionality). For each $k$, we compute and approximate the gradients for 22 reward and 22 policy values, totalling over $10000*k$ datapoints for each point in Figure \ref{fig:gradient-comparison}. 

Figure \ref{fig:gradient-comparison} shows the mean squared error between the ground-truth gradient values and the sampled values using reparameterization and different RL policy gradient algorithms on top, and the corresponding variances on the bottom. While the gradients estimated by the Gumbel softmax trick are of \textit{much} lower variance, they are also biased, with an average MSE larger than the MSE of the Policy Gradient algorithms. This is consistent with the analysis in \citep{maddison2016concrete}. As for RELAX, we found very low variance for high dimensionalities, but -- interestingly enough -- a much higher MSE than REINFORCE and even than Gumbel, suggesting a higher bias. However, this may stem from the problem that RELAX relies on a trained surrogate network, which is difficult to train for these single-sample experiments.  While we try to accommodate for this requirement by updating this network over multiple samples first, this might not suffice to fully initialize RELAX. However, as we will show in Section \ref{sec:eval}, where RELAX and its surrogate network are trained over millions of iterations, it is still very difficult to draw clear conclusions about how reparameterization techniques relate to REINFORCE based approaches in practice.

\subsection{Simultaneously Selecting Multiple Function Blocks}

While in principle a routing network that only chooses a single function block at each step is capable of learning any function, it could be useful for sample efficiency to combine the output of multiple parallel routing paths. One popular approach for implementing this idea is to use a sparse version of the mixtures of experts architecture as in \citep{largeneuralnets,ramachandran2018diversity}. This approach leverages a typical mixtures of experts gating network that combines the output of multiple experts using a weighted averaging procedure. However, in general we would like to have an even more expressive mechanism of combining the output from multiple parallel routing paths that can represent more complex relationships between the output of each path. As sparse mixtures of experts architectures only use the top K experts at each step, there is also an inherent exploitation and exploration dilemma that impacts the system despite the fact that is not modeled explicitly. As a result, sparse mixtures of experts architectures must include special procedures for adding noise to the system and load balancing to achieve strong performance \citep{largeneuralnets}. In order to arrive at a general purpose solution, it makes more sense to model the selection of a subset of experts as an explicit reinforcement learning problem. For example, in \citep{EBengio} each "expert" is gated by a Boolean routing decision that is modeled as a reinforcement learning problem. 

\subsection{Training}
\begin{figure}[ht!]
\centering
\begin{minipage}{0.42\textwidth}
\centering
    		\begin{tikzpicture}
		\scriptsize
		%\node (full_input) at (-2.0, 2.5) {input: $v$, $t$};
		% the input
		\node (input) at (0,0) {input: $x$};
		\node (output) at (0, 4.4) {$\hat{y}=f_{1}(f_{3}(v,t))$};
		% the fm set: layer 1
		\node[draw,rectangle] (fm13) at (0,1) {$f_{3}$};
		\node[draw,rectangle] (fm22) at (0,2.2) {$f_{2}$};
		\node[draw,rectangle] (fm34) at (0,3.4) {$\bot$};
		
		\node[draw,rectangle] (tm13) at (1,1) {$\theta(f_{3})$};
		\node[draw,rectangle] (tm22) at (1,2.2) {$\theta(f_{2})$};
		
		% the routing selection
		% \node [below right, red] at (.5,.75) {below right};
	    % \draw (0,0) -- (3,1)
 % node[pos=0]{0} node[pos=0.5]{1/2} node[pos=0.9]{9/10};
		\node (a1) at (-1.2,0.5) {\tiny $a_1$};
		\node (a2) at (-1.2,1.7) {\tiny $a_2$};
		\node (a3) at (-1.2,2.9) {\tiny $a_3$};
		\draw[-{latex[scale=1.5]},line width=0.2mm,dotted] (a1) [out=0, in=-140] to (fm13);
		\draw[-{latex[scale=1.5]},line width=0.2mm,dotted] (a2) [out=0, in=-140] to (fm22);
		\draw[-{latex[scale=1.5]},line width=0.2mm,dotted] (a3) [out=0, in=-140] to (fm34);
		% the information flow
		\draw[-{latex[scale=1.5]},line width=0.4mm] (input) [out=90, in=-90] to (fm13);
		\draw[-{latex[scale=1.5]},line width=0.4mm] (fm13) [out=90, in=-90] to (fm22);
		\draw[-{latex[scale=1.5]},line width=0.4mm] (fm22) [out=90, in=-90] to (fm34);
		% the output
		\draw[-{latex[scale=1.5]},line width=0.4mm] (fm34) [out=90, in=-90] to (output);
		
		\node (loss) at (2, 4.4) {$\mathcal{L}(\hat{y}, y)$};
		\node (finalreward) at (-2, 4.4) {$r_f(\hat{y}, y)$};
		
		\draw[-{latex[scale=1.5]},line width=0.4mm,dashed] (finalreward) [out=-90, in=160] to node [left] {} (a3);
		\draw[-{latex[scale=1.5]},line width=0.4mm,dashed] (a3) [out=200, in=160] to node [left] {\tiny$+r(a_2)$} (a2);
		\draw[-{latex[scale=1.5]},line width=0.4mm,dashed] (a2) [out=200, in=160] to node [left] {\tiny$+r(a_1)$} (a1);
		
		\draw[-{latex[scale=1.5]},line width=0.4mm,dashed] (output) [out=0, in=180] to node [right] {} (loss);
		\draw[-{latex[scale=1.5]},line width=0.4mm,dashed] (output) [out=180, in=0] to node [right] {} (finalreward);
		
		\draw[-{latex[scale=1.5]},line width=0.4mm,dashed] (loss) [out=-90, in=20] to node [right] {\tiny$\frac{\partial \mathcal{L}}{\partial f_{2}}$} (tm22);
		\draw[-{latex[scale=1.5]},line width=0.4mm,dashed] (tm22) [out=-20, in=20] to node [right] {\tiny$\frac{\partial \mathcal{L}}{\partial f_{3}}$} (tm13);
		
		\draw [color=white,fill=white, opacity=0.5]
       (-1.4,0.2) -- (-1.4,4.2) -- (1.46,4.2) -- (1.46,0.2) -- cycle;
	\end{tikzpicture}\vspace{-2mm}
    \captionof{figure}{Routing (backward) Example}
    \label{fig:backward}
\end{minipage}%
\begin{minipage}{0.56\textwidth}
\centering
    \input{drawings/alg_backward.tex}
    \captionof{alg}{Backward step}
    \label{alg:backward}
\end{minipage}%
\end{figure}
Training a routing network is illustrated in Figure \ref{fig:backward} and the corresponding algorithm is Algorithm \ref{alg:backward}. The core idea is that the training of the router and of the modules happens simultaneously, after completing an episode, i.e., the forward pass. After the network has been assembled in the forward pass, the output of the network is translated into a loss for the module parameters and a final reward for a reinforcement learner. That reward, in combination with any accumulated per-action rewards, can be used to define a training loss for the router, either a Bellman Error, a negative log probability loss or some other loss function used to train a decision maker. The resulting losses can be added, and then backpropagated along the decision making parameters to define a gradient for each parameter in the model. Backpropagating along the sum of the losses allows for higher mini-batch parallelization in the update process of the network. %(a parameterized stochastic learner will be exclusively trained on the module loss) -- what does this mean? 

\paragraph{SGD}
Finally, a gradient descent style algorithm can use these gradients to derive a new set of parameter values. However, it is worth noticing that routing networks can be highly sensitive to the choice of optimization algorithm. We will discuss this further in Section \ref{sec:eval}.

\paragraph{Exploratory Actions}
Another useful change to the training procedure is to limit the updates of the transformations if they were chosen by an exploratory action. The intuition is that we do not want the network to add interference to the training of modules if they were just evaluated and found not fitting to a particular sample. As, however, any exploratory action has an effect on any return from the entire trajectory, we simply squash the optimization step size \textit{of the modules} for the particular trajectory using the following formula:
\begin{align}
    \alpha &= \left(1 - \frac{\#exploratory actions}{trajectory length} \right)^{\kappa}
\end{align}
where $\alpha$ is factor to the learning rate, and $\kappa$ is a hyperparameter. For $\kappa=0$, exploratory actions will be treated no differently, while for a very large $\kappa$, the transformations are effectively not trained on the entire trajectory if even only one action was taken non-greedily.

\paragraph{Splitting training data} One of the challenges of training a routing network is overfitting (see Section \ref{sec:chall-overfitting} for a thorough explanation). To prevent this, we investigate splitting the training data into a part for training the transformations, and a part to train the router.

\subsection{Architectures}
\paragraph{Model Architectures}
In general, any machine learning model can be routed. However, for non-layered architectures, routing collapses to the better investigated model selection problem. For layered architectures, any single layer can be routed by creating parallel copies of the respective layer, each with different (hyper)parameters. Existing routing architectures have included routing several fully connected layers with identical hyperparameters but different parameters \citep{routingnets,modularnets,crl,routingnaacl}, routing entire convolutional networks \citep{modularnets}, routing the hidden to hidden transformation of recurrent neural architectures \citep{modularnets,routingnaacl}, routing the input to hidden transformation \citep{routingnaacl}; and routing word representations \citep{routingnaacl}. However, only \cite{ramachandran2018diversity} have studied the effect of routing among modules that have different architectures.

\paragraph{Router Architectures}\label{sec:routing-architectures}
\begin{figure*}[hb!]
\centering
    	\begin{tikzpicture}
		\scriptsize
		%\node (full_input) at (-2.0, 2.5) {input: $v$, $t$};
		% the input
		\node (input) at (-.4,0) {};
		% the fm set: layer 1
		\node[draw,rectangle] (fm11) at (-1.2, 1) {$t_{1}$};
		\node[draw,rectangle] (fm12) at (-0.4,1) {$t_{2}$};
		\node[draw,rectangle] (fm13) at (0.4,1) {$t_{3}$};

        \node[draw,rectangle] (fm21) at (-1.2,2.2) {$t_{1}$};
		\node[draw,rectangle] (fm22) at (-0.4,2.2) {$t_{2}$};
		\node[draw,rectangle] (fm23) at (0.4,2.2) {$t_{3}$};

        \node[draw,rectangle] (fm31) at (-1.2,3.4) {$t_{1}$};
		\node[draw,rectangle] (fm32) at (-0.4,3.4) {$t_{2}$};
		\node[draw,rectangle] (fm33) at (0.4,3.4) {$t_{3}$};
		
		\node (router) at (-2.8, 2.2) {\large $\pi(h)$};
	    \draw (-3.5,1.1) -- (-2,1.1) -- (-2,3.2) -- (-3.5,3.2) -- (-3.5,1.1);
	    \node (routerbox) at (-2.9, 1.3) {\texttt{router}};
		\draw[-{latex[scale=1.5]},line width=0.2mm,dotted] (router) [out=-60, in=-140] to (fm11);
		\draw[-{latex[scale=1.5]},line width=0.2mm,dotted] (router) [out=-20, in=-140] to (fm22);
		\draw[-{latex[scale=1.5]},line width=0.2mm,dotted] (router) [out=40, in=-140] to (fm33);
		% the information flow
		\draw[-{latex[scale=1.5]},line width=0.4mm] (input) [out=90, in=-90] to node[below] {$~h_1$} (fm11);
		\draw[-{latex[scale=1.5]},line width=0.4mm] (fm11) [out=90, in=-90] to node[below] {$~h_2$} (fm22);
		\draw[-{latex[scale=1.5]},line width=0.4mm] (fm22) [out=90, in=-90] to node[below] {$~h_3$} (fm33);
		% the output
		\node (output) at (-0.4, 4.4) {};
		\draw[-{latex[scale=1.5]},line width=0.4mm] (fm33) [out=90, in=-90] to node[above] {$h_4$} (output);
		\node (caption) at (-2.8,0.2) {\small(a) single};
    \begin{scope}[shift={(5,0)}]
        \node (input) at (-0.4,0) {};
		% the fm set: layer 1
		\node[draw,rectangle] (fm11) at (-1.2, 1) {$t_{1}$};
		\node[draw,rectangle] (fm12) at (-0.4,1) {$t_{2}$};
		\node[draw,rectangle] (fm13) at (0.4,1) {$t_{3}$};

        \node[draw,rectangle] (fm21) at (-1.2,2.2) {$t_{1}$};
		\node[draw,rectangle] (fm22) at (-0.4,2.2) {$t_{2}$};
		\node[draw,rectangle] (fm23) at (0.4,2.2) {$t_{3}$};

        \node[draw,rectangle] (fm31) at (-1.2,3.4) {$t_{1}$};
		\node[draw,rectangle] (fm32) at (-0.4,3.4) {$t_{2}$};
		\node[draw,rectangle] (fm33) at (0.4,3.4) {$t_{3}$};
		
		\node (rout1) at (-2.8, 1.7) {\small $\pi_1(h_1)$};
		\node (rout2) at (-2.8, 2.2) {\small $\pi_2(h_2)$};
		\node (rout3) at (-2.8, 2.7) {\small $\pi_3(h_3)$};
	    \draw (-3.5,1.1) -- (-2,1.1) -- (-2,3.2) -- (-3.5,3.2) -- (-3.5,1.1);
	    \node (routerbox) at (-2.9, 1.3) {\texttt{router}};
		\draw[-{latex[scale=1.5]},line width=0.2mm,dotted] (rout1) [out=0, in=-120] to (fm11);
		\draw[-{latex[scale=1.5]},line width=0.2mm,dotted] (rout2) [out=0, in=-140] to (fm22);
		\draw[-{latex[scale=1.5]},line width=0.2mm,dotted] (rout3) [out=0, in=-140] to (fm33);
		% the information flow
		\draw[-{latex[scale=1.5]},line width=0.4mm] (input) [out=90, in=-90] to node[below] {$~h_1$} (fm11);
		\draw[-{latex[scale=1.5]},line width=0.4mm] (fm11) [out=90, in=-90] to node[below] {$~h_2$} (fm22);
		\draw[-{latex[scale=1.5]},line width=0.4mm] (fm22) [out=90, in=-90] to node[below] {$~h_3$} (fm33);
		% the output
		\node (output) at (-0.4, 4.4) {};
		\draw[-{latex[scale=1.5]},line width=0.4mm] (fm33) [out=90, in=-90] to node[above] {$h_4$} (output);
		\node (caption) at (-2.35,0.2) {\small(b) per-decision};
    \end{scope}
    \begin{scope}[shift={(10.5 ,0)}]
        \node (input) at (-0.4,0) {};
		% the fm set: layer 1
		\node[draw,rectangle] (fm11) at (-1.2, 1) {$t_{1}$};
		\node[draw,rectangle] (fm12) at (-0.4,1) {$t_{2}$};
		\node[draw,rectangle] (fm13) at (0.4,1) {$t_{3}$};

        \node[draw,rectangle] (fm21) at (-1.2,2.2) {$t_{1}$};
		\node[draw,rectangle] (fm22) at (-0.4,2.2) {$t_{2}$};
		\node[draw,rectangle] (fm23) at (0.4,2.2) {$t_{3}$};

        \node[draw,rectangle] (fm31) at (-1.2,3.4) {$t_{1}$};
		\node[draw,rectangle] (fm32) at (-0.4,3.4) {$t_{2}$};
		\node[draw,rectangle] (fm33) at (0.4,3.4) {$t_{3}$};
		
		\node (rout1) at (-3.6, 2.7) {$\pi_a(h)$};
		\node (rout2) at (-3, 2.7) {$...$};
		\node (rout3) at (-2.4, 2.7) {$\pi_k(h)$};
		\node (dispatcher) at (-3, 1.8) {\small $\pi(x)$};
		\draw[-{latex[scale=1.5]},line width=0.2mm] (dispatcher) [out=90, in=-140] to (rout3);
		
		% the routing selection
		% \node [below right, red] at (.5,.75) {below right};
	    \draw (-4.,1.1) -- (-2,1.1) -- (-2,3.2) -- (-4.,3.2) -- (-4.,1.1);
	    \node (routerbox) at (-3.4, 1.3) {\texttt{router}};
		\draw[-{latex[scale=1.5]},line width=0.2mm,dotted] (rout3) [out=-90, in=-140] to (fm11);
		\draw[-{latex[scale=1.5]},line width=0.2mm,dotted] (rout3) [out=-50, in=-130] to (fm22);
		\draw[-{latex[scale=1.5]},line width=0.2mm,dotted] (rout3) [out=30, in=-140] to (fm33);
		% the information flow
		\draw[-{latex[scale=1.5]},line width=0.4mm] (input) [out=90, in=-90] to node[right] {~\hspace{1mm}~$h_1=x$} (fm11);
		\draw[-{latex[scale=1.5]},line width=0.4mm] (fm11) [out=90, in=-90] to node[below] {$~h_2$} (fm22);
		\draw[-{latex[scale=1.5]},line width=0.4mm] (fm22) [out=90, in=-90] to node[below] {$~h_3$} (fm33);
		% the output
		\node (output) at (-0.4, 4.4) {};
		\draw[-{latex[scale=1.5]},line width=0.4mm] (fm33) [out=90, in=-90] to node[above] {$h_4$} (output);
		\node (caption) at (-3,0.2) {\small(c) dispatched};
    \end{scope}
	\end{tikzpicture}\vspace{-2mm}
    \captionof{figure}{Routing architectures}
    \label{fig:arch}
\end{figure*}
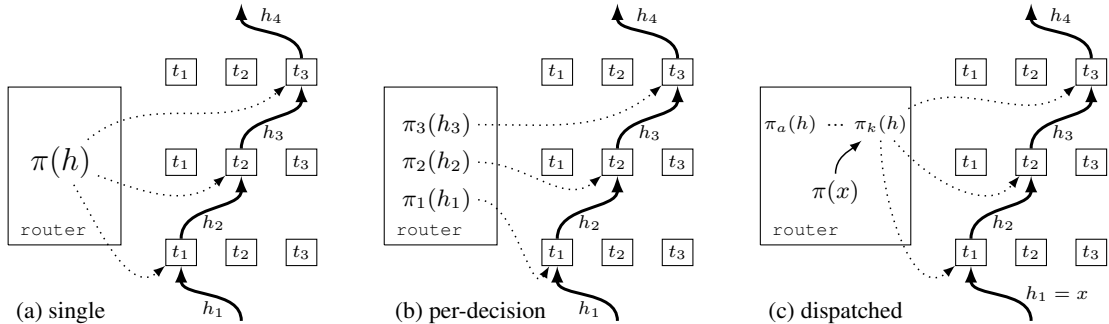

%When designing the router, the main question is how to map specific decisions to different ``subrouters''. 

In the conceptually simplest version used by \cite{crl,modularnets,ramachandran2018diversity} and unsuccessfully tried by \cite{routingnets}, there is only one router, that learns to make all required decisions (compare Figure \ref{fig:arch}(a)). This router's statespace contains the possible space of all activations, including the input activation. However, it requires additional work to stabilize -- curriculum learning for \cite{crl}, separate grouping for \cite{modularnets} and soft-routing for \cite{ramachandran2018diversity}. \cite{routingnets} introduce two other interesting router architectures that consist of multiple subrouters. The first is assigning one subrouter to each decision problem (see Figure \ref{fig:arch}(b)). Consequently, each subrouter has a statespace constrained to the activations possible at the respective depth. In general, this approach has the disadvantage that it only allows a recursion depth not deeper than the number of subrouters defined. However, it is an effective way to implement routing architectures where the modules available at different timesteps have to be different\footnote{This can result as a consequence of e.g., dimensionality constraints; other approaches to solve this problem may include solutions where the agent, if it chooses an incompatible action, is heavily penalized and the episode ends immediately. We assume that several existing approaches also use separate approximations to deal with these dimensionality problems, though they do not mention this.}. The second option discussed in \citep{routingnets} is what they call a ``dispatched'' routing network (compare Figure \ref{fig:arch}(c)). This hierarchical configuration has an extra preceding subrouter with the interesting task to assign -- or cluster -- samples in input space first, before assigning the sample to be routed to one of a set of parallel subrouters, each of which exclusively works in activation space.

Another routing architecture design choice is the available action space. One already discussed option -- particularly relevant for fully recursive routing models -- is to include a ``termination'' action to stop routing and forward the last activation. This termination action, is not required for per-decision router designs (Figure \ref{fig:arch}(b)), but could be implemented if model constraints permit. Another one, introduced by \cite{routingnets} for per-decision routing networks, is to include a ``skip'' action that does not terminate routing, but instead simply skips one sub-router. These two actions can result in identical behavior for limited-depth routing networks. In practice, dividing the decision problem over multiple subrouters, each with only a subset of actions and states to learn, can make training the router considerably easier. In particular in early stages of training, the problem of training a policy on modules that are similarly untrained can make fully recursive, single subrouter routing models hard to train. We will discuss this -- and related problems -- in more detail in the following section.
\section{Evaluation}\label{sec:eval}
For evaluation, we consider two different domains: image classification and natural language inference. For image classification, we follow the architecture of \cite{routingnets}, i.e., we route the fully connected layers of a simple convolutional network with three $3\times 3$ convolutional layers. For natural language inference, we follow the successful architecture by \cite{routingnaacl}, i.e., we route the word projections of a standard sentence to sentence comparison architecture.

Following the general architectures choices described in Section \ref{sec:routing-architectures}, we try to do each experiment twice. One experiment relies on meta-information, i.e., task-labels, using a task-wise dispatched architecture, where each task is assigned to a separate subrouter which only routes based on meta-information and stores its policy in a table. The other experiment covers architectures without a dispatching subagent that route without any meta-information based on input sample information and consecutive activations.

For image classification experiments, we use CIFAR 100, as it has predefined `tasks' in form of its `coarse' label structure, allowing us to naturally implement routing networks relying on meta-information. All experiments try to predict the `fine' label in the context of the coarse label, i.e., the problem is five-way classification. All language inference experiments are on the Stanford Corpus of Implicatives \citep{routingnaacl}, which is a three-way classificaion task. For experiments relying on meta-information, we use the provided `signatures' as task-labels.

All results shown in this section are results on test datasets. The entropy plots show the entropy of the selection distribution over the entire dataset, also at test time. As most of the results are meant to be qualitative in nature, we did not do extensive hyperparameter searches, and we do expect better results to be found. As we will show in the following section, Q-Learning is consistently among the best performing decision making algorithms. Thus, unless otherwise noted, all plots utilize Q-Learning.

\subsection{Decision Making Strategies}

In this section, we evaluate different decision making strategies. Evaluated strategies include reparameterization strategies and several reinforcement learning algorithms. 

\subsubsection{Learning with meta-information}
\begin{figure}[!ht]
    \centering
    \subcaptionbox{CIFAR 100 MTL, with depth of 1}{
    \includegraphics[width=0.32\textwidth]{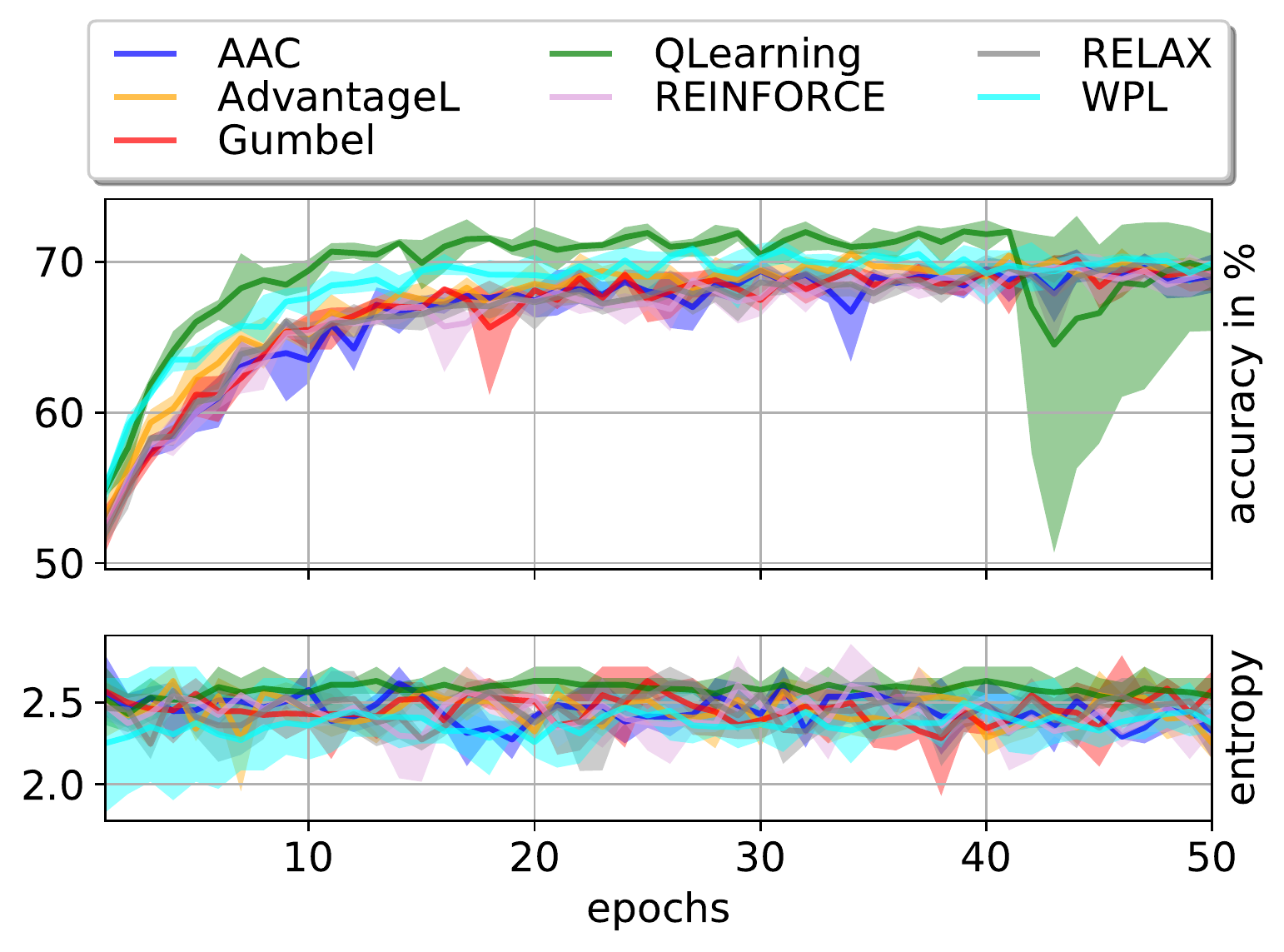}}
    \subcaptionbox{CIFAR 100 MTL, with depth of 2}{
    \includegraphics[width=0.32\textwidth]{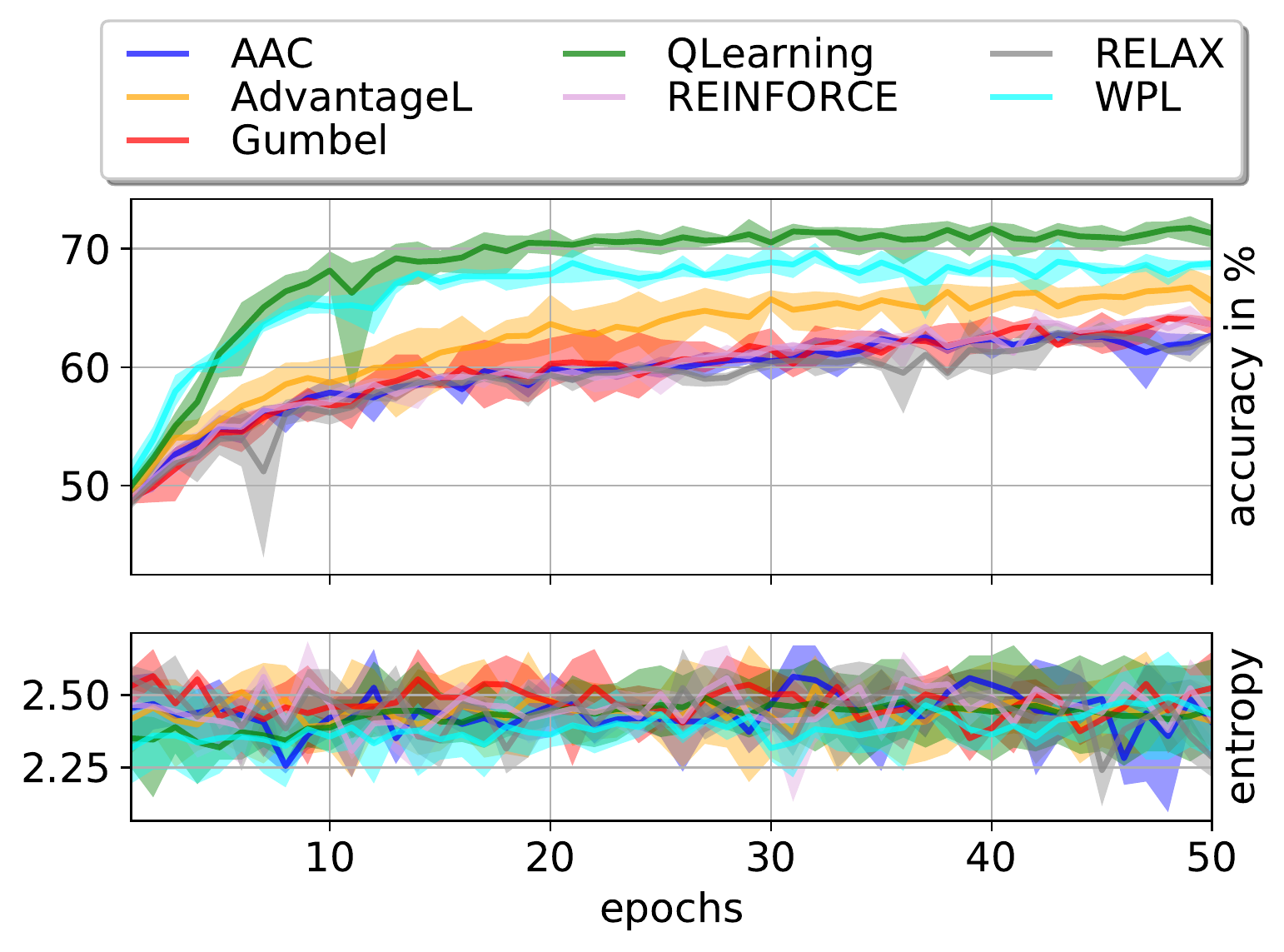}}
    \subcaptionbox{SCI MTL, with depth of 3}{
    \includegraphics[width=0.32\textwidth]{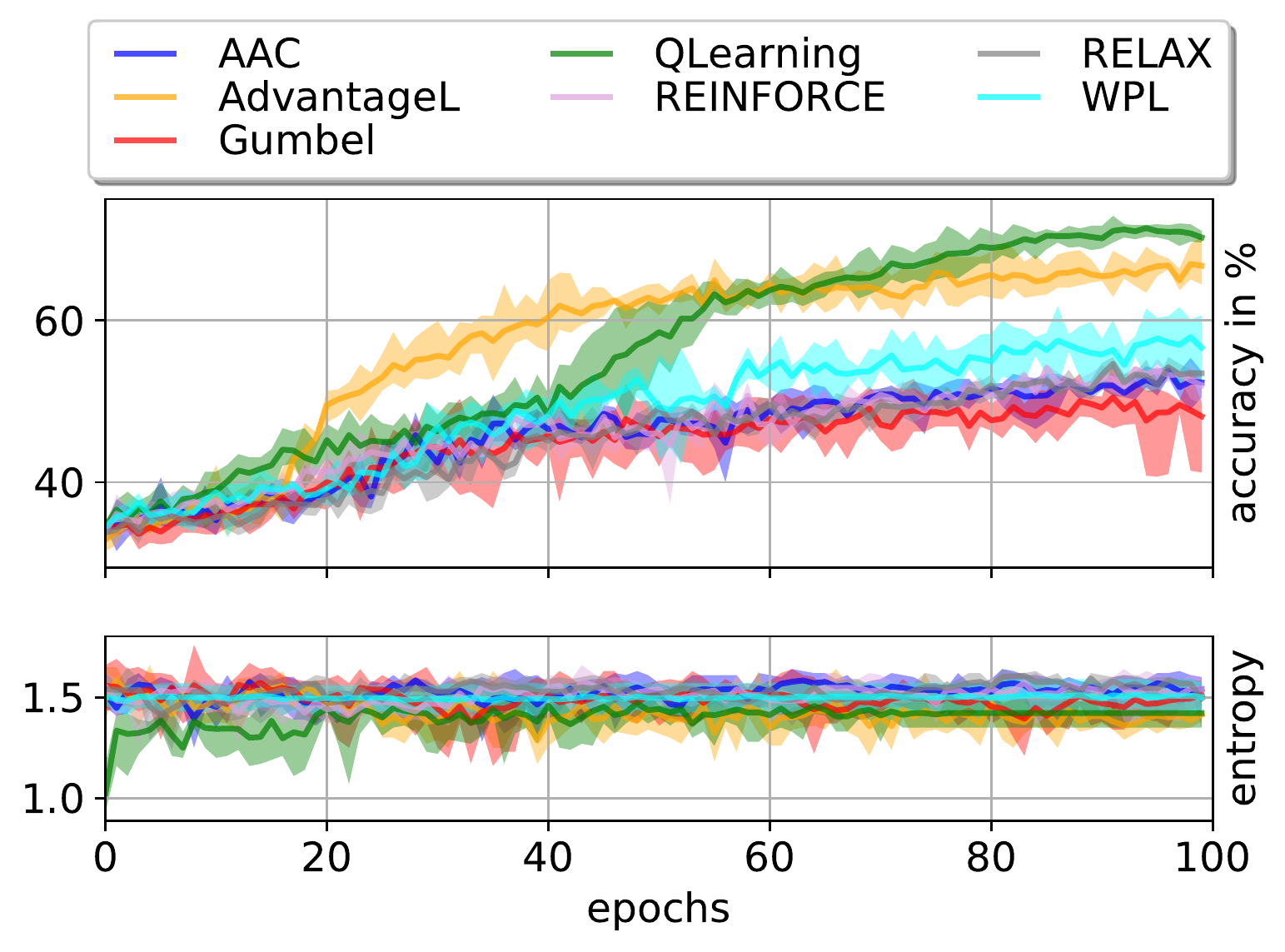}}
    \caption{Multi-task results for different decision making strategies}
    % \caption{Results on CIFAR 100 MTL (for multi-task experiments) and CIFAR 10 (for single-task experiments) for different decision making strategies; }
    \label{fig:res-dm}
\end{figure}
Figure \ref{fig:res-dm} shows results for different decision making algorithms when relying on meta-information. The starkest result is that the policy gradient based approaches, including the related reparameterization algorithms -- Gumbel and RELAX, are consistently outperformed by value-based reinforcement learning based approaches. As can be seen when comparing Figures \ref{fig:res-dm}(a) with (b), this difference grows with routing depth -- i.e., number of routed layers. We stipulate that this difference results from the effect exploration has on interference, as discussed in section \ref{sec:chall-collapse}. 
In contrast to policy gradient strategies, $\epsilon$-greedy strategies are guaranteed to interfere with another transformation, even of near-identical value, only a fraction of the time purely determined by $\epsilon$.
This also explains the best performing policy gradient based approach, the weighted policy learner (WPL) \citep{abdallah_wpl}, which was already successfully tested in the routing multi-agent setting of \cite{routingnets}. WPL is a non-standard policy gradient algorithm (in that it does not use the REINFORCE log-probability trick) that was explicitly designed for stochastic games, and that has specific properties to emphasize exploitation even for similar-value actions. This effectively lowers its exploration and thereby its interference.
%We stipulate that this is a consequence of stability discussed in section \ref{sec:chall-collapse}, as policy gradient based approaches cannot stabilize by simply acting greedily. 

In Figure \ref{fig:res-dm}(c), the language inference experiments, it is apparent how different algorithms stabilize (or not) over time. As for CIFAR, the value-based approaches Q-Learning and Advantage learning have a clear edge over the policy gradient based approaches. It is particularly interesting to see how both algorithms have nearly the same learning behavior for the first 20 epochs, until Advantage learning, and 20 epochs later, Q-Learning, apparently stabilize to then quickly leave the other algorithms behind. We speculate that the benefit of Advantage learning stems from the argument in Section \ref{sec:routing-algs}, as it is able to offset the general increase in value of the different transformations.

\subsubsection{Learning without meta-information} \label{sec:eval-wo-meta}
Routing decisions that do not rely on meta-information rely on the activation at the input to the routed layer instead. That is, they only consider the activation produced by the previous layer, which may or may not be routed, and which may even be the input. This allows the most fine-grained control over the routing path, as -- possibly -- each sample may be routed through a different path. The most straightforward implementation is a `single router' architecture depicted in Figure \ref{fig:arch}(a), where the router consists of only one subrouter -- one parameterization that gets passed in activations from any layer. This also allows arbitrary depth routing networks, as the router may decide to apply an arbitrary number of transformations before stopping. Unfortunately, this complicates the routing process in at least two ways. The first is that the subrouter does not only need to learn where to route based on an activation alone, but even needs to do so for activations that may have been produced at completely different steps in the routing network. This makes the distribution of activations the router will have to decide on dramatically more difficult for the router to interpret, as it is likely that each `depth' of activations adds another mode to the distribution. The second, related, reason is that the activations are highly volatile, as any change to the subrouter’s policy may dramatically transform the distribution of activations, making the modes non-stationary and even more difficult to disentangle. For illustration, consider a simple single router model where the router can select from $\{t_1,t_2,t_3\}$. At a given training time, the router only selects $t_1$ or $t_2$. When an update changes the router to use $t_3$ instead of $t_2$, the space of activations will now have been created by $\{t_1,t_3\}^k$, instead of by $\{t_1,t_2\}^k$. It is obvious how this can `confuse' a router, in particular when this non-stationarity is combined with the already existing non-stationarity produced by updating the transformations.

\begin{wrapfigure}{r}{0.68\textwidth}
    \centering
     \vspace{-4mm}
    \subcaptionbox{CIFAR, single router}{
    \includegraphics[width=0.32\textwidth]{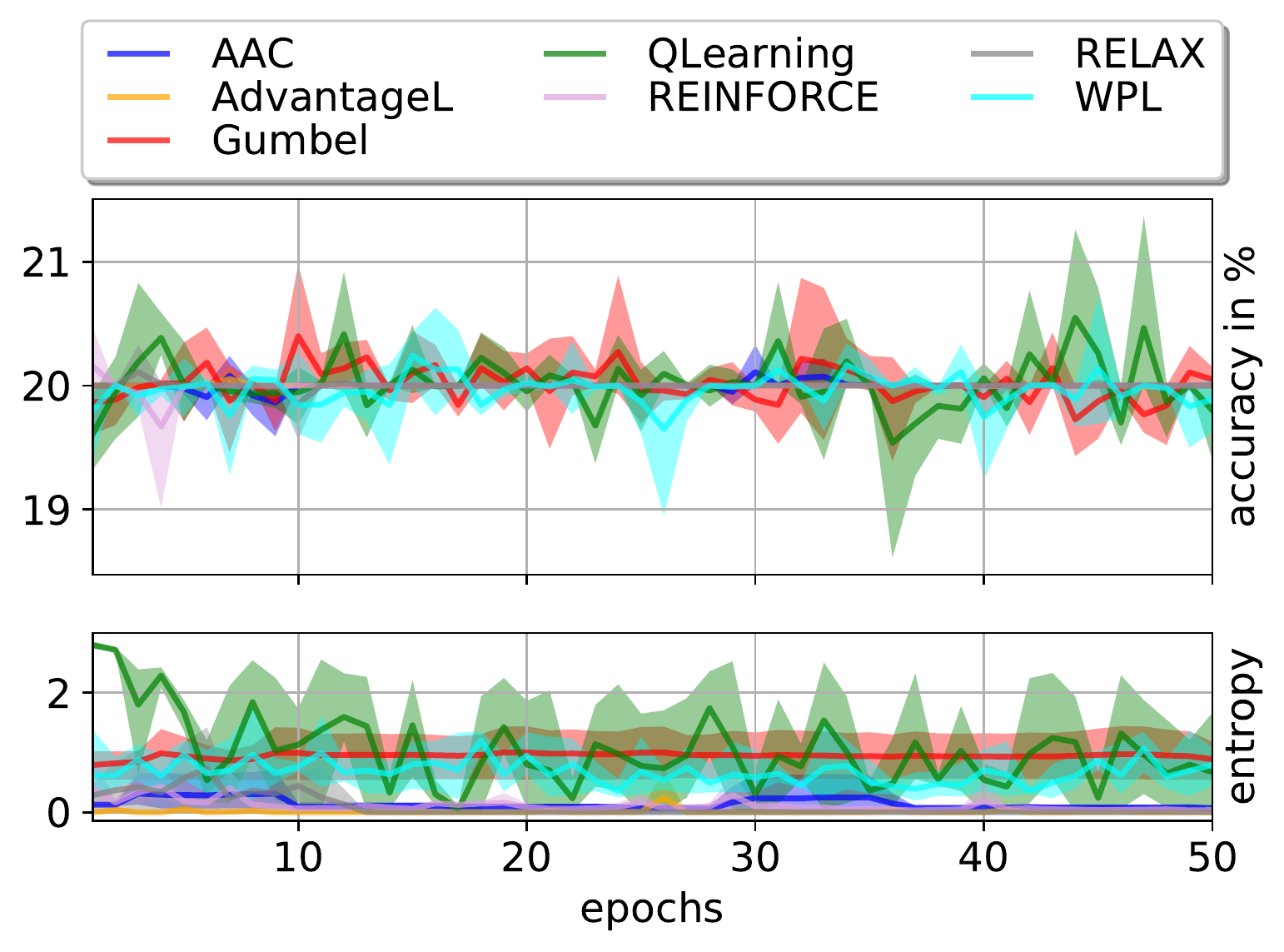}}
    \hspace{0.01\textwidth}
    \subcaptionbox{SCI, single router}{
    \includegraphics[width=0.32\textwidth]{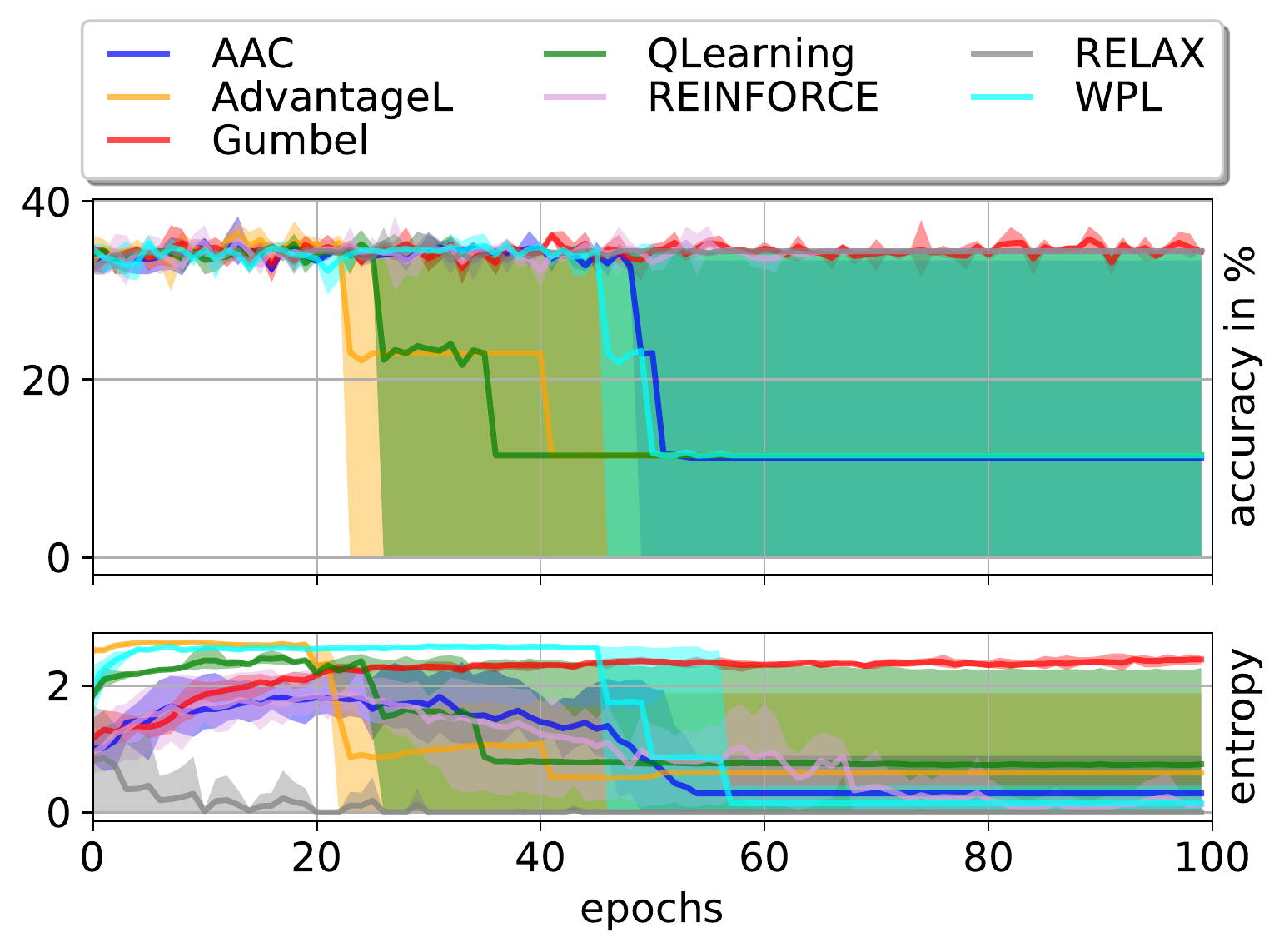}}
    \caption{Single router results}
    \vspace{-3mm}
    % \caption{Results on CIFAR 100 MTL (for multi-task experiments) and CIFAR 10 (for single-task experiments) for different decision making strategies; }
    % \vspace{-5mm}
    \label{fig:res-dm-single}
\end{wrapfigure}
Consequently, performance on both CIFAR and SCI, as depicted in Figure \ref{fig:res-dm-single} is basically the same as random. For the results on language inference in Figure \ref{fig:res-dm-single}(b), the routing problem becomes so complex  that the router even picks a fourth class for language inference (which generally only has three) that only exists for implementation reasons, thereby consistently achieving 0\% accuracy instead of the 33\% of the random baseline.

To investigate the problem posed by these results, we additionally experiment with two more architectures not relying on meta-information. In the first, we fix the depth for the single router architecture, and in the second, we design a per-layer subrouter architecture depicted in Figure \ref{fig:arch}(b). Other than the single router architecture used for the experiments in Figures \ref{fig:res-dm-single},  this architecture does not have one approximation network over all routing layers, of which there may be an arbitrary number, but instead only makes (at most) $k$ decisions, with a separate subrouter for each layer. 
\begin{wrapfigure}{r}{0.68\textwidth}
    \vspace{-2mm}
    \centering
    \subcaptionbox{Singe router, fixed depth of 3}{
    \includegraphics[width=0.32\textwidth]{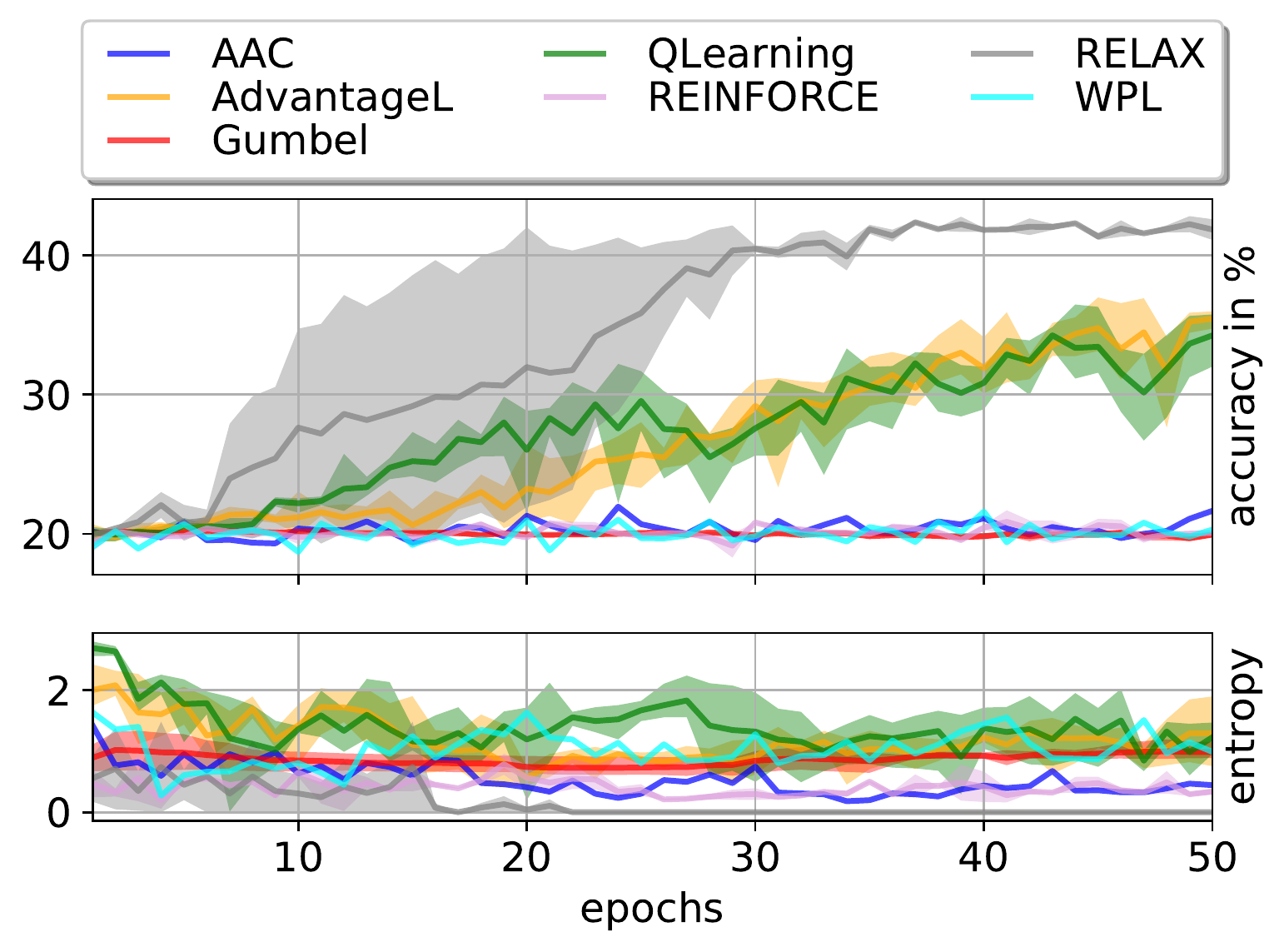}}
    \hspace{0.01\textwidth}
    \subcaptionbox{Separate subrouter per layer}{
    \includegraphics[width=0.32\textwidth]{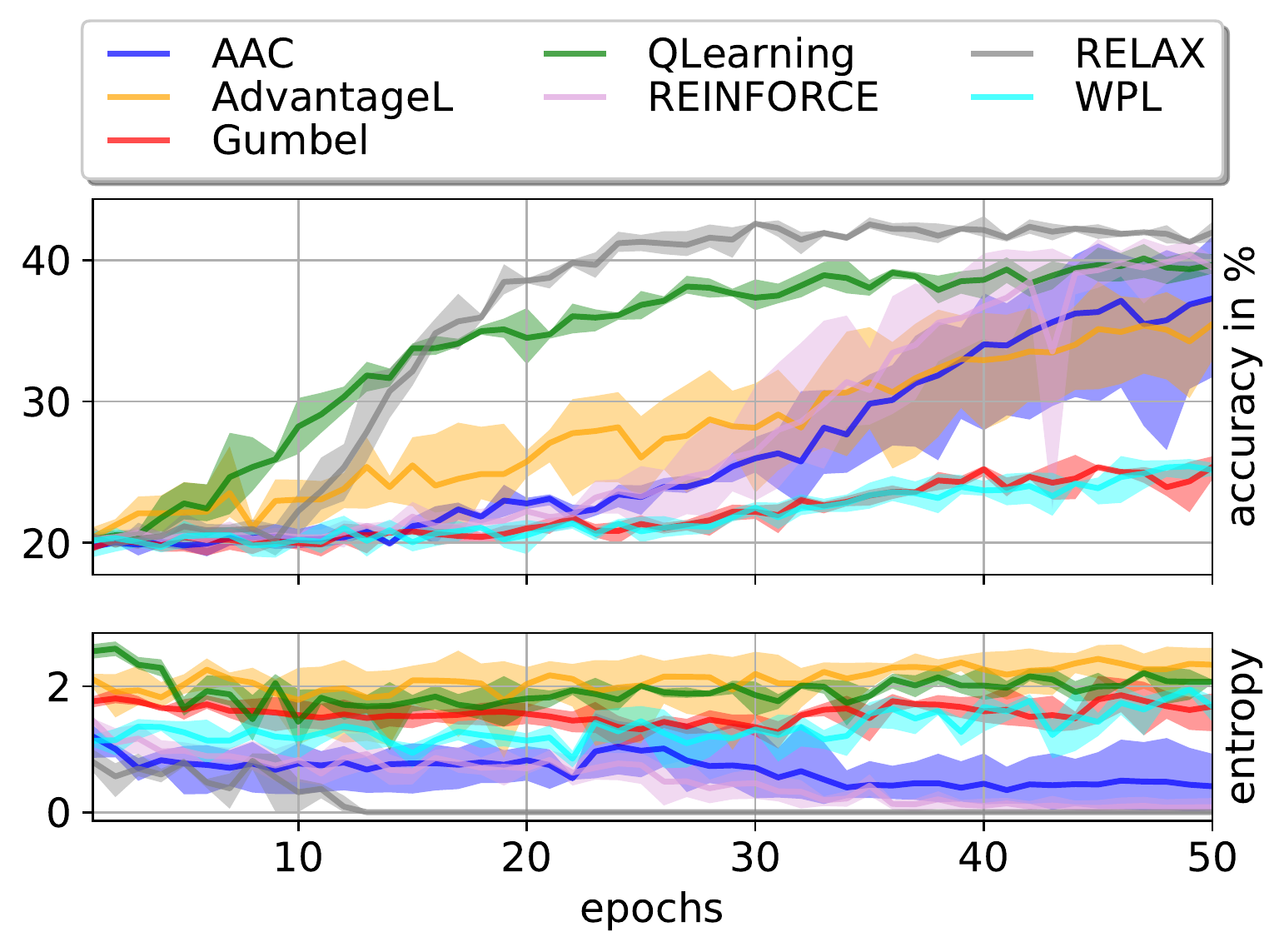}}
    \caption{No-metainformation ablation studies on CIFAR}
    \label{fig:res-dm-single-alt}
    \vspace{-3mm}
\end{wrapfigure}
As shown in Figure \ref{fig:res-dm-single-alt}(a), fixing the routing depth does help to stabilize the training for some algorithms. As with other experiments, the (Q-)value based approaches start learning, but, more surprisingly, RELAX is able to stabilize, although at the cost of complete collapse. We assume that in this highly volatile environment the rich gradient information provided by RELAX can help the most. In Figure \ref{fig:res-dm-single-alt}, we show results for a separate subrouter for each layer, thereby limiting the maximum number of applied transformations to 3 (each subrouter is able to terminate earlier). This makes the decision making problem easier so that some algorithms are able to stabilize. However, most policy gradient approaches, including RELAX, collapse to achieve this stability, while the remaining policy gradient algorithms, WPL and Gumbel, do not achieve noteworthy performance. Only Q-Learning and Advantage learning learn while maintaining selection entropy. Interestingly, the router apparently does not learn to completely minimize interference, as it does not achieve non-routed performance, not even when it collapses. It should be noted, though, that even for  algorithms that successfully stabilize, the solutions found are local optima of very bad performance. For both the results in Figure \ref{fig:res-dm-single-alt}, the final performance stays about 30\% below the performance achieved by models relying on meta-information.

These results strongly suggest that routing networks -- and quite probably any other kind of approach to compositional computation -- have difficulty overcoming the initial instability of the training process, if they are not provided with a strong external signal. One such signal, as discussed in the previous section, are task-labels or other pieces of meta-information that allow the routing network to find paths at the dimensionality of the meta-information, which tends to be \textit{much} smaller than the full space of activations. Another approach, as introduced by \cite{crl}, is to carefully curate the order in which samples are provided to the network. This, also, drastically limits the complexity of the decision problem, as only `similar' samples are shown at any given time during training, but relies on such an order to exist for a given dataset.

\subsubsection{Reparameterization Techniques and Exploration Strategies}
\begin{wrapfigure}{r}{0.34\textwidth}
    \centering
    \vspace{-4mm}
    % \vspace{-12mm}
    \includegraphics[width=0.32\textwidth]{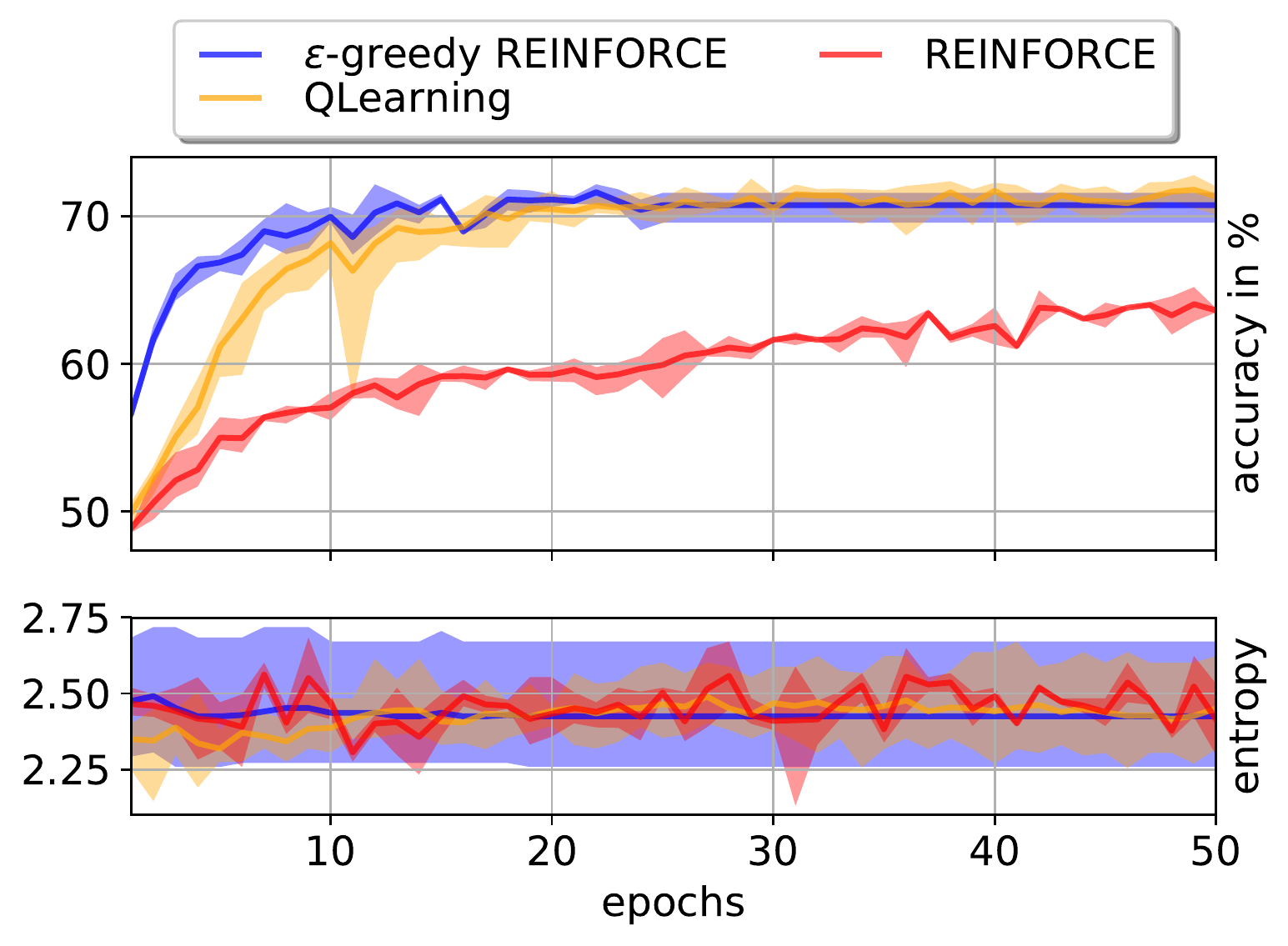}
    % \caption{Results on CIFAR 100 MTL (for multi-task experiments) and CIFAR 10 (for single-task experiments) for different decision making strategies; }
    \caption{A greedy version of REINFORCE on CIFAR 100 MTL}
    \label{fig:res-pg-exp}
    \vspace{-3mm}
\end{wrapfigure}
In general, the results for reparameterized routing, i.e., using Gumbel and RELAX, are very similar to the results for policy gradient approaches. We even found that for deeper architectures, the reparameterization techniques suffer from the same decrease in performance as other PG algorithms, as shown in Figure \ref{fig:res-dm}(b). As mentioned above, we assume that this stems from inferior exploration behavior for on-policy stochastic sampling. To verify, we designed a version of REINFORCE that samples actions using an $\epsilon$-greedy policy, i.e., that takes the best action $\epsilon$ of the time, and that samples from the actual policy $1-\epsilon$ of the time. When training, we compensate for the difference in sampling and training policies using importance sampling \citep{precup_importance}. As shown in Figure \ref{fig:res-pg-exp}, this indeed changes the behavior of REINFORCE to act more like a value-based approach.\footnote{We should also note that exploration for the Q-Learning experiments is \textit{very} high, with over 0.4 for the first 5 epochs, which suggests that the amount of exploration may play a role, but that the exploitation strategy dominates the results.} 
These results suggest that routing may greatly benefit from a targeted decision making algorithm. In particular exploration, with its added complexity of causing interference, seems to be a promising direction for future research. For example, it may be useful to consider algorithms that have separate exploitation and exploration policies, along the lines of \cite{francisco_exploration}, or algorithms that can incorporate more complex exploration mechanics into reparameterization techniques.

\subsection{Reward Design}

\begin{wrapfigure}{r}{0.68\textwidth}
    \centering
    \vspace{-1mm}
    \subcaptionbox{Final reward on CIFAR 100 MTL}{
    \includegraphics[width=0.32\textwidth]{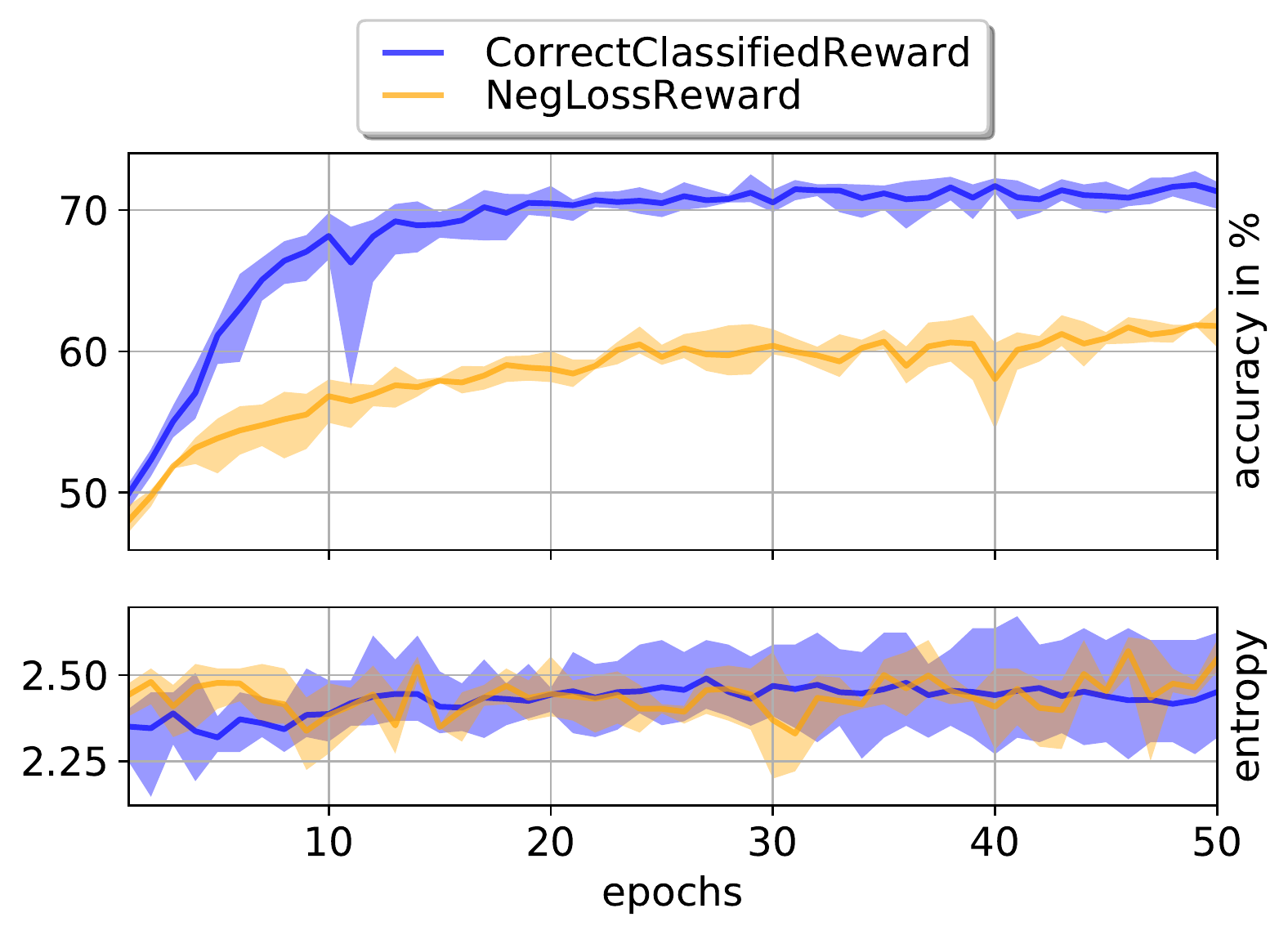}}
    \subcaptionbox{Final reward on SCI MTL}{
    \includegraphics[width=0.32\textwidth]{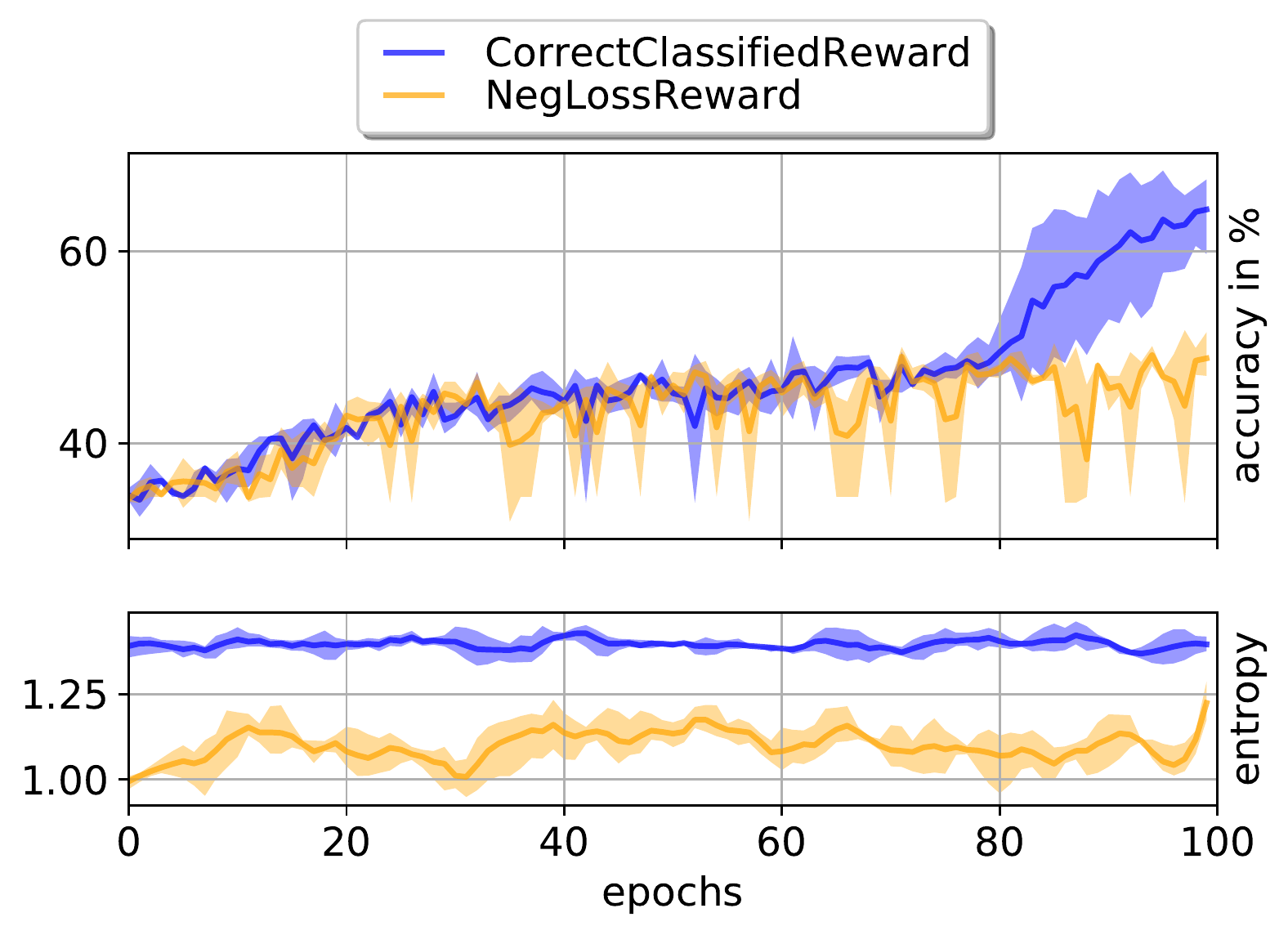}}
    \caption{Multi-task results for different reward functions}
    \label{fig:res-rewardtypes}
    \vspace{-3mm}
\end{wrapfigure}

Figure \ref{fig:res-rewardtypes} shows results for different final reward functions. In general, the correct/incorrect final reward strategy, $\pm 1$ for correct and incorrect classification, appears clearly superior to the negative classification (cross-entropy) loss reward ($r_f=-\mathcal{L}(\hat{y}, y)$. While this seems initially surprising, as the negative loss reward contains more information, we believe that this is exactly what makes learning more difficult, for two reasons. First, a negative classification loss may overemphasize outliers, as, relatively to other samples, it may add a much larger part to the total loss. While this may be useful for supervised learning, a reinforcement learning router may update to accommodate these samples at the cost of other, non-outlyer samples.
Second, as suggested by the high fluctuation in the learning on SCI, the finer granularity of the negative classification loss reward may reduce the difference between the value of different transformations, resulting in even small updates changing the greedy strategy. As this reduces stability, it will also decrease overall performance.

\begin{wrapfigure}{r}{0.68\textwidth}
    \vspace{-3mm}
    \centering
    \subcaptionbox{Per-action reward on CIFAR 100 MTL}{
    \includegraphics[width=0.32\textwidth]{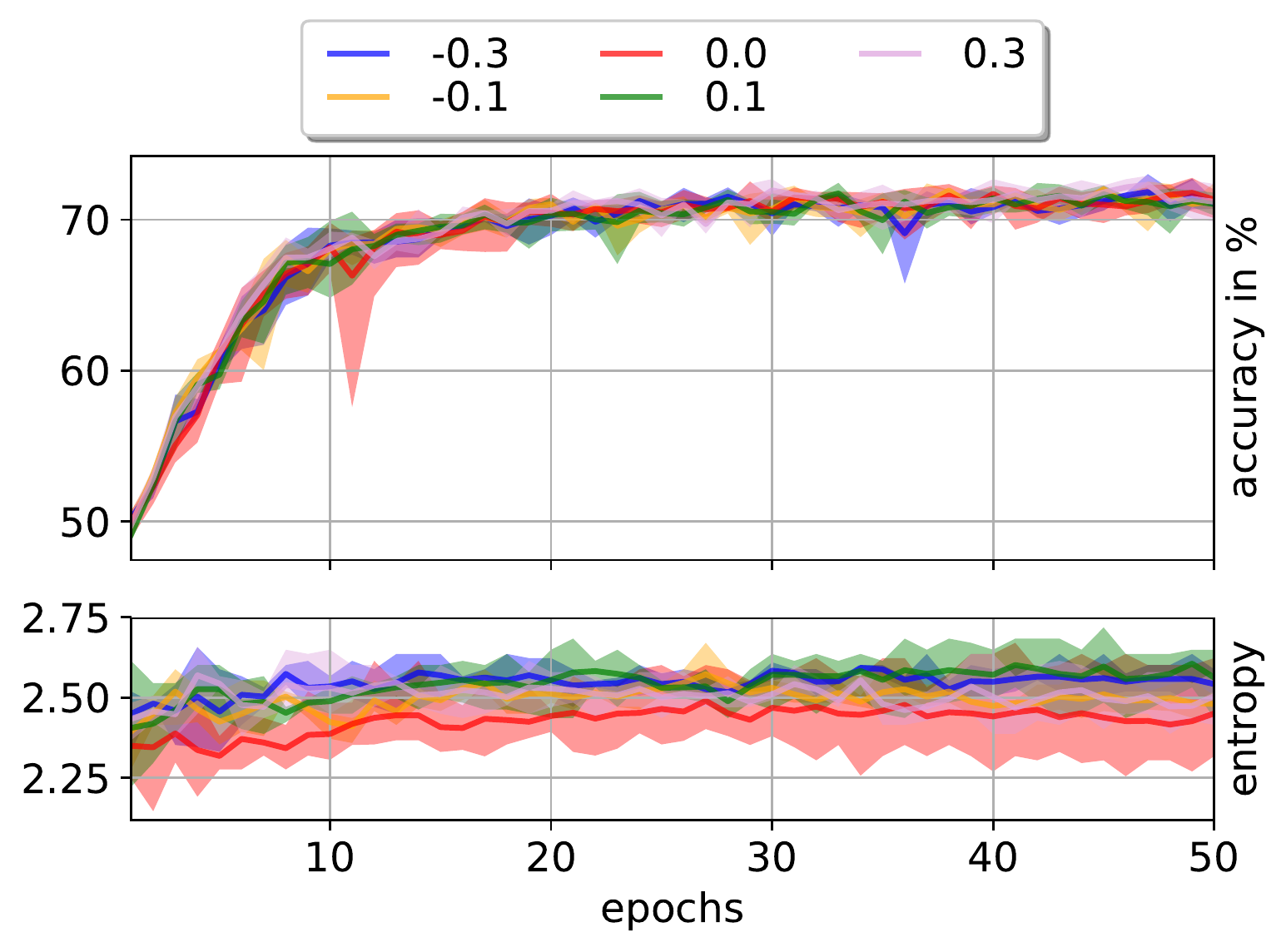}}
    \subcaptionbox{Per-action reward on SCI MTL}{
    \includegraphics[width=0.32\textwidth]{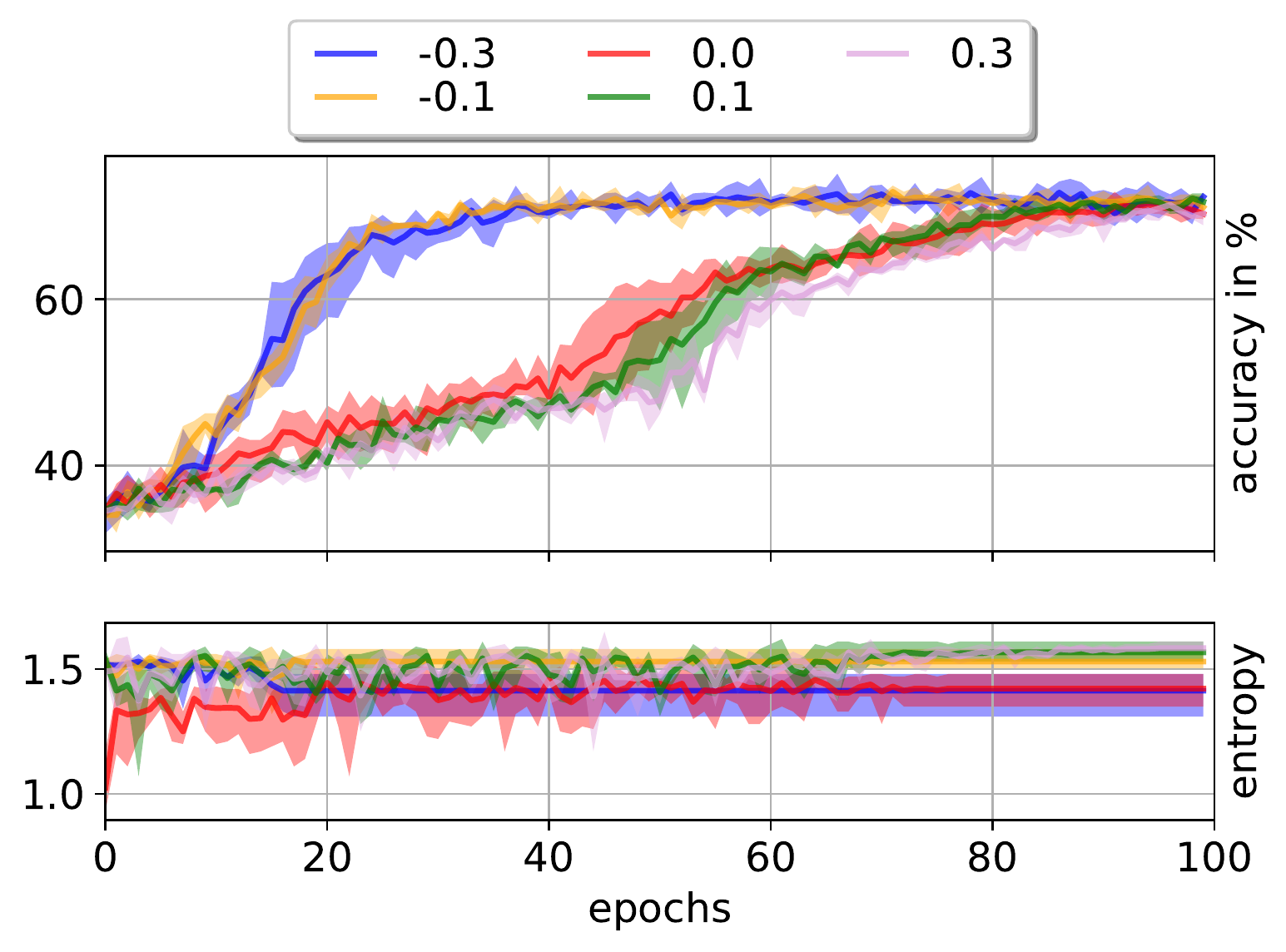}}
    \caption{Multi-task results for different intermediate reward values}
    \label{fig:res-rewardvals}
    \vspace{-2mm}
\end{wrapfigure}

Figure \ref{fig:res-rewardvals} shows plots for different intermediate reward values. These rewards are computed as a fraction of the overall probability of choosing a particular transformation. Positive values are expected to increase transfer and lower entropy, while negative values are expected to decrease interference and increase entropy. Comparing the results for CIFAR in Figure \ref{fig:res-rewardvals}(a) with the results for SCI in Figure \ref{fig:res-rewardvals}(b) it becomes apparent that the domain -- or the architecture -- plays a major role in the effect of this regularization reward. While the reward has no discernable effect on CIFAR, it dramatically changes the convergence behavior -- if not the final performance -- on SCI. Interestingly, negative reward values, 
meant to increase diversity, appear to stabilize learning as the model learns much faster, but have no effect on the routing entropy. While a complete interpretation of this result is very difficult, we speculate that this results from the higher potential for interference on SCI, and maybe even on language domains in general. The potential for interference is higher when training on SCI because the effective input dimensionality is nearly an order of magnitude lower than it is for CIFAR, which leads to a larger average overlap between examples in the input space. Additionally, as is also the case for us here, models used for NLP tend to use smaller hidden representations than computer vision models, resulting in a larger average overlap between examples in the activation space as well. This would not only explain why `pushing' the router to diversify stabilizes learning, but also why the results with the negative loss reward function have such a high variance, as any change in routing decision may end up with a much higher amount of interference.

These results suggest that future research should consider different new reward functions. In particular, it could be worthwhile to find a final reward function that can rely on the information contained in the negative loss, but that is less susceptible to its problems. Additionally, it would be interesting to investigate an adaptive intermediate reward that incentivizes diversity as long as it is needed for stability, but that eventually anneals to zero, so that the router only optimizes for the overall model performance.

\subsection{Other Design Choices}\label{sec:eval-optimizer}
\begin{figure*}[ht!]
    \centering
    \captionbox{Squashing training for exploratory trajectories on CIFAR100 MTL \label{fig:res-exp-squashing}}{
    \includegraphics[width=0.31\textwidth]{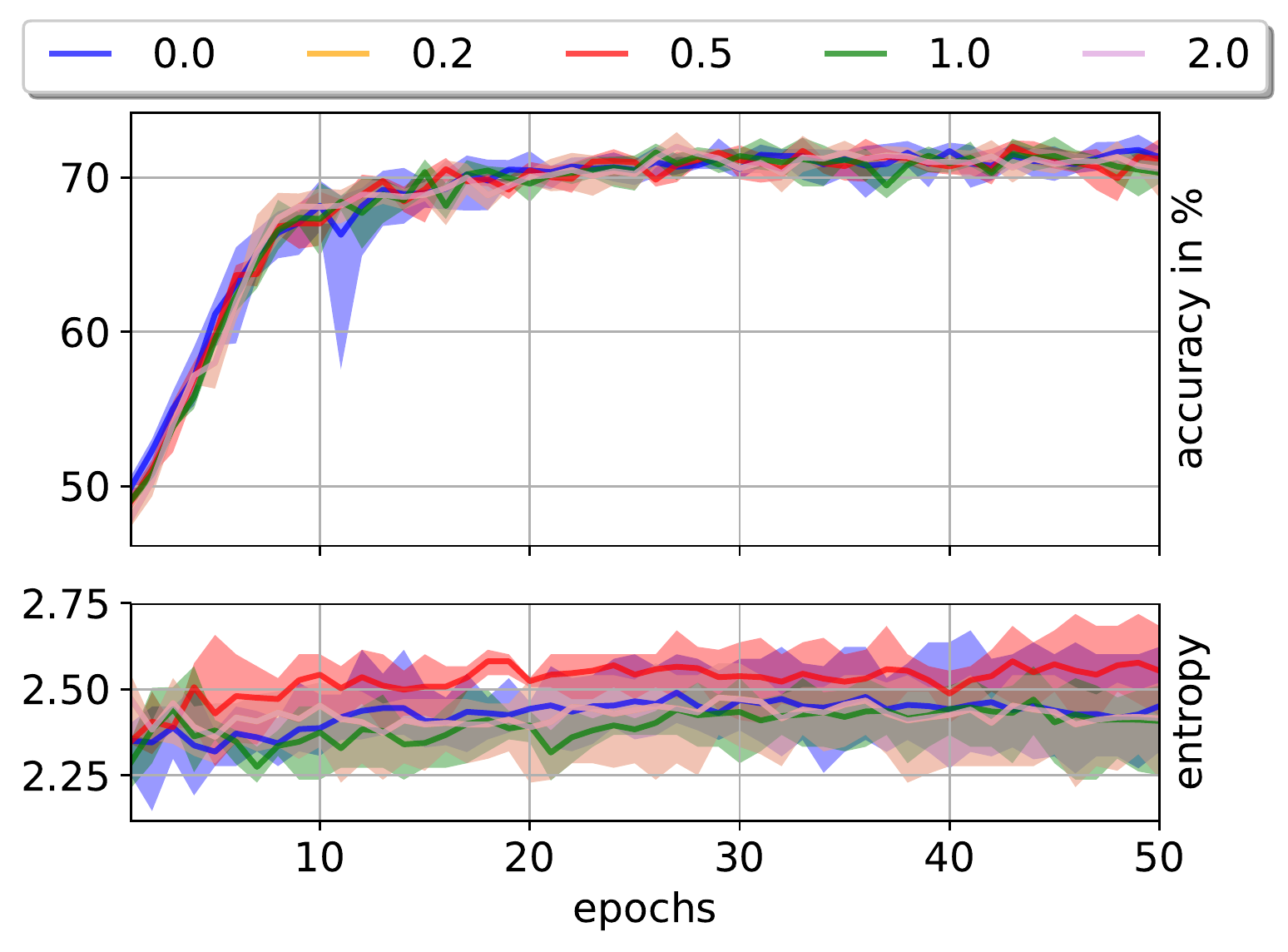}}
    % \subcaptionbox{Final reward in a single-task setting}{
    % \includegraphics[width=0.32\textwidth]{plots/results/cifar_rout_reward_type_stl.pdf}}
    \hspace{0.01\textwidth}
    \captionbox{Stochastic router-transformation training  on CIFAR100 MTL\label{fig:res-stoch-dec-mod}}{
    \includegraphics[width=0.31\textwidth]{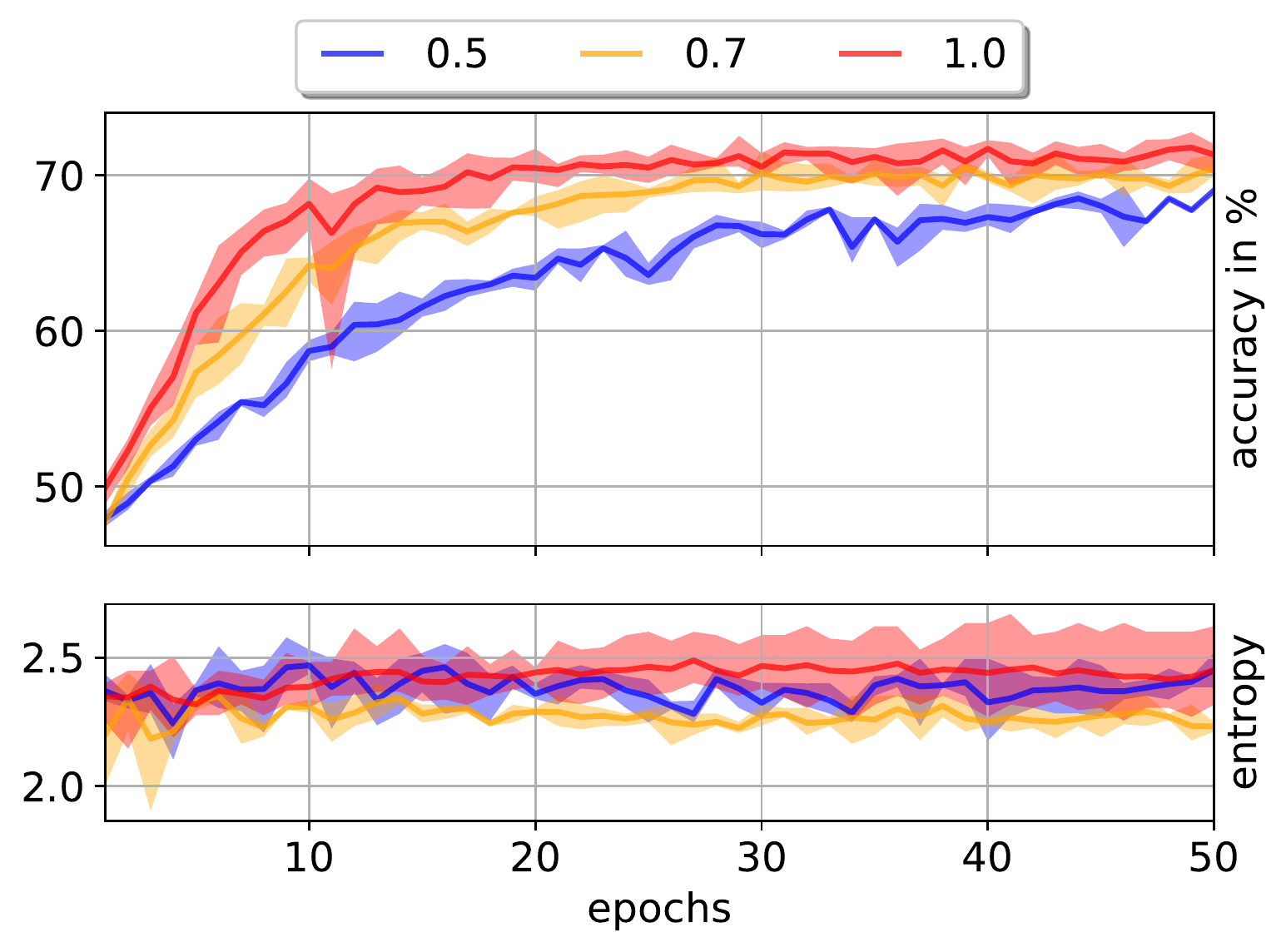}}
    \hspace{0.01\textwidth}
    \captionbox{The effect of the optimizer choice on CIFAR100 MTL\label{fig:res-optim}}{
    \includegraphics[width=0.31\textwidth]{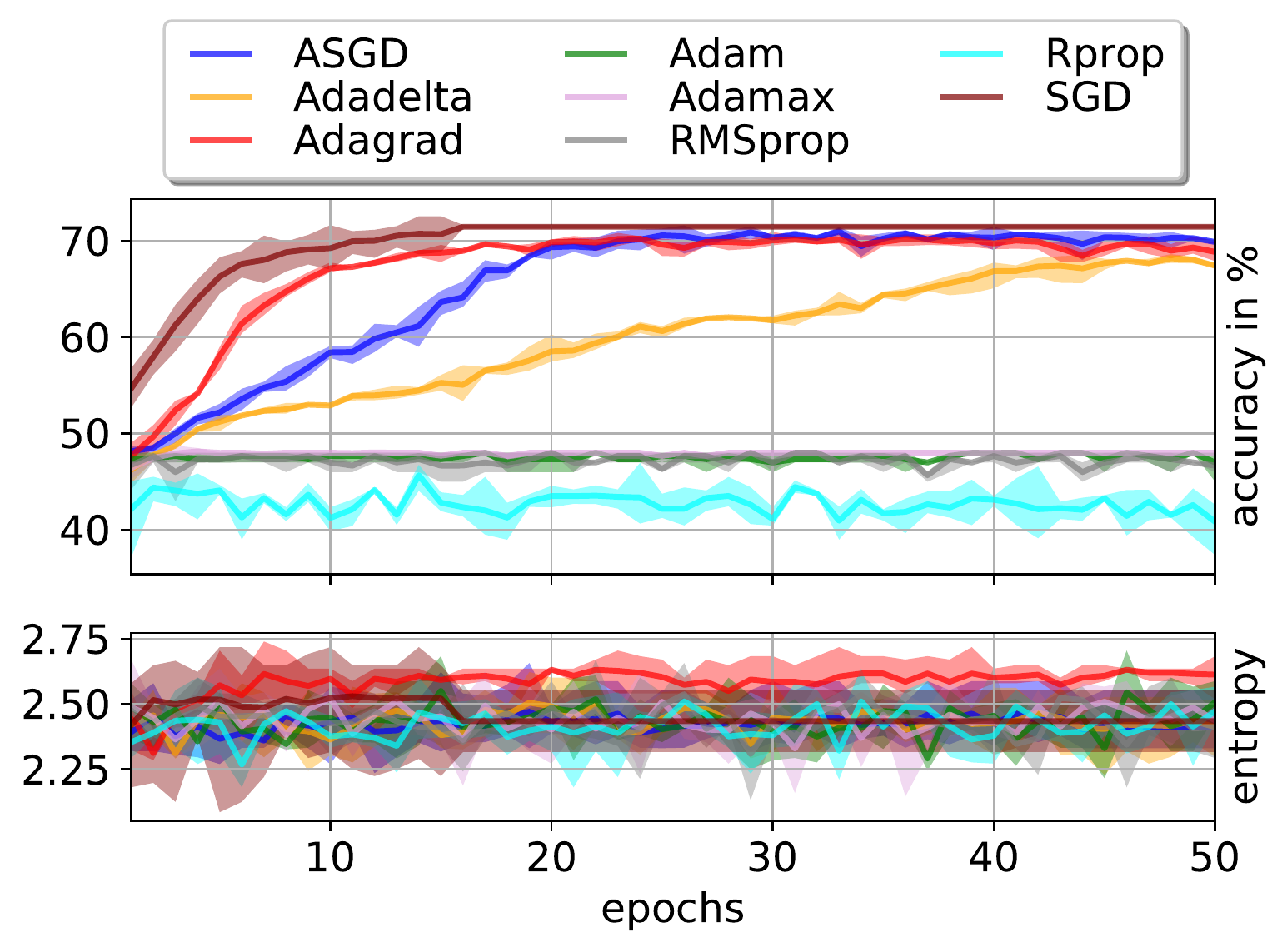}}
    % \caption{Results on CIFAR 100 MTL for different reward strategies}
\end{figure*}
Figure \ref{fig:res-exp-squashing} shows the effect that lowering the training of transformations chosen non-greedily has. As one can see, $\kappa$ has very little effect on the overall performance and only mild effect on the entropy.

Splitting the training set into samples to train the transformations and samples to train the router, on the other hand, has a clear effect on the performance, as shown in Figure \ref{fig:res-stoch-dec-mod}. Here, the curves show the percentage of the data used to train the transformations. As one can see, using the full data benefits both performance and entropy.

More interesting is Figure \ref{fig:res-optim}. As already reported by \cite{routingnets,routingnaacl}, routing becomes unstable for many optimizers, and consistently yields the best performance for `plain' SGD. We stipulate that similar problems can be observed for any architecture with an adaptive computation graph. Consider a simple example, where we route a sample $x$ through a two-layer routing network. The first routing decision can choose from transformations $\{t_{1,1}, t_{1,2}\}$ and the second from $\{t_{2,1},t_{2,2}\}$. 
%These are parameterized by parameters $\theta_{1,1}, \theta_{1,2}, \theta_{2,1}, \theta_{2,2}$, respectively. 
Now consider the gradients for $t_{1,2}$ over two different routing paths, $t_{1,2}(t_{2,1}(x))$ and $t_{1,2}(t_{2,2}(x))$. It is obvious that for those two paths, $t_{2,1}$ and $t_{2,2}$ may yield vastly different activations, and thereby vastly different inputs to $t_{1,2}$. 
If an optimizer relies on parameter-specific approximations, such as, e.g., momentum, an approximation for $t_{2,2}$ may be completely incompatible with $t_{2,1}$, thereby making training difficult, or even impossible. This explains why plain optimization strategies that do not compute parameter-specific information generally do better in the context of dynamic computation graphs.
% $\frac{\delta}{\delta \theta_{1,2}}t_{1,2}(t_{2,1}(x))$ and $\frac{\delta}{\delta \theta_{1,2}}t_{1,2}(t_{2,2}(x))$

% \subsection{Transfer Learning}

\subsection{Stability}\label{sec:eval-stability}
As we argued before, stability is a consequence of the routing `chicken-and-egg' problem: Upon initialization of a routing network, both the modules and the router do not yet have any information on the problem, and act randomly. The router cannot stabilize, as it cannot discern which selected module caused any increase in performance, and the modules cannot stabilize as they are trained with  samples and activations so different that interference can destroy any learning that may happen. This, in turn, may destroy any progress the router may have made with the credit assignment. This problem is never worse than for the `single router' architecture (compare Figure \ref{fig:arch}), for the reasons described in Section \ref{sec:eval-wo-meta}. Consequently, a routing network may fail to learn anything, as depicted in Figure \ref{fig:res-dm-single}.

\subsection{Module Collapse}\label{sec:eval-collapse}
\begin{wrapfigure}{r}{0.34\textwidth}
    \centering
    \vspace{-1mm}
    \includegraphics[width=0.32\textwidth]{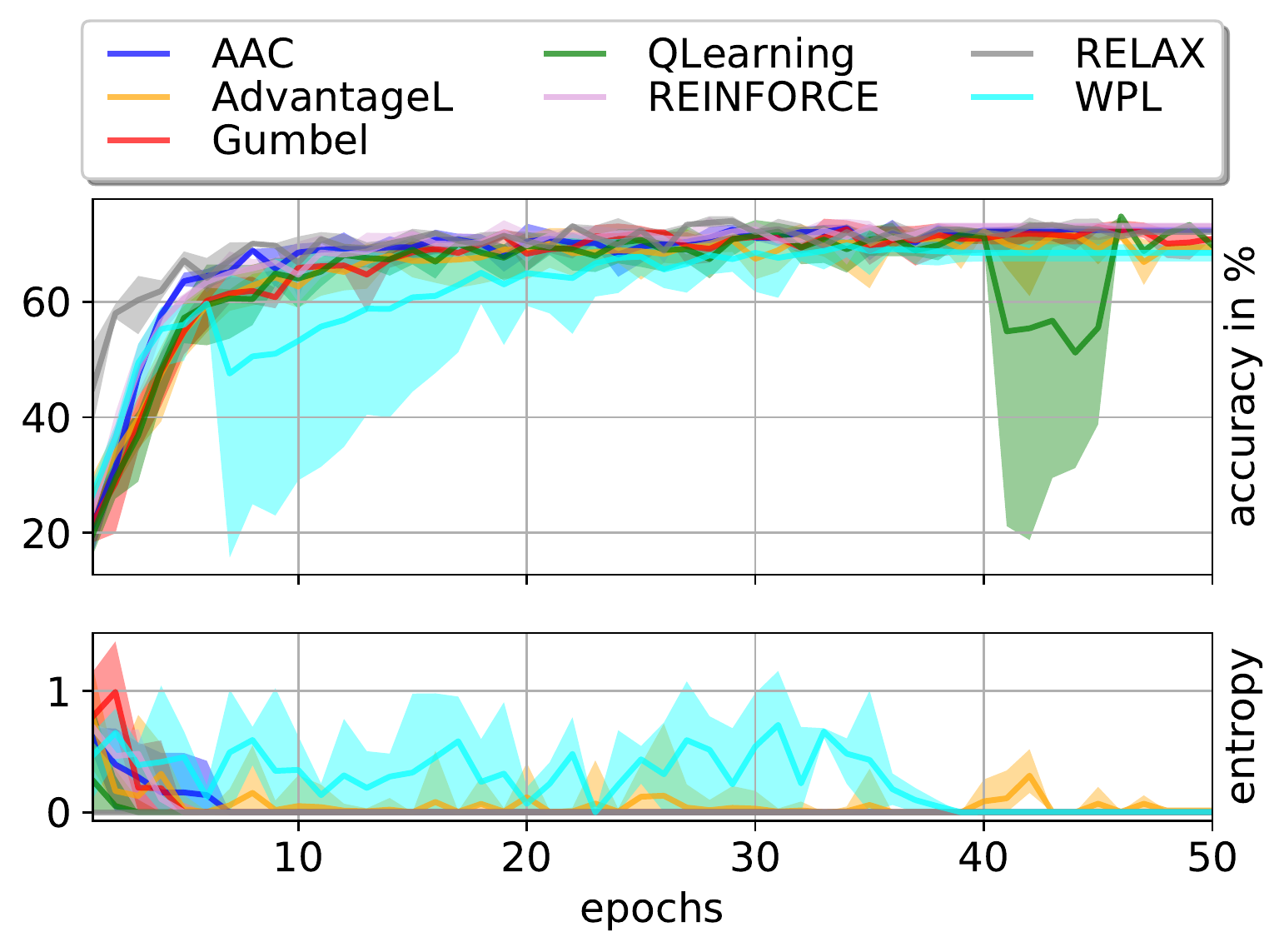}
    % \caption{Results on CIFAR 100 MTL (for multi-task experiments) and CIFAR 10 (for single-task experiments) for different decision making strategies; }
    \caption{Collapse on CIFAR with a dispatched architecture}
    \label{fig:res-collapse}
    \vspace{-3mm}
\end{wrapfigure}

Figure \ref{fig:res-dm-single-alt} already showed how different router training algorithms can lead to collapse in different domains. Additionally, consider Figure \ref{fig:res-collapse}, where the experiment is on CIFAR with a \textit{dispatched} architecture\footnote{The same fc layers are routed as in the other experiments. However, the dispatching action is based on the activation after the convolutional layers. The dispatched subrouters are tabular, and do not consider the intermediate activations.}. For this architecture, collapse is not a consequence of the routing algorithm, but of the general architecture, as \textit{all} algorithms lead to collapse. However, in difference to the results in Figure \ref{fig:res-dm-single-alt}, the results achieved are good, reaching standard non-routed performance.
This is a not surprising result, as collapse \textit{will} stabilize the router, and as there are no later activations to consider. This is reflected by the test-time selection entropy of nearly all algorithms going to zero within ten epochs. Surprisingly, this extends to the stochastic reparameterization algorithms Gumbel and RELAX which do not rely purely on rewards, further establishing that in the context of routing, Gumbel and RELAX behave `just as' any other PG algorithm.

The only algorithms that do not collapse completely are WPL and Advantage learning. We assume that WPL does not collapse as much as it was specifically designed for non-stationary returns, and that Advantage learning performs well as a consequence of the argument put forth in \ref{sec:routing-algs}.

\subsection{Overfitting}\label{sec:eval-overfitting}
\begin{wrapfigure}{r}{0.34\textwidth}
    \centering
    \vspace{-5mm}
    % \vspace{-12mm}
    \includegraphics[width=0.32\textwidth]{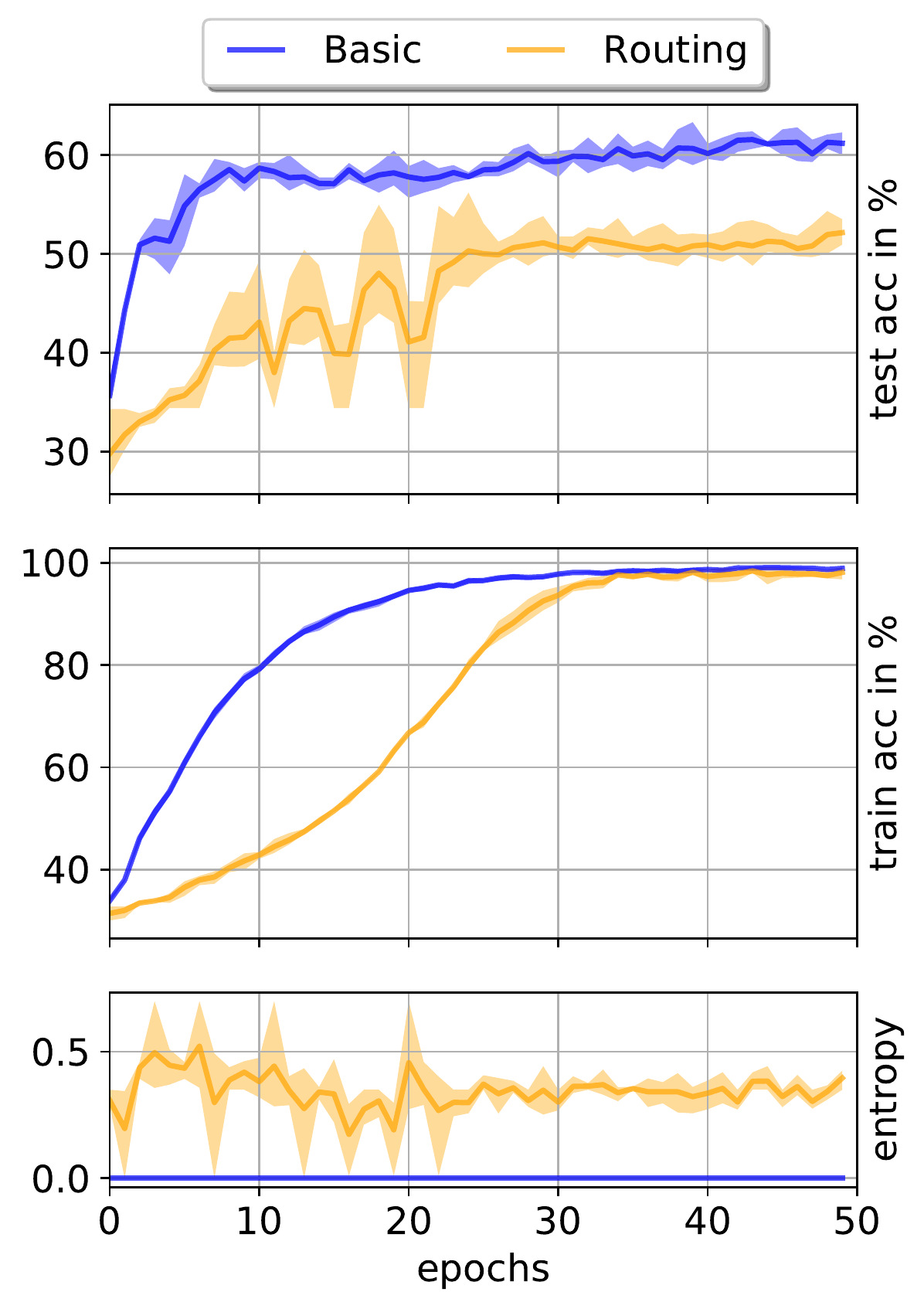}
    % \caption{Results on CIFAR 100 MTL (for multi-task experiments) and CIFAR 10 (for single-task experiments) for different decision making strategies; }
    \caption{Overfitting on SCI}
    \label{fig:res-overfit}
    \vspace{-7mm}
\end{wrapfigure}

As the main reference architectures for this section -- routing the classifier of a convolutional network, and routing the word representations of a natural language inference model -- were chosen because of their performance, they overfit only little. 

However, if we consider an sequence-to-sequence architecture for language inference with a dispatched routed classifier, we can show how routing can lead a model to overfit. Consider Figure \ref{fig:res-overfit}, which shows from top to bottom the test accuracy, train accuracy and test entropy on SCI. While both a basic, non-routed network and the routed network easily reach perfect train accuracy ($>99.5\%$), the routing architecture starts flattening out even earlier than the basic architecture, reaching accuracy around 10\% lower than the basic model, and over 20\% lower than the routed model.

As discussed in section \ref{sec:chall-overfitting}, overfitting is less of a problem in the presence of metainformation, as the router can ignore spurious activations in these cases. However, as the goal of modular architectures is to adaptively modularize without any kind of external information guiding the composition, future work should investigate how routing architectures can be regularized to achieve good generalization properties.

\section{Conclusion}
\paragraph{Summary of Results:}
Considering the results presented in this paper, it becomes clear how much of the performance of a routing network still depends on design- and hyperparameter decisions. However, many other results are also clear: 

\begin{itemize}\itemsep-0.2em 
    \item In the context of routing, reparameterization techniques behave similar to RL policy gradient approaches, and have oftentimes even identical performance and selection behavior.
    \item All policy gradient approaches, including reparameterization techniques, fall short compared to value-based learning. We established that the problem of PG approaches is their exploitation strategy. 
    \item Routing networks often need externally provided structure to properly stabilize and to achieve maximal performance given typical dataset sizes today. This structure may come in form of meta-information, or as a guided curricula of training samples.
    \item We evaluated the role that the reward design has on training reinforcement learning routers, observing that simple rewards tend to perform better than high information rewards that may be more difficult to interpret.
    \item The choice of optimization algorithm has a huge impact on the overall stability of routing networks, and presumably on any kind of compositional model of learning. Here, modern, more sophisticated algorithms can fail completely, with a simple SGD update performing the best.
    \item We investigated each of the challenges introduced and discussed in Section \ref{sec:challenges}, and illustrated how different design decisions can lead to instability, to collapse, and how the high expressivity of a routing network can lead to overfitting.
    \item We additionally identified a dilemma of flexibility for modular learning. This arises out of a need to balance the flexibility (or locality) of modular approaches to avoid both collapse and overfitting.
\end{itemize}

\paragraph{Open Questions:} Research into compositional architectures has accelerated in recent years. In this area, routing, as a general formulation of functional composition relying on trainable modules, can provide good insights into the challenges ahead. Stability, selection collapse, overfitting, mathematical justifications and learning over different architectures may only be some of the challenges that need to be solved before a fully compositional and modular architecture that is able to quickly adapt to \textit{any} new problem, task, or context can be designed. With this paper we hope to contribute an overview that is useful for establishing these problems within the research community and lay out some future steps. Moreover, our analysis in this paper reveals that decision making strategies for routing, different routing architectures, and new reward design strategies seem to be very promising directions for future research that have the potential to lead to significant improvements in general purpose models with compositional architectures.
\newpage
\bibliographystyle{humannat}
\bibliography{main}

\begin{thebibliography}{}

\bibitem[\protect\astroncite{Abdallah and Lesser}{2006}]{abdallah_wpl}
Abdallah, S. and V.~Lesser\leavevmode\nopagebreak\newline 2006.
\newblock Learning the task allocation game.
\newblock In {\em Proceedings of the fifth international joint conference on
  {Autonomous} agents and multiagent systems}, Pp.~ 850--857. ACM.

\bibitem[\protect\astroncite{Aertsen et~al.}{1989}]{aertsen1989}
Aertsen, A., G.~Gerstein, M.~Habib, and G.~Palm\leavevmode\nopagebreak\newline
  1989.
\newblock Dynamics of neuronal firing correlation: modulation of" effective
  connectivity".
\newblock {\em Journal of neurophysiology}, 61(5):900--917.

\bibitem[\protect\astroncite{Alet et~al.}{2018}]{modular_meta_learning}
Alet, F., T.~Lozano{-}P{\'{e}}rez, and L.~P.
  Kaelbling\leavevmode\nopagebreak\newline 2018.
\newblock Modular meta-learning.
\newblock {\em CoRR}, abs/1806.10166.

\bibitem[\protect\astroncite{Aljundi et~al.}{2017}]{Aljundi}
Aljundi, R., J.~Chakravarty, and T.~Tuytelaars\leavevmode\nopagebreak\newline
  2017.
\newblock Expert gate: Lifelong learning with a network of experts.
\newblock In {\em Proceedings CVPR 2017}, Pp.~ 3366--3375.

\bibitem[\protect\astroncite{Andreas et~al.}{2015}]{Andreas2015}
Andreas, J., M.~Rohrbach, T.~Darrell, and
  D.~Klein\leavevmode\nopagebreak\newline 2015.
\newblock Deep compositional question answering with neural module networks.
\newblock {\em CoRR}, abs/1511.02799.

\bibitem[\protect\astroncite{Bacon et~al.}{2017}]{optioncritic}
Bacon, P.-L., J.~Harb, and D.~Precup\leavevmode\nopagebreak\newline 2017.
\newblock The option-critic architecture.
\newblock {\em AAAI}.

\bibitem[\protect\astroncite{Baker et~al.}{2017}]{bakerICLR}
Baker, B., O.~Gupta, N.~Naik, and R.~Raskar\leavevmode\nopagebreak\newline
  2017.
\newblock Designing neural network architectures using reinforcement learning.
\newblock {\em ICLR}.

\bibitem[\protect\astroncite{Bechtel and
  Abrahamsen}{2002}]{bechtel2002connectionism}
Bechtel, W. and A.~Abrahamsen\leavevmode\nopagebreak\newline 2002.
\newblock {\em Connectionism and the mind: Parallel processing, dynamics, and
  evolution in networks}.
\newblock Blackwell Publishing.

\bibitem[\protect\astroncite{Bechtel and
  Richardson}{2010}]{BechtelRichardson2010}
Bechtel, W. and R.~C. Richardson\leavevmode\nopagebreak\newline 2010.
\newblock {\em Discovering Complexity: Decomposition and Localization as
  Strategies in Scientific Research}.
\newblock Mit Press.

\bibitem[\protect\astroncite{Bender et~al.}{2018}]{bender2018}
Bender, G., P.-J. Kindermans, B.~Zoph, V.~Vasudevan, and
  Q.~Le\leavevmode\nopagebreak\newline 2018.
\newblock Understanding and simplifying one-shot architecture search.
\newblock In {\em International Conference on Machine Learning}, Pp.~ 549--558.

\bibitem[\protect\astroncite{Bengio et~al.}{2015}]{EBengio}
Bengio, E., P.~Bacon, J.~Pineau, and D.~Precup\leavevmode\nopagebreak\newline
  2015.
\newblock Conditional computation in neural networks for faster models.
\newblock {\em CoRR}, abs/1511.06297.

\bibitem[\protect\astroncite{Bengio et~al.}{2013}]{conditionalcomputation}
Bengio, Y., N.~L{\'{e}}onard, and A.~C.
  Courville\leavevmode\nopagebreak\newline 2013.
\newblock Estimating or propagating gradients through stochastic neurons for
  conditional computation.
\newblock {\em CoRR}, abs/1308.3432.

\bibitem[\protect\astroncite{Bengio et~al.}{2009}]{curriculumlearning}
Bengio, Y., J.~Louradour, R.~Collobert, and
  J.~Weston\leavevmode\nopagebreak\newline 2009.
\newblock Curriculum learning.
\newblock In {\em Proceedings of the 26th annual international conference on
  machine learning}, Pp.~ 41--48. ACM.

\bibitem[\protect\astroncite{Brock et~al.}{2017}]{SMASH}
Brock, A., T.~Lim, J.~M. Ritchie, and N.~Weston\leavevmode\nopagebreak\newline
  2017.
\newblock {SMASH:} one-shot model architecture search through hypernetworks.
\newblock {\em CoRR}, abs/1708.05344.

\bibitem[\protect\astroncite{Buschman and Miller}{2010}]{bio2010}
Buschman, T.~J. and E.~K. Miller\leavevmode\nopagebreak\newline 2010.
\newblock Shifting the spotlight of attention: evidence for discrete
  computations in cognition.
\newblock {\em Frontiers in human neuroscience}, 4.

\bibitem[\protect\astroncite{Cases et~al.}{2019}]{routingnaacl}
Cases, I., C.~Rosenbaum, M.~Riemer, A.~Geiger, T.~Klinger, A.~Tamkin, O.~Li,
  S.~Agarwal, J.~D. Greene, D.~Jurafsky, C.~Potts, and
  L.~Karttunen\leavevmode\nopagebreak\newline 2019.
\newblock Recursive routing networks: Learning to compose modules for language
  understanding.
\newblock In {\em Proceedings of the 2019 Conference of the North American
  Chapter of the Association for Computational Linguistics: Human Language
  Technologies, Volume 1 (Long Papers)}.

\bibitem[\protect\astroncite{Chang et~al.}{2019}]{crl}
Chang, M., A.~Gupta, S.~Levine, and T.~L.
  Griffiths\leavevmode\nopagebreak\newline 2019.
\newblock Automatically composing representation transformations as a means for
  generalization.
\newblock In {\em International Conference on Learning Representations}.

\bibitem[\protect\astroncite{Cortes et~al.}{2016}]{Adanet}
Cortes, C., X.~Gonzalvo, V.~Kuznetsov, M.~Mohri, and
  S.~Yang\leavevmode\nopagebreak\newline 2016.
\newblock Adanet: Adaptive structural learning of artificial neural networks.
\newblock {\em arXiv preprint arXiv:1607.01097}.

\bibitem[\protect\astroncite{Davis and Arel}{2013}]{davis2013low}
Davis, A. and I.~Arel\leavevmode\nopagebreak\newline 2013.
\newblock Low-rank approximations for conditional feedforward computation in
  deep neural networks.
\newblock {\em arXiv preprint arXiv:1312.4461}.

\bibitem[\protect\astroncite{Engel et~al.}{2001}]{engel2001}
Engel, A.~K., P.~Fries, and W.~Singer\leavevmode\nopagebreak\newline 2001.
\newblock Dynamic predictions: oscillations and synchrony in top--down
  processing.
\newblock {\em Nature Reviews Neuroscience}, 2(10):704.

\bibitem[\protect\astroncite{Eysenbach et~al.}{2018}]{DIAYAN}
Eysenbach, B., A.~Gupta, J.~Ibarz, and S.~Levine\leavevmode\nopagebreak\newline
  2018.
\newblock Diversity is all you need: Learning skills without a reward function.
\newblock {\em arXiv preprint arXiv:1802.06070}.

\bibitem[\protect\astroncite{Farah}{1994}]{Farah1994}
Farah, M.~J.\leavevmode\nopagebreak\newline 1994.
\newblock Neuropsychological inference with an interactive brain: A critique of
  the "locality" assumption.
\newblock {\em Behavioral and Brain Sciences}, 17(1), 43-104.

\bibitem[\protect\astroncite{Fernando et~al.}{2017}]{Pathnet}
Fernando, C., D.~Banarse, C.~Blundell, Y.~Zwols, D.~Ha, A.~A. Rusu, A.~Pritzel,
  and D.~Wierstra\leavevmode\nopagebreak\newline 2017.
\newblock Pathnet: Evolution channels gradient descent in super neural
  networks.
\newblock {\em arXiv preprint arXiv:1701.08734}.

\bibitem[\protect\astroncite{Florensa et~al.}{2017}]{florensa}
Florensa, C., Y.~Duan, and P.~Abbeel\leavevmode\nopagebreak\newline 2017.
\newblock Stochastic neural networks for hierarchical reinforcement learning.
\newblock {\em arXiv preprint arXiv:1704.03012}.

\bibitem[\protect\astroncite{Fodor}{1975}]{Fodor1975-FODTLO}
Fodor, J.~A.\leavevmode\nopagebreak\newline 1975.
\newblock {\em The Language of Thought}.
\newblock Harvard University Press.

\bibitem[\protect\astroncite{Fodor}{1983}]{Fodor1983-FODTMO}
Fodor, J.~A.\leavevmode\nopagebreak\newline 1983.
\newblock {\em The Modularity of Mind}.
\newblock MIT Press.

\bibitem[\protect\astroncite{Fodor and Pylyshyn}{1988}]{Fodor1988-FODCAC}
Fodor, J.~A. and Z.~W. Pylyshyn\leavevmode\nopagebreak\newline 1988.
\newblock Connectionism and cognitive architecture: A critical analysis.
\newblock {\em Cognition}, 28(1-2):3--71.

\bibitem[\protect\astroncite{Fries}{2005}]{fries2005}
Fries, P.\leavevmode\nopagebreak\newline 2005.
\newblock A mechanism for cognitive dynamics: neuronal communication through
  neuronal coherence.
\newblock {\em Trends in cognitive sciences}, 9(10):474--480.

\bibitem[\protect\astroncite{Garcia and Thomas}{2019}]{francisco_exploration}
Garcia, F.~M. and P.~S. Thomas\leavevmode\nopagebreak\newline 2019.
\newblock A meta-mdp approach to exploration for lifelong reinforcement
  learning.
\newblock {\em CoRR}, abs/1902.00843.

\bibitem[\protect\astroncite{Grathwohl et~al.}{2018}]{relax-estimator}
Grathwohl, W., D.~Choi, Y.~Wu, G.~Roeder, and
  D.~Duvenaud\leavevmode\nopagebreak\newline 2018.
\newblock Backpropagation through the void: Optimizing control variates for
  black-box gradient estimation.
\newblock In {\em International Conference on Learning Representations}.

\bibitem[\protect\astroncite{Gregor et~al.}{2016}]{VIC}
Gregor, K., D.~J. Rezende, and D.~Wierstra\leavevmode\nopagebreak\newline 2016.
\newblock Variational intrinsic control.
\newblock {\em arXiv preprint arXiv:1611.07507}.

\bibitem[\protect\astroncite{Gurney et~al.}{2001}]{bio2001}
Gurney, K., T.~J. Prescott, and P.~Redgrave\leavevmode\nopagebreak\newline
  2001.
\newblock A computational model of action selection in the basal ganglia. i. a
  new functional anatomy.
\newblock {\em Biological cybernetics}, 84(6):401--410.

\bibitem[\protect\astroncite{Harutyunyan et~al.}{2019}]{termcritic}
Harutyunyan, A., W.~Dabney, D.~Borsa, N.~Heess, R.~Munos, and
  D.~Precup\leavevmode\nopagebreak\newline 2019.
\newblock The termination critic.
\newblock {\em arXiv preprint arXiv:1902.09996}.

\bibitem[\protect\astroncite{Hausman et~al.}{2018}]{hausman}
Hausman, K., J.~T. Springenberg, Z.~Wang, N.~Heess, and
  M.~Riedmiller\leavevmode\nopagebreak\newline 2018.
\newblock Learning an embedding space for transferable robot skills.
\newblock In {\em International Conference on Learning Representations}.

\bibitem[\protect\astroncite{Hay et~al.}{2014}]{metamdp}
Hay, N., S.~Russell, D.~Tolpin, and S.~E.
  Shimony\leavevmode\nopagebreak\newline 2014.
\newblock Selecting computations: Theory and applications.
\newblock {\em arXiv preprint arXiv:1408.2048}.

\bibitem[\protect\astroncite{Jacobs et~al.}{1991a}]{JACOBS1991219}
Jacobs, R.~A., M.~I. Jordan, and A.~G. Barto\leavevmode\nopagebreak\newline
  1991a.
\newblock Task decomposition through competition in a modular connectionist
  architecture: The what and where vision tasks.
\newblock {\em Cognitive Science}, 15(2):219 -- 250.

\bibitem[\protect\astroncite{Jacobs et~al.}{1991b}]{Hinton91}
Jacobs, R.~A., M.~I. Jordan, S.~J. Nowlan, and G.~E.
  Hinton\leavevmode\nopagebreak\newline 1991b.
\newblock Adaptive mixtures of local experts.
\newblock {\em Neural computation}, 3(1):79--87.

\bibitem[\protect\astroncite{Jang et~al.}{2016}]{jang2016categorical}
Jang, E., S.~Gu, and B.~Poole\leavevmode\nopagebreak\newline 2016.
\newblock Categorical reparameterization with gumbel-softmax.
\newblock {\em arXiv preprint arXiv:1611.01144}.

\bibitem[\protect\astroncite{Jordan and Jacobs}{1994}]{Jordan94}
Jordan, M.~I. and R.~A. Jacobs\leavevmode\nopagebreak\newline 1994.
\newblock Hierarchical mixtures of experts and the em algorithm.
\newblock {\em Neural computation}, 6(2):181--214.

\bibitem[\protect\astroncite{Kell et~al.}{2018}]{Kell2018}
Kell, A.~J., D.~L. Yamins, E.~N. Shook, S.~V. Norman-Haignere, and J.~H.
  McDermott\leavevmode\nopagebreak\newline 2018.
\newblock A task-optimized neural network replicates human auditory behavior,
  predicts brain responses, and reveals a cortical processing hierarchy.
\newblock {\em Neuron}, 98(3):630 -- 644.e16.

\bibitem[\protect\astroncite{Kirsch et~al.}{2018}]{modularnets}
Kirsch, L., J.~Kunze, and D.~Barber\leavevmode\nopagebreak\newline 2018.
\newblock Modular networks: Learning to decompose neural computation.
\newblock In {\em Advances in Neural Information Processing Systems}, Pp.~
  2414--2423.

\bibitem[\protect\astroncite{Kostas et~al.}{2019}]{kostas2019reinforcement}
Kostas, J., C.~Nota, and P.~S. Thomas\leavevmode\nopagebreak\newline 2019.
\newblock Reinforcement learning without backpropagation or a clock.
\newblock {\em arXiv preprint arXiv:1902.05650}.

\bibitem[\protect\astroncite{Liang et~al.}{2018}]{evol_multitask_nets}
Liang, J.~Z., E.~Meyerson, and R.~Miikkulainen\leavevmode\nopagebreak\newline
  2018.
\newblock Evolutionary architecture search for deep multitask networks.
\newblock {\em CoRR}, abs/1803.03745.

\bibitem[\protect\astroncite{Liu et~al.}{2018}]{Liu_2018_ECCV}
Liu, C., B.~Zoph, M.~Neumann, J.~Shlens, W.~Hua, L.-J. Li, L.~Fei-Fei,
  A.~Yuille, J.~Huang, and K.~Murphy\leavevmode\nopagebreak\newline 2018.
\newblock Progressive neural architecture search.
\newblock In {\em The European Conference on Computer Vision (ECCV)}.

\bibitem[\protect\astroncite{Maddison et~al.}{2016}]{maddison2016concrete}
Maddison, C.~J., A.~Mnih, and Y.~W. Teh\leavevmode\nopagebreak\newline 2016.
\newblock The concrete distribution: A continuous relaxation of discrete random
  variables.
\newblock {\em arXiv preprint arXiv:1611.00712}.

\bibitem[\protect\astroncite{Marcus}{2001}]{Marcus2001-MARTAM-10}
Marcus, G.~F.\leavevmode\nopagebreak\newline 2001.
\newblock {\em The Algebraic Mind}.
\newblock MIT Press.

\bibitem[\protect\astroncite{Miikkulainen}{1993}]{Miikkulainen1993}
Miikkulainen, R.\leavevmode\nopagebreak\newline 1993.
\newblock {\em Subsymbolic natural language processing - an integrated model of
  scripts, lexicon, and memory}, Neural network modeling and connectionism.
\newblock {MIT} Press.

\bibitem[\protect\astroncite{Miikkulainen et~al.}{2017}]{evolving}
Miikkulainen, R., J.~Liang, E.~Meyerson, A.~Rawal, D.~Fink, O.~Francon,
  B.~Raju, A.~Navruzyan, N.~Duffy, and B.~Hodjat\leavevmode\nopagebreak\newline
  2017.
\newblock Evolving deep neural networks.
\newblock {\em arXiv preprint arXiv:1703.00548}.

\bibitem[\protect\astroncite{Misra et~al.}{2016}]{CrossStitch}
Misra, I., A.~Shrivastava, A.~Gupta, and
  M.~Hebert\leavevmode\nopagebreak\newline 2016.
\newblock Cross-stitch networks for multi-task learning.
\newblock In {\em Proceedings of the IEEE Conference on Computer Vision and
  Pattern Recognition}, Pp.~ 3994--4003.

\bibitem[\protect\astroncite{Pham et~al.}{2018}]{pham2018}
Pham, H., M.~Y. Guan, B.~Zoph, Q.~V. Le, and
  J.~Dean\leavevmode\nopagebreak\newline 2018.
\newblock Efficient neural architecture search via parameter sharing.
\newblock {\em arXiv preprint arXiv:1802.03268}.

\bibitem[\protect\astroncite{Precup et~al.}{2000}]{precup_importance}
Precup, D., R.~S. Sutton, and S.~P. Singh\leavevmode\nopagebreak\newline 2000.
\newblock Eligibility traces for off-policy policy evaluation.
\newblock In {\em Proceedings of the Seventeenth International Conference on
  Machine Learning}, Pp.~ 759--766. Morgan Kaufmann Publishers Inc.

\bibitem[\protect\astroncite{Rajendran et~al.}{2017}]{AAT}
Rajendran, J., P.~Prasanna, B.~Ravindran, and M.~M.
  Khapra\leavevmode\nopagebreak\newline 2017.
\newblock {ADAAPT:} attend, adapt, and transfer: Attentative deep architecture
  for adaptive policy transfer from multiple sources in the same domain.
\newblock {\em ICLR}, abs/1510.02879.

\bibitem[\protect\astroncite{Ramachandran and
  Le}{2019}]{ramachandran2018diversity}
Ramachandran, P. and Q.~V. Le\leavevmode\nopagebreak\newline 2019.
\newblock Diversity and depth in per-example routing models.
\newblock In {\em International Conference on Learning Representations}.

\bibitem[\protect\astroncite{Rice}{1976}]{algorithmselection}
Rice, J.~R.\leavevmode\nopagebreak\newline 1976.
\newblock The algorithm selection problem.
\newblock In {\em Advances in computers}, volume~15, Pp.~ 65--118.
\newblock Elsevier.

\bibitem[\protect\astroncite{Riemer et~al.}{2019}]{MER}
Riemer, M., I.~Cases, R.~Ajemian, M.~Liu, I.~Rish, Y.~Tu, and
  G.~Tesauro\leavevmode\nopagebreak\newline 2019.
\newblock Learning to learn without forgetting by maximizing transfer and
  minimizing interference.
\newblock In {\em In International Conference on Learning Representations
  (ICLR)}.

\bibitem[\protect\astroncite{Riemer et~al.}{2018}]{abstractoptions}
Riemer, M., M.~Liu, and G.~Tesauro\leavevmode\nopagebreak\newline 2018.
\newblock Learning abstract options.
\newblock In {\em Advances in Neural Information Processing Systems}, Pp.~
  10424--10434.

\bibitem[\protect\astroncite{Riemer et~al.}{2016}]{Riemer2016}
Riemer, M., A.~Vempaty, F.~Calmon, F.~Heath, R.~Hull, and
  E.~Khabiri\leavevmode\nopagebreak\newline 2016.
\newblock Correcting forecasts with multifactor neural attention.
\newblock In {\em International Conference on Machine Learning}, Pp.~
  3010--3019.

\bibitem[\protect\astroncite{Rosenbaum et~al.}{2017}]{routingnets}
Rosenbaum, C., T.~Klinger, and M.~Riemer\leavevmode\nopagebreak\newline 2017.
\newblock Routing networks: Adaptive selection of non-linear functions for
  multi-task learning.
\newblock {\em CoRR}, abs/1711.01239.

\bibitem[\protect\astroncite{Ruder et~al.}{2017}]{SLUICE}
Ruder, S., J.~Bingel, I.~Augenstein, and
  A.~S{\o}gaard\leavevmode\nopagebreak\newline 2017.
\newblock Sluice networks: Learning what to share between loosely related
  tasks.
\newblock {\em arXiv preprint arXiv:1705.08142}.

\bibitem[\protect\astroncite{Salinas and Sejnowski}{2001}]{salinas2001}
Salinas, E. and T.~J. Sejnowski\leavevmode\nopagebreak\newline 2001.
\newblock Correlated neuronal activity and the flow of neural information.
\newblock {\em Nature reviews neuroscience}, 2(8):539.

\bibitem[\protect\astroncite{Shallice}{1988}]{shallice1988neuropsychology}
Shallice, T.\leavevmode\nopagebreak\newline 1988.
\newblock {\em From neuropsychology to mental structure}.
\newblock Cambridge University Press.

\bibitem[\protect\astroncite{Shazeer et~al.}{2017}]{largeneuralnets}
Shazeer, N., A.~Mirhoseini, K.~Maziarz, A.~Davis, Q.~V. Le, G.~E. Hinton, and
  J.~Dean\leavevmode\nopagebreak\newline 2017.
\newblock Outrageously large neural networks: The sparsely-gated
  mixture-of-experts layer.
\newblock {\em CoRR}, abs/1701.06538.

\bibitem[\protect\astroncite{Siegel and K{\"o}nig}{2003}]{siegel2003}
Siegel, M. and P.~K{\"o}nig\leavevmode\nopagebreak\newline 2003.
\newblock A functional gamma-band defined by stimulus-dependent synchronization
  in area 18 of awake behaving cats.
\newblock {\em Journal of Neuroscience}, 23(10):4251--4260.

\bibitem[\protect\astroncite{Smolensky}{1988}]{Smolensky1988-SMOOTP-2}
Smolensky, P.\leavevmode\nopagebreak\newline 1988.
\newblock On the proper treatment of connectionism.
\newblock {\em Behavioral and Brain Sciences}, 11(1):1--23.

\bibitem[\protect\astroncite{Stocco et~al.}{2010}]{bio2010_2}
Stocco, A., C.~Lebiere, and J.~R. Anderson\leavevmode\nopagebreak\newline 2010.
\newblock Conditional routing of information to the cortex: A model of the
  basal ganglia’s role in cognitive coordination.
\newblock {\em Psychological review}, 117(2):541.

\bibitem[\protect\astroncite{Stollenga et~al.}{2014}]{Schmidhuber}
Stollenga, M.~F., J.~Masci, F.~Gomez, and
  J.~Schmidhuber\leavevmode\nopagebreak\newline 2014.
\newblock Deep networks with internal selective attention through feedback
  connections.
\newblock In {\em Advances in Neural Information Processing Systems 27},
  Z.~Ghahramani, M.~Welling, C.~Cortes, N.~D. Lawrence, and K.~Q. Weinberger,
  eds., Pp.~ 3545--3553.
\newblock Curran Associates, Inc.

\bibitem[\protect\astroncite{Sutton and Barto}{1998}]{suttonbarto}
Sutton, R.~S. and A.~G. Barto\leavevmode\nopagebreak\newline 1998.
\newblock {\em Introduction to Reinforcement Learning}, 1st edition.
\newblock Cambridge, MA, USA: MIT Press.

\bibitem[\protect\astroncite{Sutton et~al.}{1999}]{Options}
Sutton, R.~S., D.~Precup, and S.~Singh\leavevmode\nopagebreak\newline 1999.
\newblock Between mdps and semi-mdps: A framework for temporal abstraction in
  reinforcement learning.
\newblock {\em Artificial intelligence}, 112(1-2):181--211.

\bibitem[\protect\astroncite{Thomas}{2011}]{thomas2011policy}
Thomas, P.~S.\leavevmode\nopagebreak\newline 2011.
\newblock Policy gradient coagent networks.
\newblock In {\em Advances in Neural Information Processing Systems}, Pp.~
  1944--1952.

\bibitem[\protect\astroncite{Thomas and Barto}{2011}]{coagents}
Thomas, P.~S. and A.~G. Barto\leavevmode\nopagebreak\newline 2011.
\newblock Conjugate markov decision processes.
\newblock In {\em Proceedings of the 28th International Conference on Machine
  Learning (ICML-11)}, Pp.~ 137--144.

\bibitem[\protect\astroncite{Tiesinga et~al.}{2002}]{tiesinga2002}
Tiesinga, P., J.-M. Fellous, J.~Jos, and
  T.~Sejnowski\leavevmode\nopagebreak\newline 2002.
\newblock Information transfer in entrained cortical neurons.
\newblock {\em Network: Computation in Neural Systems}, 13(1):41--66.

\bibitem[\protect\astroncite{Tononi et~al.}{1994}]{Tononi1994}
Tononi, G., O.~Sporns, and G.~M. Edelman\leavevmode\nopagebreak\newline 1994.
\newblock A measure for brain complexity: relating functional segregation and
  integration in the nervous system.
\newblock {\em Proceedings of the National Academy of Sciences},
  91(11):5033--5037.

\bibitem[\protect\astroncite{Touretzky and Hinton}{1988}]{TouretzkyHinton1988}
Touretzky, D.~S. and G.~E. Hinton\leavevmode\nopagebreak\newline 1988.
\newblock A distributed connectionist production system.
\newblock {\em Cognitive Science}, 12(3):423--466.

\bibitem[\protect\astroncite{Tucker et~al.}{2017}]{rebar-estimator}
Tucker, G., A.~Mnih, C.~J. Maddison, and
  J.~Sohl{-}Dickstein\leavevmode\nopagebreak\newline 2017.
\newblock {REBAR:} low-variance, unbiased gradient estimates for discrete
  latent variable models.
\newblock {\em CoRR}, abs/1703.07370.

\bibitem[\protect\astroncite{Usrey and Reid}{1999}]{usrey1999}
Usrey, W.~M. and R.~C. Reid\leavevmode\nopagebreak\newline 1999.
\newblock Synchronous activity in the visual system.
\newblock {\em Annual review of physiology}, 61(1):435--456.

\bibitem[\protect\astroncite{Williams}{1992}]{williams_simple_1992}
Williams, R.~J.\leavevmode\nopagebreak\newline 1992.
\newblock Simple statistical gradient-following algorithms for connectionist
  reinforcement learning.
\newblock {\em Machine learning}, 8(3-4):229--256.

\bibitem[\protect\astroncite{Zhang et~al.}{2016}]{overfitting}
Zhang, C., S.~Bengio, M.~Hardt, B.~Recht, and
  O.~Vinyals\leavevmode\nopagebreak\newline 2016.
\newblock Understanding deep learning requires rethinking generalization.
\newblock {\em CoRR}, abs/1611.03530.

\bibitem[\protect\astroncite{Zoph and Le}{2017}]{zophICLR}
Zoph, B. and Q.~V. Le\leavevmode\nopagebreak\newline 2017.
\newblock Neural architecture search with reinforcement learning.
\newblock {\em ICLR}.

\end{thebibliography}

% \clearpage

% \setcounter{table}{0}
% \renewcommand{\thetable}{A\arabic{table}}
% \setcounter{figure}{0}
% \renewcommand{\thefigure}{A\arabic{figure}}

% \appendix

% \section*{Appendices}
% \label{sec:appendix}
% \input{appendix_architecture.tex}
% \input{appendix_multi.tex}
% \input{appendix_implicatives.tex}

\end{document}